\documentclass[11pt, letterpaper]{berkeley}

\usepackage{amsmath}
\usepackage{hyperref}
\usepackage{url}
\usepackage{graphicx}
\usepackage{booktabs}
\usepackage{wrapfig}
\usepackage{multirow}
\usepackage{lipsum}
\usepackage{subcaption}
\usepackage{amssymb}
\usepackage{tikz} 
\usepackage{xcolor}
\usepackage{multicol}
\usepackage{natbib}
\usepackage{listings}
\usepackage{fancyvrb}
\usepackage{fvextra}
\usepackage{tcolorbox}
\usepackage{xcolor}
\usepackage{hyperref}

\definecolor{brownishyellow}{RGB}{245, 194, 66}
\newtcolorbox{solutionbox}{
  colframe=black,
  colback=gray!10,
  boxrule=1pt,
  arc=0pt,
  title=,
  fonttitle=\bfseries
}

\newenvironment{sol}
  {\begin{solutionbox}}
  {\end{solutionbox}}

\newcommand{\BeginSol}{\begin{sol}}
\newcommand{\EndSol}{\end{sol}}

\definecolor{mylightgray}{RGB}{240,240,240}

\usepackage{tcolorbox}
\tcbuselibrary{skins}
\newtcolorbox{text_full}[1]{
    enhanced,
    left=4mm,
    right=4mm,
    top=2mm,
    bottom=2mm,
    boxsep=0mm,
    rounded corners,
    title=#1,
    fontupper=\footnotesize\linespread{0.9}\fontfamily{lmr}\selectfont,
    }
\newtcolorbox{text_half}[1]{
    enhanced,
    left=2mm,
    right=2mm,
    top=2mm,
    bottom=2mm,
    boxsep=0mm,
    rounded corners,
    title=#1,
    width=0.475\textwidth,
    fontupper=\footnotesize\linespread{0.9}\fontfamily{lmr}\selectfont,
    }

\newcommand\blfootnote[1]{%
  \begingroup
  \renewcommand\thefootnote{}\footnote{#1}%
  \addtocounter{footnote}{-1}%
  \endgroup
}

\title{Towards System 2 Reasoning in LLMs: Learning How to Think With Meta Chain-of-Thought}

\author[2]{Violet Xiang}
\author[3]{Charlie Snell}
\author[2]{Kanishk Gandhi}
\author[1]{Alon Albalak}
\author[2]{Anikait Singh}
\author[1]{Chase Blagden}
\author[1]{Duy Phung}
\author[2,1]{Rafael Rafailov}
\author[1]{Nathan Lile}
\author[1]{Dakota Mahan}
\author[1]{Louis Castricato}
\author[2]{Jan-Philipp Fränken}
\author[2]{Nick Haber}
\author[2]{Chelsea Finn}

\affil[1]{SynthLabs.ai}
\affil[2]{Stanford University}
\affil[3]{UC Berkeley}

\begin{document}

\blfootnote{Authors are ordered randomly. Correspondence to \href{mailto:team@synthlabs.ai}{team@synthlabs.ai}.}

\begin{abstract}

We propose a novel framework, Meta Chain-of-Thought (Meta-CoT), which extends traditional Chain-of-Thought (CoT) by explicitly modeling the underlying reasoning required to arrive at a particular CoT. We present empirical evidence from state-of-the-art models exhibiting behaviors consistent with in-context search, and explore methods for producing Meta-CoT via process supervision, synthetic data generation, and search algorithms. Finally, we outline a concrete pipeline for training a model to produce Meta-CoTs, incorporating instruction tuning with linearized search traces and reinforcement learning post-training. Finally, we discuss open research questions, including scaling laws, verifier roles, and the potential for discovering novel reasoning algorithms. This work provides a theoretical and practical roadmap to enable Meta-CoT in LLMs, paving the way for more powerful and human-like reasoning in artificial intelligence.

\end{abstract}

\maketitle

\begin{flushright}
\textit{Give a man a fish and you feed him for a day; \\ teach a man to fish and you feed him for a lifetime. \\ -Proverb}
\end{flushright}


\tableofcontents

\clearpage

\section{Introduction}

\subsection{Motivation}
A key aspect of the current era of Large Language Models has been the foundational principle of next-token prediction~\citep{elman1990finding, jordan1997serial}. That is, tokenizing text (or other continuous modalities) into a discrete sequence in the following way:

$$\text{"The quick brown fox jumps over the lazy dog."}\to \mathbf{y}_1, \mathbf{y}_2, \ldots, \mathbf{y}_n,$$

where $\mathbf{y}_i$ are elements of some finite vocabulary and, subsequently, train a large parameterized neural network $p_{\theta}$ (transformer) model with the following maximum likelihood objective:

$$\mathcal{L}_{\theta} =\mathbb{E}_{\mathcal{D}_{\text{train}}}\left[-\sum_t \log p_{\theta}(\mathbf{y}_{t+1}|\mathbf{y}_{\leq t})\right].$$

Behind this approach is a simple principle often abbreviated as "compression is intelligence", or the model must approximate the distribution of data and perform implicit reasoning in its activations in order to predict the next token (see Solomonoff Induction;~\citealt{solomonoff1964formal}). That is, language models learn the implicit meaning in text, as opposed to the early belief some researchers held that sequence-to-sequence models (including transformers) simply fit correlations between sequential words.

A fundamental question, however, is: What is the relationship between the complexity of the underlying data stream and the model's capability to learn the data-generating algorithm? While natural language has some irreducible entropy, this question holds even in deterministic scenarios (as demonstrated below). Mathematical reasoning is a good domain to demonstrate and evaluate these questions. One can ask an LLM to generate a response to questions like:

``What is 1+2?''

\noindent to which most immediately answer ``3''. We can increase the complexity of the question by asking, for example: 
\begin{tcolorbox}[colframe=black, colback=white, boxrule=1pt, arc=0pt]
``What is the value of \[
\frac{(x^2 - 1)(x + 1)}{x^3 - x} - \frac{1}{x}
\]
evaluated at $x=\pi$?''
\end{tcolorbox}

\noindent This is actually a simple question since the expression is canonically equal to 1, yet even powerful LLMs such as GPT-4o and Claude fail to answer it correctly even once. That is, the \textbf{true} conditional generative process $p(\mathbf{y}_{t+1}|\mathbf{y}_t)$ could have arbitrarily high computational complexity even when it is deterministic as in the example above. In contrast, if we instruct models to ``think step by step'' and produce a chain-of-thought we can significantly improve their capabilities producing traces like the following:

\BeginSol
\noindent We are tasked with evaluating 
\[
\frac{(x^2 - 1)(x + 1)}{x^3 - x} - \frac{1}{x}
\]
at \(x = \pi\). Let us simplify the expression before substituting \(x = \pi\).

\textbf{Step 1: Simplify \((x^2 - 1)(x + 1)\)} \\

The numerator of the first term can be rewritten:
\[
x^2 - 1 = (x - 1)(x + 1),
\]
so
\[
(x^2 - 1)(x + 1) = (x - 1)(x + 1)^2.
\]
\textbf{Step 2: Simplify \(x^3 - x\) in the denominator} \\
Factorize \(x^3 - x\):
\[
x^3 - x = x(x^2 - 1) = x(x - 1)(x + 1).
\]

Thus, the first term becomes:
\[
\frac{(x^2 - 1)(x + 1)}{x^3 - x} = \frac{(x - 1)(x + 1)^2}{x(x - 1)(x + 1)}.
\]

Cancel \((x - 1)\) and one factor of \((x + 1)\):
\[
\frac{(x^2 - 1)(x + 1)}{x^3 - x} = \frac{x + 1}{x}.
\]

\textbf{Step 3: Subtract \(\frac{1}{x}\)} \\
Now substitute this simplified result into the full expression:
\[
\frac{x + 1}{x} - \frac{1}{x}.
\]

Combine the terms over the common denominator \(x\):
\[
\frac{x + 1}{x} - \frac{1}{x} = \frac{x + 1 - 1}{x} = \frac{x}{x} = 1.
\]
\textbf{Step 4: Evaluate at \(x = \pi\)}
The expression simplifies to \(1\), which is independent of \(x\). Thus, the value at \(x = \pi\) is:
\[
\boxed{1}.
\]
\EndSol

\noindent There have been multiple explanations for this phenomenon~\citep{merrill2023expresssive,prystawski2024think,ye2024physicslanguagemodels21,Dutta2024HowTT}, however we focus on the complexity hypothesis~\citep{merrill2023expresssive}. Specifically, a CoT expansion allows for (in-theory) arbitrarily large amounts of compute to be deployed towards the prediction of answer tokens. A large theoretical literature has been developed arguing for the representational complexity of transformers with and without CoT~\citep{merrill2023expresssive, nowak2024representational, li2024chainthoughtempowerstransformers}. The literature says that endowing LLMs with CoT allows them to represent new orders of complexity and (with assumptions, such as infinite memory) may even make them Turing complete. However, in practice, today's LLMs can only reliably solve problems of limited complexity~\citep{snell2024scalingllmtesttimecompute}.

\subsection{Outline}

In this paper, we investigate the limitations of current LLMs in handling complex reasoning tasks and propose a novel framework, Meta Chain-of-Thought (Meta-CoT), to address these shortcomings. We argue that traditional Chain-of-Thought (CoT) methods, while effective for simpler problems, fail to capture the true data-generating process of complex reasoning which often involves a non-linear, iterative, and latent process of exploration and verification. Meta-CoT extends CoT by explicitly modeling this latent ``thinking'' process, which we hypothesize is essential for solving problems that require advanced reasoning capabilities.

We draw inspiration from Cognitive Science's dual-process theory, framing Meta-CoT as a form of System 2 reasoning. We establish the theoretical foundations of Meta-CoT, demonstrating how it can be realized through systematic search processes, and how these processes can be internalized within a single auto-regressive model. We then present empirical evidence supporting our claims, including analyses on state-of-the-art models like OpenAI's o1~\citep{openai2024_o1} and DeepSeek-R1~\citep{deepseek2024r1}, which exhibit behaviors consistent with internalized (in-context) search. We further explore methods for training models on Meta-CoT through process supervision, and synthetic data generation via search algorithms like Monte Carlo Tree Search (MCTS) and A*.

Finally, we outline a concrete pipeline for achieving Meta-CoT in a single end-to-end system, incorporating instruction tuning with linearized search traces and reinforcement learning (RL) post-training. We discuss open research questions, including the scaling laws of reasoning and search, the role of verifiers, and the potential for discovering novel reasoning algorithms through meta-RL. We also present the ``Big MATH'' project, an effort to aggregate over 1,000,000 high-quality, verifiable math problems to facilitate further research in this area. Our work provides both theoretical insights and a practical road map to enable Meta-CoT in LLMs, paving the way for more powerful and human-like reasoning in artificial intelligence.




\section{Meta Chain-Of-Thought}
\label{sec:meta_cot} 

In this section, we first formulate the meta chain-of-thought process and discuss how it can describe the problem solving process for complex reasoning problems. Then, we describe and demonstrate why classical chain-of-thought fails under certain circumstances.

\subsection{Deriving The Meta-CoT Process}
\label{sec:deriving_meta_cot}
A question to ask ourselves is: Should language models with Chain-Of-Thought prompting really be able to express any function, and thus solve arbitrarily complex problems, which was the theoretical point of the previous section? We will stick with the mathematical reasoning domain for the purpose of the discussion. Today, the capabilities of frontier models are enough for a large class of mathematical reasoning problems. Current state-of-the art systems such as GPT-4o and Claude largely solve the Hendrycks MATH Levels 1-3 Benchmark~\citep{hendrycks2021measuringmathematicalproblemsolving}, however, they still struggle with advanced problems such as those in Levels 4 and 5, HARP~\citep{yue2024harp} and Omni-MATH~\citep{gao2024omnimathuniversalolympiadlevel} (as well as other advanced reasoning tasks). We put forward the following theory to explain these empirical observations. 
\begin{tcolorbox}[colback=blue!5!white,colframe=blue!75!black]
Reasoning data present in pre-training corpuses does not represent the true data generation process, especially for complex problems, which is a product of extensive latent reasoning. Moreover, this process generally does \textit{not} occur in a left-to-right, auto-regressive, fashion.
\end{tcolorbox}
In more details, the CoT reasoning data prevalent in the pre-training corpus and post-training instruction tuning follows the \textbf{true data-generating process} for solutions of simple problems such as algebraic computations, counting, basic geometry etc.. That is, for example, the textbook solutions for solving high-school algebra present the general process of generating those solutions. If we follow some set of steps or approaches present in existing textbooks, we can eventually arrive at the solution. Hence, these are learnable with a constant-depth transformers that can express the complexity of each individual step in the process.  In contrast, complex reasoning problems \textit{do not} follow that pattern. We may have a set of triples $(\mathbf{q},\mathbf{S},\mathbf{a})$ of questions $\mathbf{q}$, solution steps $\mathbf{S}=(\mathbf{s}_1, \ldots, \mathbf{s}_n)$ and (optionally) answers $\mathbf{a}$, \textbf{but the true data generation process is not auto-regressive}:

\begin{equation}\label{eq:data_generative_eq}
\mathbf{q}\to\mathbf{z}_1\to\ldots\to\mathbf{z}_K\to(\mathbf{s_1}, \ldots, \mathbf{s}_n, \mathbf{a}), 
\end{equation}
where $\mathbf{z}_i$ are the latent "thoughts" left out of the solutions steps, which can be represented fully with left-to-right generation, while the dataset solution $\mathbf{S}=(\mathbf{s}_1, \ldots, \mathbf{s}_n)$ is generated jointly. Let us illustrate this with an example from the International Mathematics Olympiad 2011. This is the (in)famous "windmill" problem:
\begin{tcolorbox}[colframe=black, colback=white, boxrule=1pt, arc=0pt]
``Let $\mathcal{S}$ be a finite set of at least two points in the plane. Assume that no three points of $\mathcal{S}$ are collinear. A windmill is a process that starts with a line $\ell$ going through a single point $P \in \mathcal{S}$. The line rotates clockwise about the pivot $P$ until the first time that the line meets some other point belonging to $\mathcal{S}$. This point, $Q$, takes over as the new pivot, and the line now rotates clockwise about $Q$, until it next meets a point of $\mathcal{S}$. This process continues indefinitely. Can we choose a point $P$ in $\mathcal{S}$ and a line $\ell$ going through $P$ such that the resulting windmill uses each point of $\mathcal{S}$ as a pivot infinitely many times.''
\end{tcolorbox}

\noindent which has the following solution:

\BeginSol
``Let $|S|=n$. Now consider an arbitrary point $P$ in $S$ and choose a line $l$ through $P$ which splits the points in the plane into roughly equal chunks. Next notice that as the line rotates it will sweep a full $2\pi$ angle against some fixed reference frame. Now take another random point $P'$ and similarly constructed stationary line $l'$. At some point in the windmill process we will have $l||l'$. However notice that $l$ and $l'$ split the points into the same two sets and are parallel. Therefore we must have that $l\equiv l'$ and thus $l$ passes through $P'$. This of course holds for any $P'\in S$. Applying the same argument recursively yields the final proof that it is in fact possible to make such a construction for any set $S$ with these properties.''
\EndSol

The solution above does not use any prior knowledge and fits within a few sentences. Yet, this problem was considered among the most difficult in the competition (there were only a handful of solutions among the 600+ participants). What makes the problem difficult is the highly non-linear structure of the solution.
Most participants would follow the standard "generative" solution process and investigate approaches based on the convex hull construction or use tools from Hamiltonian graph theory, none of these leading to a solution. Instead, participants who solved the problem followed an experimental approach with a lot of geometric exploration and inductive reasoning. 
Moreover, the solution itself is not linear. It's hard to see the utility of the proposed construction in the beginning without the analysis of the dynamics of $l$. Essentially, \textbf{to start generating the solution requires that we already know the full approach}. The underlying generative process of the solution is not auto-regressive from left-to-right. 

We can formalize this argument through the interpretation of reasoning as a latent variable process~\citep{phan2023trainingchainofthoughtlatentvariableinference}. In particular, classical CoT can be viewed as

$$p_{\text{data}}(\mathbf{a}|\mathbf{q}) \propto \int \underbrace{p_{\text{data}}(\mathbf{a}|\mathbf{s}_1, \ldots, \mathbf{s}_n, \mathbf{q})}_{\text{Answer Generation}} \underbrace{\prod_{t=1}^n p_{\text{data}}(\mathbf{s}_t | \mathbf{s}_{< t}, \mathbf{q})}_{\text{CoT}} d\mathbf{S},$$

i.e., the probability of the final answer being produced by a marginalization over latent reasoning chains. We claim that for complex problems, the true solution generating process should be viewed as

$$p_{\text{data}}(\mathbf{a}, \mathbf{s}_1, \ldots, \mathbf{s}_n|\mathbf{q}) \propto \int \underbrace{p_{\text{data}}(\mathbf{a}, \mathbf{s}_1, \ldots, \mathbf{s}_n|\mathbf{z}_1, \ldots, \mathbf{z}_k, \mathbf{q})}_{\text{Joint Answer+CoT}} \underbrace{\prod_{t=1}^K p_{\text{data}}(\mathbf{z}_t | \mathbf{z}_{< t},\mathbf{q})}_{\text{Meta-CoT}} d\mathbf{Z},$$

i.e., the joint probability distribution of the solution $(\mathbf{a}, \mathbf{s}_1, \ldots, \mathbf{s}_n)$ is conditioned on the latent generative process. Notice that this argument is a meta-generalization of the prior CoT argument, hence why we will refer to the process $\mathbf{q}\to\mathbf{z}_1\to\ldots\to\mathbf{z}_K$ as \textbf{Meta-CoT}.

\subsection{Why Does (Classical) CoT Fail?}
\begin{figure}[t]
    \centering
    \includegraphics[width=0.975\linewidth]{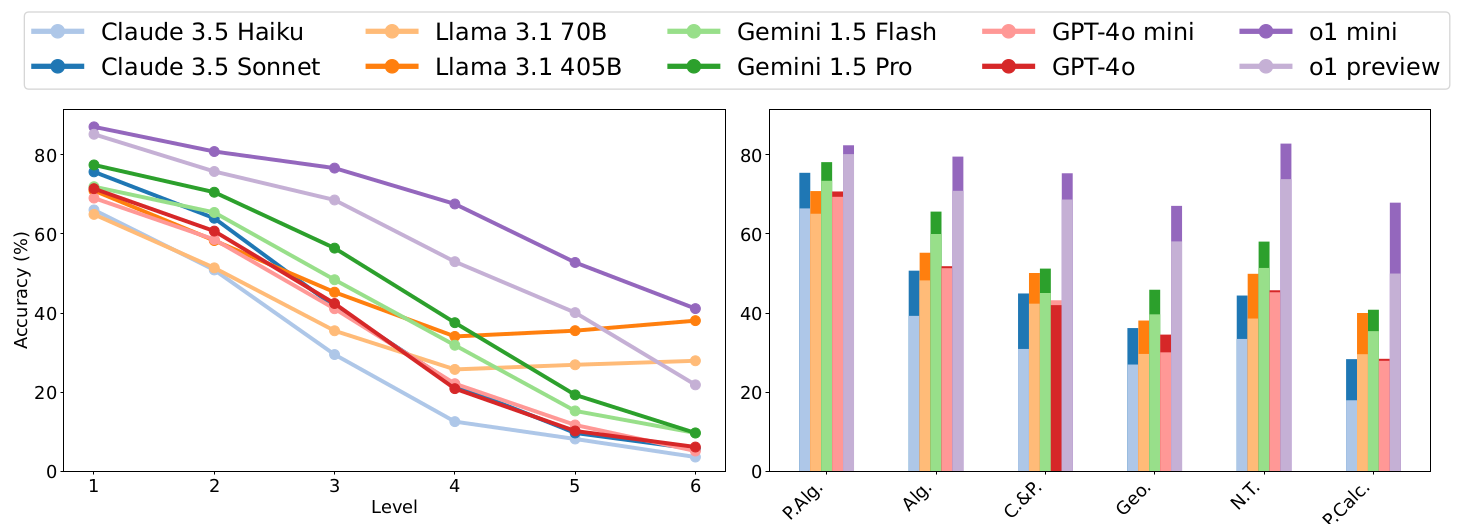}
    \includegraphics[width=0.975\linewidth]{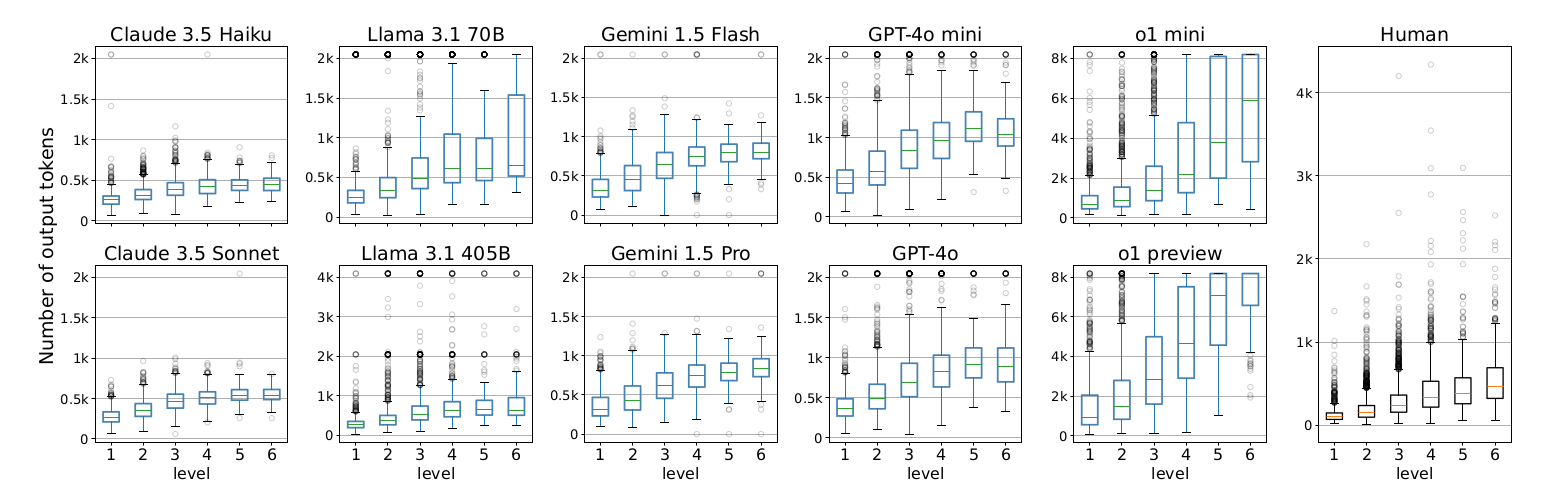}
    \caption{\textbf{Top:} Performance of current frontier models by size on the HARP mathematics benchmark~\citep{yue2024harp} by difficulty level and topic. The OpenAI O1 series significantly out-performs prior generation models across the board. Source: Figure 3 in~\citep{yue2024harp}. \textbf{Bottom} Average number of tokens generated by each model grouped by difficulty level, as well as average number of tokens in human-generated solutions (using GPT4 tokenizer). Source: Figure 4 in~\citep{yue2024harp}.}
    \label{fig:hardmath_solutions}
\end{figure}
Based on the previous discussion, a natural question follows: Why do LLMs fail at these advanced reasoning tasks? Above we proposed that the pre-training and instruction-tuning corpora consist of data of the type $(\mathbf{q}, \mathbf{s}_1, \ldots, \mathbf{s_n, \mathbf{a}})$, which do not contain the true data generating process as shown in Equation~\ref{eq:data_generative_eq}. Indeed, the solution to the windmill problem above is widely available on the internet, but there is little to no discussion about the ways in which commonly used convex hull or planar graph arguments fail. This is true in general - \textit{textbooks contain advanced proofs but not the full thought process of deriving these proofs}. We can then apply the same general meta-argument of why CoT is necessary to the Meta-CoT case: simply because the conditional solution-level distribution $p_{\text{data}}(\mathbf{a}, \mathbf{s}_1, \ldots, \mathbf{s}_n|\mathbf{q})$ (without the intermediate Meta-CoT) on hard reasoning questions can have arbitrarily high complexity in the same way that $p_{\text{data}}(\mathbf{a}|\mathbf{q})$ can have arbitrarily high complexity in the standard CoT setting. We will examine some empirical evidence for our stance in the following sections.

We will argue in the following chapters that the OpenAI o1 model series performs full Meta-CoT reasoning in an auto-regressive fashion at inference time. A useful analysis is presented in a new mathematics benchmark with challenging high-school Olympiad-level problems~\citep{yue2024harp}. Figure~\ref{fig:hardmath_solutions} sourced from that work shows the relevant results. First, we see that the o1 family of models significantly outperforms ``standard'' reasoning models across the board. However, the gap between o1 and other models' performance increases on higher difficulty problems (with the interesting exception of the LLaMa 3.1 models), that is, problems which have higher solution complexity.

Furthermore, the bottom half of Figure~\ref{fig:hardmath_solutions} shows the average number of tokens generated grouped by problem difficulty level. First, we see that outside of the o1 series of models, LLMs generate solutions of comparable lengths to humans. While this may initially appear quite intriguing, suggesting that models are learning to approximate or replicate human reasoning, the simple explanation is that models are learning solutions to match the training data - i.e.  $p_{\text{data}}(\mathbf{a}, \mathbf{s}_1, \ldots, \mathbf{s}_n|\mathbf{q})$. Much more intriguingly, the o1 series of models exhibits significantly different token behavior. We see that:

\begin{enumerate}
    \item On level 1 problems the o1 series generates a comparable number of tokens to human-written solutions. These are the types of problems where the training solutions likely match the true data generation process and each individual logical step can be internalized in a constant-depth transformer.
    \item At higher difficulty, the o1 series of models generates significantly more tokens per problem and also widens the performance gap over the classical reasoning models. In fact the gap between the inference compute used by the o1 model and prior series of models seems to \textbf{scale with the complexity of the problems}. We hypothesize that in those more challenging problems the solutions \textbf{do NOT} in fact represent the true data generative process, which is instead better approximated by the more extensive Meta-CoT generated by the o1 family of models.
\end{enumerate}

Of course, in practice the distinction between these two is not so clear cut, and in fact the constant-depth transformer can likely internalize part of the Meta-CoT generative process as evidenced by the gradation of (Meta-)CoT lengths from Levels 2-6 in Figure~\ref{fig:hardmath_solutions}. In the next chapter we will discuss in greater detail what the Meta-CoT process actually represents. 

\section{Towards Deliberate Reasoning With Language Models - Search}\label{sec:search}

In the previous section we introduced the Meta-CoT process and argued that LLMs fail on advanced reasoning tasks because the \textbf{training data does not adequately represent the true data generation process}, i.e. text corpora do not include (or only include limited amounts of) Meta-CoT data. So the remaining question is: what does the true data generating process look like? 

\begin{enumerate}
    \item First, we argue that for many advanced reasoning or goal-oriented problems there exist meaningful gaps between the complexity of generation and verification. This is of course one of the fundamental open problems of theoretical computer science and any attempt to prove this is significantly beyond the scope of the current writing, but we will review what we believe to be compelling empirical evidence from the literature. 
    \item Second, assuming a non-trivial generator-verifier gap, we argue that the solutions to challenging problems present in text corpora are the outcomes of an \textbf{extended search process}, which itself is not represented in the data.
\end{enumerate}
\noindent We will dive deeper into these two points in the remainder of this section.
\subsection{Inference-Time Compute: Search}\label{sec:best-of-N}

The first point above (generation-verification gap) has recently become a popular research and discussion direction under the framework of ``deploying inference-time compute'' and we explore this in our first experiment. We start with a LLaMa 3.1 8B base model~\citep{dubey2024llama3herdmodels} and carry out extensive supervised fine-tuning on the Numina MATH dataset~\citep{numina_math_datasets}. Refer to Figure~\ref{fig:sft_scaling} for results and Section~\ref{sec:dataset} for dataset details. For each intermediate checkpoint we evaluate performance on the Hendrycks MATH~\citep{hendrycks2021measuringmathematicalproblemsolving} 500 problems evaluation dataset~\citep{lightman2023let}. Based on the results, we make a few observations here:

\begin{figure}[t]
    \centering
\begin{minipage}{.5\textwidth}
    \centering
    \includegraphics[width=\textwidth]{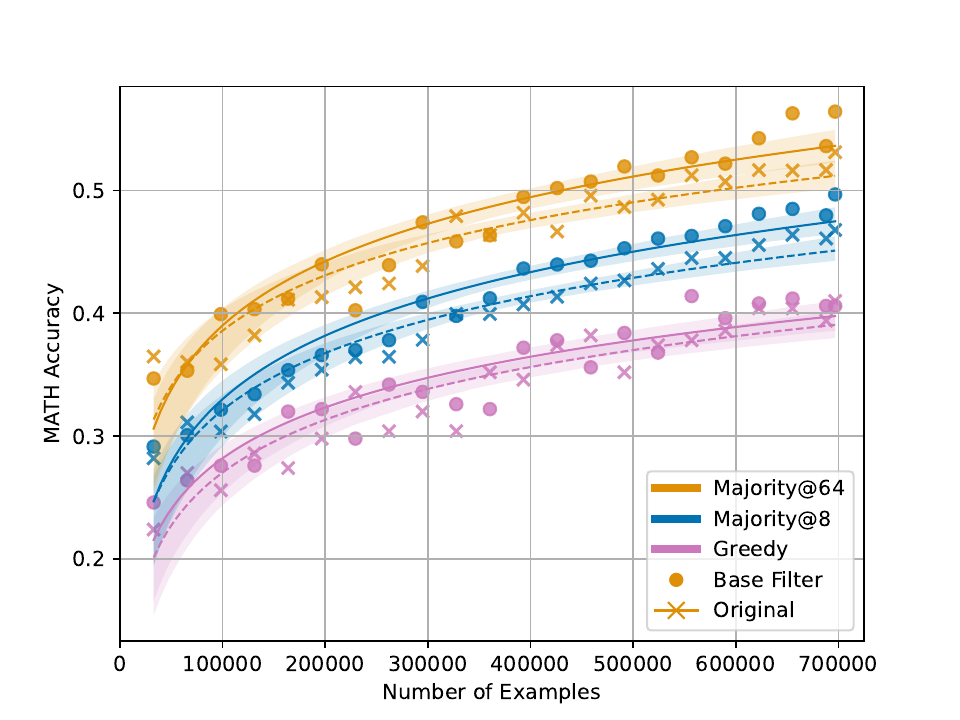}
\end{minipage}%
\begin{minipage}{.5\textwidth}
    \centering
    \includegraphics[width=\textwidth]{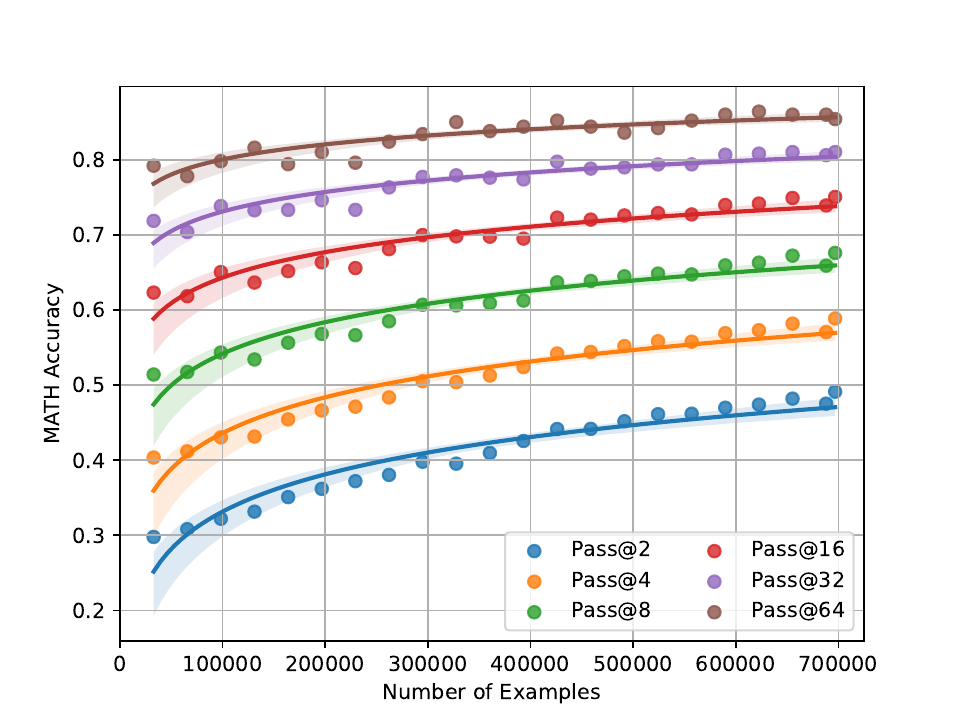}
\end{minipage}
\caption{\textbf{Train-time scaling curves} for increasing quantities of training data during supervised fine-tuning of Llama3.1 8B, when evaluating for accuracy on the MATH test set. \textbf{Left}: A comparison of training on on the original data (x) and base filtered data (o) and evaluated using either greedy or majority vote decoding. For all settings, the filtered dataset shows improved scaling, and has not plateaued.
\textbf{Right}: A comparison of pass@k for increasing $k$ with a model trained on the base filtered dataset. The chart shows how increasing $k$ leads to a much greater probability of at least 1 correct solution, even for a relatively small model. Additionally, the performance improvement from $k=32$ to 64 does not show plateauing, suggesting that $k$ can further be increased for improved performance.}
\label{fig:sft_scaling}
\end{figure}

\begin{enumerate}
    \item We evaluate pass@$k$ (i.e.\ using an oracle verifier) on intermediate checkpoints and see a significant jump in performance for increasing $k$. While zero-shot performance with greedy decoding improves from about 20\% to 40\% (see the base filter on the left side of Figure~\ref{fig:sft_scaling}), even the first model checkpoint outperforms these results at pass@$4$ (right side of Figure~\ref{fig:sft_scaling}). Moreover, the pass@64 for the final checkpoint of an 8B model achieves accuracy close to 85\%, outperforming the zero-shot performance of many current frontier models.
    \item We also evaluate performance under majority voting with $k=8$ and $k=64$. There is continuous improvement for both increased training and samples, with maj@64 outperforming the greedy model performance \textbf{with only 15\% of the training compute} without access to a ground-truth verifier.
\end{enumerate}

These results demonstrate that even as we directly optimize for answer generation ability by finetuning on increasing amounts of SFT data, there remains a consistent verifier-generator gap, as evidenced by the improved performance in botht eh pass@k and majority voting settings. Recent literature has observed similar results on post-training sampling~\citep{lightman2023let, brown2024large, snell2024scalingllmtesttimecompute}. However, most of these studies do not systematically evaluate the effects of varying amounts of training data, compute, and model size which we believe is a fruitful direction for additional empirical work. These questions are important as the observed gains from additional inference might disappear at larger scales and training - i.e. the model may be able to fully internalize the reasoning process. This definitely seems to be the case for advanced models and simpler benchmarks like GSM8k~\citep{cobbe2021trainingverifierssolvemath}. While we observe the opposite result in our experiments, we admit that our results are the outcomes of preliminary study and additional work is required, but we will argue from a theoretical point in Section~\ref{sec:meta_RL} that \textit{a persistent search gap remains in domains with high enough epistemic uncertainty}.
Besides this point, the question remains whether the improvement from increased inference can be effectively achieved without oracle verifiers or environment feedback. In theory, it is possible to generate correct solutions under an increased inference budget, but we may not be able to verify them effectively, as verification complexity may be just as high as, or even higher than, generation complexity. We will address this issue next.

\subsection{Inference-Time Compute: Verification}
\label{sec:verifiers}

\begin{figure}
    \centering
    \includegraphics[width=0.95\linewidth]{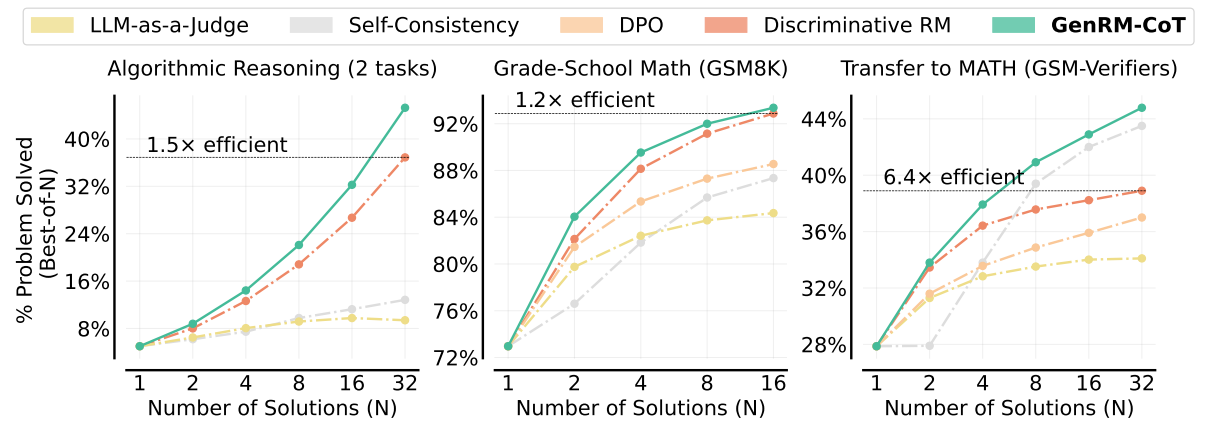}
    \caption{\textbf{Scaling trends for verifier models} on algorithmic reasoning, grade-school math (GSM8k), and transfer from GSM8k to MATH. The performance of all verifiers improves in the best-of-N setting, as N increases. Figure sourced from~\citep{zhang2024generativeverifiersrewardmodeling}.}
    \label{fig:verfier_scaling}
\end{figure}


Several works focus on \textbf{training verifier models}, which explicitly evaluate the correctness of reasoning steps and solutions. Verifiers can be trained either using explicit binary classification~\citep{cobbe2021trainingverifierssolvemath,lightman2023let, snell2024scalingllmtesttimecompute, anonymous2024improving, setlur2024rewarding} or modeling evaluation directly in natural language, using the LLM-as-a-judge prior~\citep{zhang2024generativeverifiersrewardmodeling, mahan2024generative}. The unifying formulation of these approaches is the model $v_{\theta}$ which evaluates a reasoning process $v_{\theta}(\mathbf{q}, \mathbf{S})\to[0, 1]$. 
Under this framework, $K$ candidate solutions $(\mathbf{S}^1, \ldots, \mathbf{S}^K)$ can be generated \textbf{from a fixed generator} $\pi_{\theta}(\cdot|\mathbf{q})$ and ranked based on their evaluation score. 
$$\mathbf{S}^*=\arg\max \{v_{\theta}(\mathbf{q}, \mathbf{S}^1), \ldots, v_{\theta}(\mathbf{q}, \mathbf{S}^K)\}$$
For empirical results, we refer the reader to Figure~\ref{fig:verfier_scaling} sourced from~\citep{zhang2024generativeverifiersrewardmodeling} which evaluates a number of verifier models $v_{\theta}$. Regardless of the efficiency of the verifier, there is a significant improvement in performance with additional online sampling. Moreover using explicitly trained verifier models outperforms naive inference-compute scaling strategies such as self-consistency or majority voting. 

A question remains regarding the effect of using a fixed generation model (policy): Could this model be under-trained, and if it were further trained, could its zero-shot performance improve to the point where additional online search no longer provides meaningful improvement? We will address this in Section~\ref{sec:search_scaling}.

\subsection{From Best-of-N To General Search}\label{sec:general_search}

\begin{figure}
    \centering
    \includegraphics[width=0.95\linewidth]{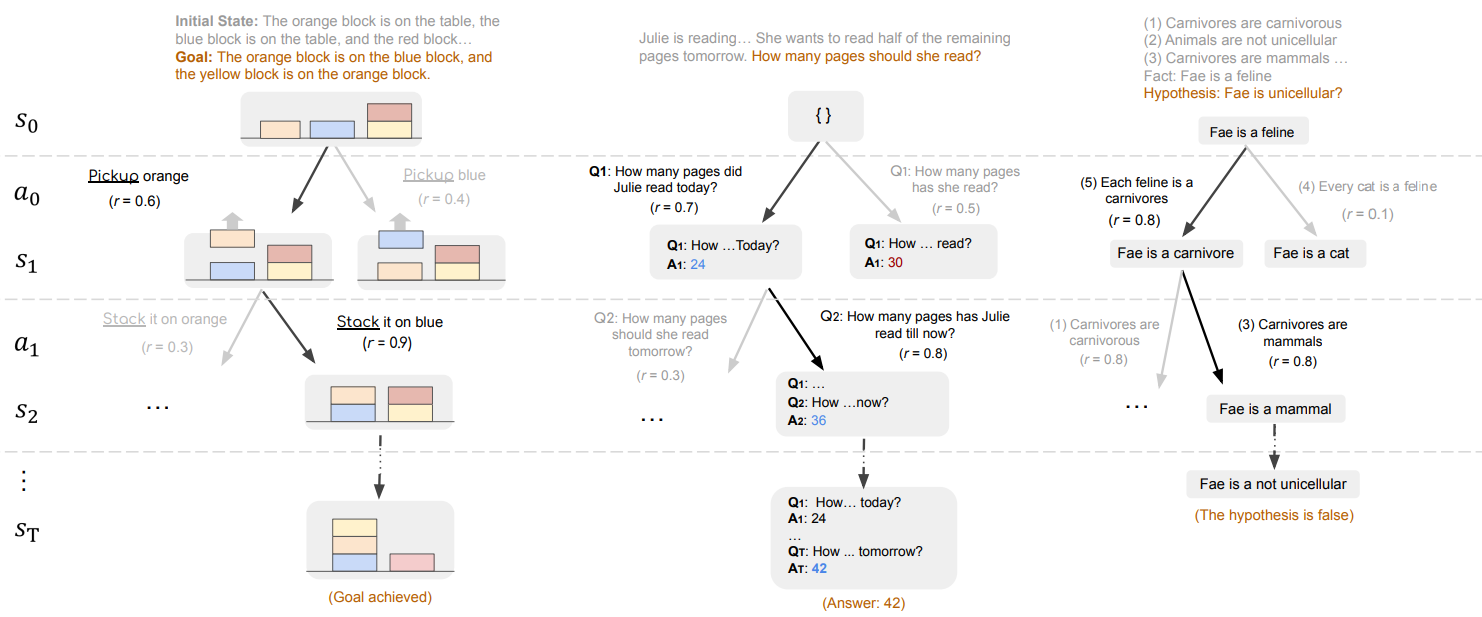}
    \caption{Reasoning via Planning (RAP) demonstrates the search procedure described here. If we have access to a state evaluator, we can truncate branches with low values and backtrack to promising nodes, without resampling the same steps again. Source: Figure 2 in~\citep{hao2023reasoninglanguagemodelplanning}.}
    \label{fig:RAP}
\end{figure}

So far, we empirically explored best-of-N approaches, generating multiple full solutions independently and selecting the most promising one based on scores. However, this approach is inefficient because it requires exploring full solution paths, even if a mistake occurs early on, and may repeatedly sample the same correct steps. Instead, we can model reasoning as a Markov Decision Process (MDP), defined by the tuple \( \mathcal{M} = (\mathcal{S}, \mathcal{A}, P, R, \gamma) \), where:

\begin{itemize}
    \item \( \mathcal{S} \): the set of states, where each state \( \mathbf{S} \in \mathcal{S} \), consists of the prompt and generations so far, i.e. $\mathbf{S}_t = (\mathbf{q}, \mathbf{s}_1, \ldots, \mathbf{s}_t)$.
    \item \( \mathcal{A} \): the set of actions, where each action \( \mathbf{a} \in \mathcal{A} \) will be represented as the next reasoning step $\mathbf{a}_{t+1}=\mathbf{s}_{t+1}$. 
    \item \( P(\mathbf{s}' \mid \mathbf{s}, \mathbf{a}) \): the transition probability function, representing the probability of transitioning to state \( \mathbf{s}' \) when taking action \( \mathbf{a} \) in state \( \mathbf{s} \). For simplicity, we will mostly consider the deterministic transition function $P(\cdot|\mathbf{s}_{t+1}, (\mathbf{q}, \mathbf{s}_1, \ldots, \mathbf{s}_t))\to (\mathbf{q}, \mathbf{s}_1, \ldots, \mathbf{s}_t, \mathbf{s}_{t+1})$ that appends the next reasoning step to the context. In general, the environment dynamics can be more complex. For example, models with tool access have to call the actual tool and receive the environment feedback in context or even modify their environment such as the cases of SWE and Web agents.
    \item \( R(\mathbf{s}, \mathbf{a}) \): the reward function, which provides a scalar reward for taking action \( \mathbf{a} \) in state \( \mathbf{s} \).  We will assume zero intermediate rewards and final reward of 1 for a correct solution and zero otherwise, although this is not strictly necessary in the presence of a good process reward model~\citep{setlur2024rewardingprogressscalingautomated}.
    \item \( \gamma \in [0, 1] \): the discount factor, balancing the trade-off between further computation and rewards.
\end{itemize}

We refer to the LLM generating the reasoning steps as the policy $\mathbf{s}_{t+1}\sim\pi_{\theta}(\cdot|\mathbf{S}_t)$. In addition we refer to a solution starting from $\mathbf{s}_0=\mathbf{q}$ as an episode or a trajectory. We will also use the notation $\mathbf{z}_t$ to represent individual reasoning steps that are part of the Meta-CoT and correspondingly denote $\mathbf{Z}_t=(\mathbf{q}, \mathbf{z}_1, \ldots, \mathbf{z}_t)$.

In the prior section we considered generating and ranking full solutions, which may be inefficient. We can extend the concept of a solution verifier from the prior section, to estimating the probability that a particular intermediate state will lead to a solution: $v_{\theta}(\mathbf{q}, \mathbf{S}_t) \to [0, 1]$. These models have become more widely known as Process Reward Models (PRMs)~\citep{lightman2023let}. If we have access to such a model, we can improve the efficiency of the search process with the following steps: 

\begin{enumerate}
    \item Terminate a solution attempt that is not making progress, or is incorrect prior to reaching the final answer.
    \item Reset the agent to any intermediate, previously visited, state that has a high likelihood of success.
\end{enumerate}

Notice that with these two operations, and the general structure of language, we can implement any tree search procedure. This is the premise of several approaches~\citep{yao2023treethoughtsdeliberateproblem, hao2023reasoninglanguagemodelplanning, zhou2024languageagenttreesearch} with the RAP method~\citep{hao2023reasoninglanguagemodelplanning} illustrated in Figure~\ref{fig:RAP}.
\begin{wrapfigure}{r}{0.45\textwidth}
\vspace{-10pt}
  \begin{center}
    \includegraphics[width=0.42\textwidth]{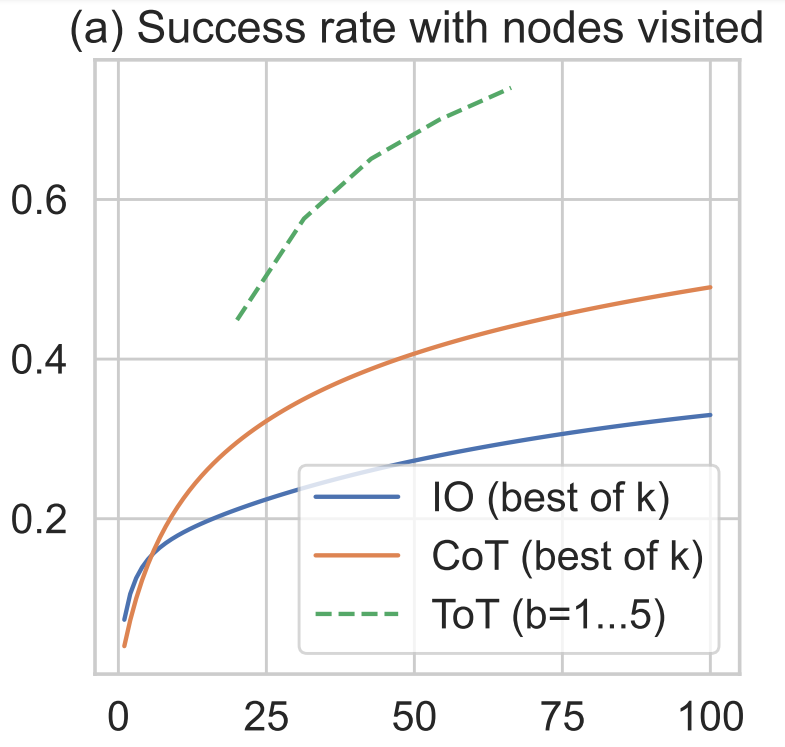}
  \end{center}
  \caption{ToT efficiency on the game of 24 shown as accuracy (y-axis) vs. \# nodes visited (x-axis). Source: Figure 3 in~\citet{yao2023treethoughtsdeliberateproblem}.}
  \label{fig:ToTEfficiency}
\end{wrapfigure}
These approaches use differing search strategies (DFS/BFS vs. MCTS) and process guidance evaluation (generative self-evaluation vs. Monte-Carlo rollouts), but they all share the same core idea: formulate the reasoning problem as tree search guided by an intermediate heuristic function. As noted above, in theory, tree search does not induce a fundamental capability shift over parallel sampling, however, it may induce significant efficiency gains as demonstrated by~\citet{yao2023treethoughtsdeliberateproblem}. In particular, Figure~\ref{fig:ToTEfficiency} (sourced from~\citet{yao2023treethoughtsdeliberateproblem}) shows nearly 4 times increased efficiency, in terms of inference budget, on a toy reasoning problem (Game of 24)  when using a tree-structured search approach compared to parallel sampling. While these earlier works focus on zero-shot (or close to zero-shot) performance on simple reasoning tasks, it is important to note that tree-search methods have been successfully scaled and deployed to a number of realistic agentic applications~\citep{koh2024tree, putta2024agent, brown2024large, yu2024exactteachingaiagents}.  

\subsection{Is Search (Inference Time Compute) A Fundamental Capability Shift?}\label{sec:search_scaling}
\begin{figure}
    \centering
    \includegraphics[width=0.490\linewidth]{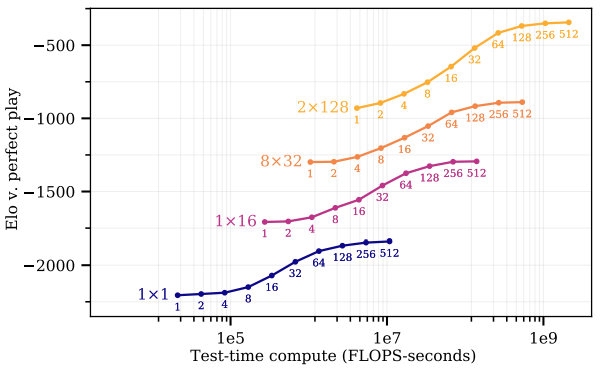}
    \includegraphics[width=0.475\linewidth]{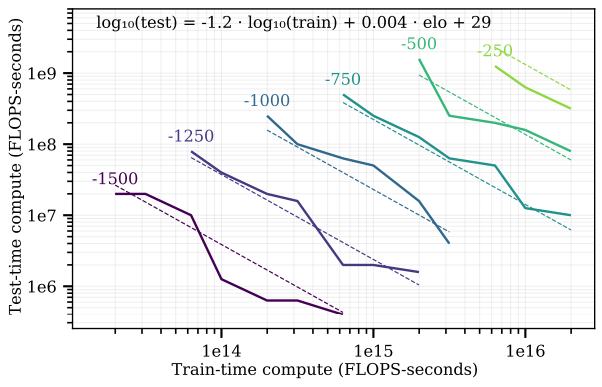}
    \caption{\textbf{Scaling trends for MCTS at training and inference-time} on board games. \textbf{Left:} Elo scores of models trained with different architectures (depth $\times$ width) where each point represents the Elo score of that model evaluated with the labeled tree size (between 1 to 512 nodes). The curves demonstrate that the performance of each model snapshot follows a sigmoid pattern with respect to the test-time compute budget. Source: Figure 8 in~\citep{jones2021scalingscalinglawsboard}. \textbf{Right:} The trade-off between train-time and test-time compute, with progressively improving Elo (from bottom-left to top-right). Source: Figure 9 in~\citep{jones2021scalingscalinglawsboard}.}
    \label{fig:board_scalingt_scaling}
\end{figure}

As pointed out earlier, the question remains whether inference-time search is a fundamental new capability or whether it is accessible with additional training. Results from classical RLHF tuning~\citep{dubois2024alpacafarm} suggest that this is a learnable capability, where zero-shot performance of post-trained models matches or outperforms the best-of-N paradigm.

\begin{tcolorbox}[colback=blue!5!white,colframe=blue!75!black]
  We stipulate that performance on complex reasoning tasks is governed by a scaling law, which involves model size, training data (compute) and inference time compute.
\end{tcolorbox}

This is indeed consistent with the theoretical results of~\citet{li2024chainthoughtempowerstransformers} and the intuition presented in Section~\ref{sec:meta_cot}.
Larger models are more capable of internalizing the Meta-CoT process in their activations, and are also capable of using longer inference-time Meta-CoT to approximate solutions with significantly higher computational complexity.
Empirically, we have limited (but promising) evidence towards this hypothesis. A major prior work to study these questions is~\citet{jones2021scalingscalinglawsboard} which carries
\begin{wraptable}{r}{0.5\textwidth}
    \centering
    \begin{tabular}{l|l|l|c}
        \hline
        \textbf{Method} & \textbf{Policy} & \textbf{Value} & \textbf{Accuracy(\%)} \\
        \hline
        \multirow{2}{*}{Greedy} & $\pi_{\theta_0}$ & - & 41.4 \\
                & $\pi_{\theta_1}$ & - & 47.9 \\
        \hline
        \multirow{4}{*}{MCTS-$\alpha$} & $\pi_{\theta_0}$ & $\{v, \hat{r}\}_{\phi_0}$ & 51.9 \\
                      & $\pi_{\theta_0}$ & $\{v, \hat{r}\}_{\phi_1}$ & 53.2 \\
                      & $\pi_{\theta_1}$ & $\{v, \hat{r}\}_{\phi_0}$ & 54.1 \\
                      & $\pi_{\theta_1}$ & $\{v, \hat{r}\}_{\phi_1}$ & \textbf{56.5} \\
        \hline
    \end{tabular}
    \caption{Iterative update results on GSM8k. $\theta_0, \phi_0$ are the old parameters while $\theta_1, \phi_1$ are the new ones. TS-LLM can boost performance by training LLM policy, value, or both. Source: Table 4 in~\citep{feng2024alphazerolike}.}
    \label{tab:iterative_update_MCTS}
\end{wraptable}
out studies using the AlphaZero algorithm~\citep{Silver2018AlphaZero} on board games. This approach fits our desiderata very well as the underlying MCTS algorithm jointly scales the policy and value (verifier) models' training in conjunction with search. Moreover, this family of board games have a clear generator-verifier gap as generating optimal strategies at intermediate steps can be quite computationally complex, while verifying a winning condition is trivial. The major empirical results on scaling are shown in Figure~\ref{fig:board_scalingt_scaling}. On the right side we see that performance increases both with increased training compute and model size, as outlined earlier. Interestingly, on the left we see the performance of using different quantities of compute (i.e.,\ search with a value function) during inference. There is also a clear scaling trend, showing improved performance with additional online search \textbf{at each intermediate checkpoint of training}. In fact, the results in this domain indicate there is a clear log-log scaling trade-off between train-time and test-time compute deployment. Currently, we have limited evidence of similar scaling laws in LLMs because such a training pipeline requires significant resources. One major work towards that goal is~\citet{feng2024alphazerolike} which carries out two iterations of MCTS fine-tuning using a LLaMa 7B on the GSM8k dataset~\citep{cobbe2021trainingverifierssolvemath}. They show improved performance in zero-shot evaluations of the policy, as well as significant gains from using additional inference-time search, at both iterations 1 and 2 (full results are shown in Table~\ref{tab:iterative_update_MCTS}). However, their work does not ablate the model size, data scaling, or inference-time search scaling, which remain under-explored in the literature for LLM reasoning.

\section{Towards Meta-CoT Reasoning}\label{sec:internal_reasoning}
In prior sections we: introduced the concept of Meta-CoT and argued that it is necessary for advanced reasoning, discussed the generator-verifier gap as a fundamental limitation, argued for search as a fundamental building block of the Meta-CoT, and discussed the utility of approaches integrating generator, verifier, and search components. \textit{However, the question remains on how to integrate these into a model to perform Meta-CoT or ``System 2'' reasoning.} The first question we need to answer is: why do we actually need to internalize deliberate reasoning inside a single model? We propose two main reasons:

\begin{enumerate}
    \item \textbf{Efficiency:} By incorporating search within the context of an auto-regressive model, exploration can be done efficiently since the model has access to all previously visited nodes, in context. Unique to the case of reasoning in natural language, many branches may contain semantically similar content, unlike other domains (e.g.,\ board games), motivating the need for improved efficiency. In fact, even advanced reasoning models carry out many repeated steps of semantically identical reasoning as we show in Figure~\ref{fig:backtrack_examples_1} and \ref{fig:backtrack_examples_2}.

    \item \textbf{Super-Intelligence:} If an auto-regressive model can learn to implement search algorithms in-context, then additional RL training may \textbf{enable the model to discover novel reasoning approaches}. Essentially, we propose that training a model capable of internal System 2 reasoning (e.g.\ Meta-CoT) and search is an optimization over algorithms rather than specific outputs, possibly yielding novel modes of problem solving. This will potentially allow the model to solve classes of problems previously unsolvable under symbolic-bases tree-search approaches as we've outlined in Sections~\ref{sec:general_search} and~\ref{sec:search_scaling}.
\end{enumerate}


In the remainder of this section, we explore how to train a model to internalize such a reasoning system.

\subsection{Bootstrapping Meta-CoT}
In this subsection, we overview the core idea behind the Self-Taught Reasoner (STaR) approach~\citep{zelikman2022star,singh2024humandatascalingselftraining,yuan2023scalingrelationshiplearningmathematical} to bootstrapping intermediate CoT steps and how to use a similar concept to generalize to meta-reasoning strategies.

\subsubsection{Self-Taught Reasoner}
The STaR method introduces an iterative bootstrapping approach designed to improve the reasoning capability of LLMs~\citep{zelikman2022star}. STaR focuses on training models to generate and refine rationales, particularly for tasks requiring complex reasoning in a reinforcement learning-based manner. In this formulation we assume we have access to a dataset $\mathcal{D}=\{\mathbf{q}^{(i)}, \mathbf{a}^{(i)}\}_{i=1}^N$ of questions $\mathbf{q}$ that require reasoning along with corresponding answers $\mathbf{a}$. Notice that we do not require access to ground-truth rationales for these problems. We begin by prompting a model $\hat{\mathbf{a}}^{(i)}, \hat{\mathbf{S}}^{(i)}\sim\pi(\mathbf{a},\mathbf{S}|\mathbf{q}^{(i)})$ to provide CoT rationale $\hat{\mathbf{S}}^{(i)}=\mathbf{s}_1^{(i)},\ldots,\mathbf{s}_{N_i}^{(i)}$ and final answer $\hat{\mathbf{a}}^{(i)}$. We then filter the generated data, keeping only rationales that lead to a correct final answer (i.e., $\hat{\mathbf{a}}^{(i)}=\mathbf{a}^{(i)}$) to create a dataset of questions, (bootstrapped) rationales and answers $\mathcal{D}_{\text{STaR}}=\{\mathbf{q}^{(i)}, \hat{\mathbf{S}}^{(i)}, \mathbf{a}^{(i)}\}_{i=1}^N$. $\mathcal{D}_{\text{STaR}}$ is then used to train a model with the standard supervised fine-tuning objective:
\begin{equation}\label{eq:STaR}
   \mathcal{L}_{\mathrm{STaR}}(\pi_{\phi}) = -\mathbb{E}_{(\mathbf{q}, \hat{\mathbf{S}}, \mathbf{a}) \sim \mathcal{D}_{\text{STaR}}} \left[ -\log \pi_{\phi}(\mathbf{a}, \hat{\mathbf{S}}|\mathbf{q}) \right].
\end{equation}
The above procedure is repeated over several iterations. The core idea behind STaR is to generate a training dataset of synthetic rationales through sampling and verification. We will extend that idea to the the concept of Meta-CoT below.

\subsubsection{Meta-STaR}
We can generalize the above idea to Meta-CoT in a straightforward way. Consider a base policy $\pi_{\theta}$ combined with some \textit{general search procedure} over intermediate steps. Given a question $\mathbf{q}$ we perform the search procedure repeatedly to generate search traces $\hat{\mathbf{z}}_1, \ldots, \hat{\mathbf{z}}_K$ until we find a final solution $(\mathbf{s}_1, \ldots, \mathbf{s}_n)$. If we can verify the \textit{final produced solution} $v(\mathbf{S})\to\{0,1\}$, for example by using a formalization and verification approach (as in AlphaProof\footnote{\url{https://deepmind.google/discover/blog/ai-solves-imo-problems-at-silver-medal-level/}}) or some other outcome verification, we can then apply a similar approach to STaR. For example,  we can construct a dataset $\mathcal{D}_{\text{STaR}}=\{\mathbf{q}^{(i)}, \hat{\mathbf{Z}}^{(i)}, \hat{\mathbf{S}}^{(i)}\}_{i=1}^N$ and use a similar training objective as before:
\begin{equation}\label{eq:Meta_STaR}
   \mathcal{L}_{\mathrm{Meta-STaR}}(\pi_{\phi}) = -\mathbb{E}_{(\mathbf{q}, \hat{\mathbf{Z}}, \hat{\mathbf{S}}) \sim \mathcal{D}_{\text{STaR}}} \left[ -\log \pi_{\phi}(\hat{\mathbf{S}}, \hat{\mathbf{Z}}|\mathbf{q}) \right].
\end{equation}
Essentially, we can use a base policy and search procedure to generate synthetic search data and then train the model to implement these in-context through the Meta-CoT concept. We are effectively proposing to linearize the search approaches described in Section~\ref{sec:search} and teach an auto-regressive model to run them sequentially. 
So far we have deliberately been vague about how these search procedures and datasets look. We will now provide examples and proof of concept from the literature on practical approaches to this problem as well as synthetic examples of realistic training data.

\subsection{Empirical Examples Of Internalizing Search}
When we formulate search in a sequential fashion we can explicitly parameterize each component in language, or choose leave it implicit~\citep{gandhi2024stream}. Note that models trained with standard next token prediction still need to implicitly internalize all of these components anyway in order to accurately model the search sequence, even if they are not explicitly verbalized. However, allowing the model to vocalize it's certainty or estimated progress could allow for additional modeling capacity or be useful for interpretability purposes. We will present some examples of auto-regressive search procedures from the literature in the following section.

\subsubsection{Small-Scale Empirical Results on Internalizing Search}
\label{subsec:linear_search}
\begin{figure}
    \centering
    \includegraphics[width=0.95\linewidth]{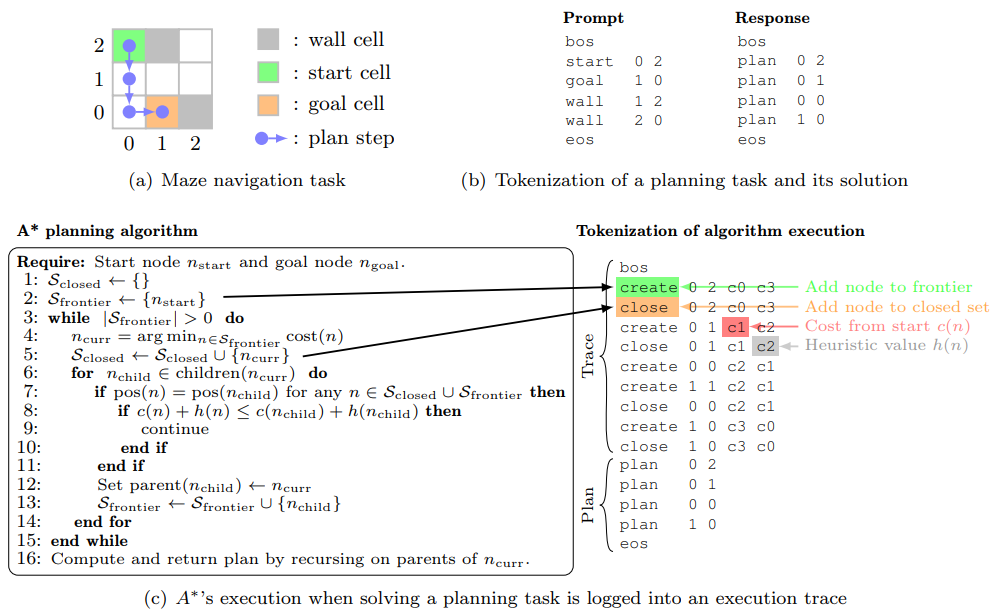}
    \caption{\textbf{A$^*$ planning algorithm outline} for a simple maze navigation task, along with a state and action tokenization scheme. The search representation explicitly models nodes and queue state, the search procedure and the cost and heuristic evaluation.  Source: Figure 1 in~\citep{lehnert2024abetterplanningtransformers}.}
    \label{fig:searchformer}
\end{figure}

\begin{figure}
    \centering
    \includegraphics[width=0.95\linewidth]{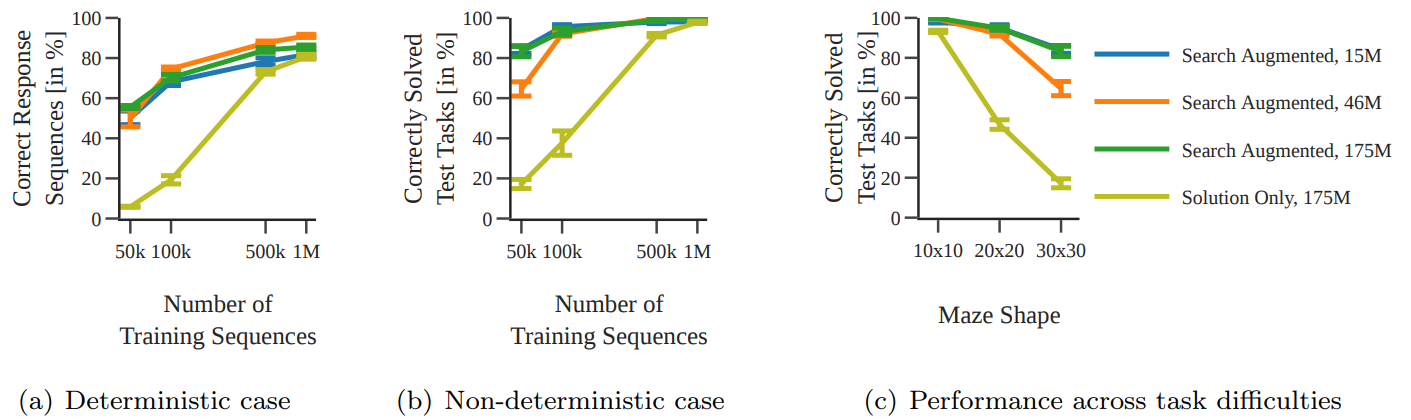}
    \caption{\textbf{Model performance vs. training compute} when using the A$^*$ planning algorithm (Search Augmented) vs. no search (Solution Only). We see that the search augmented models perform much better across all training scales (charts a and b). In particular performance is consistent with the search formulation of Section~\ref{sec:search_scaling}. Figure c) shows performance in terms of task complexity as maze size increases. Results are consistent with the Meta-CoT complexity argument presented in Section~\ref{sec:meta_cot} and results on the HARP benchmark in Figure~\ref{fig:hardmath_solutions}. Source: Figure 2 in~\citep{lehnert2024abetterplanningtransformers}.}
    \label{fig:searchformer_results}
\end{figure}

Two particular prior works that explore the idea of in-context search are~\citet{yang2022chainthoughtimitationprocedure} and~\citet{lehnert2024abetterplanningtransformers} which focus on mazes and other classical RL environments. The formulation from~\citet{lehnert2024abetterplanningtransformers} is shown in Figure~\ref{fig:searchformer}, which illustrates linearizing A* search. In our framework the ``Trace'' corresponds to the Meta-CoT $\mathbf{Z}$, and the ``Plan'' is the CoT output $\mathbf{S}$. In this setting the search procedure is stated explicitly as it shows node states, actions, costs and heuristic values. In this ``stream'' format we can then use standard auto-regressive language models with a next token-prediction objective to train a model to internalize the search process. Evaluation results are shown in Figure~\ref{fig:searchformer_results} sourced from the same paper. We observe empirical effects consistent with the scaling law hypothesis presented in~\ref{sec:search_scaling}; there is \textbf{consistent improvement with additional training data and model size (train-time compute)} across the board. A particularly interesting observation is the complexity scaling relationship in part (c) of the figure. At smaller mazes (lower complexity) the model directly producing the Plan (CoT) and performs comparably to smaller search (Meta-CoT) augmented models, however as maze size (complexity) increases we see a widening gap in performance between the search-augmented and zero-shot models. This is essentially identical to the results shown in Figure~\ref{fig:hardmath_solutions} on the challenging HARP benchmark~\citep{yue2024harp} between the prior frontier models and the o1 series. These empirical observations are well aligned with the intuition we presented in Section~\ref{sec:meta_cot}. For small mazes (low complexity problems) models are capable of internalizing the reasoning process, but as problem complexity (maze size) increases this becomes more challenging and model performance falls off compared to models which explicitly carries out a search procedure. Unfortunately,~\citet{lehnert2024abetterplanningtransformers} did not publish inference compute scaling laws, but given the algorithmic structure of the training data we can presume that inference-time tokens scale with the same complexity as the A* search algorithm, which can be exponential in the branching factor, while the plan length is linear in $n$. These results would also be consistent with the inference costs on advanced math reasoning tasks reported in Figure~\ref{fig:hardmath_solutions}.

\begin{figure}
  \centering
  \includegraphics[width=0.245\linewidth]{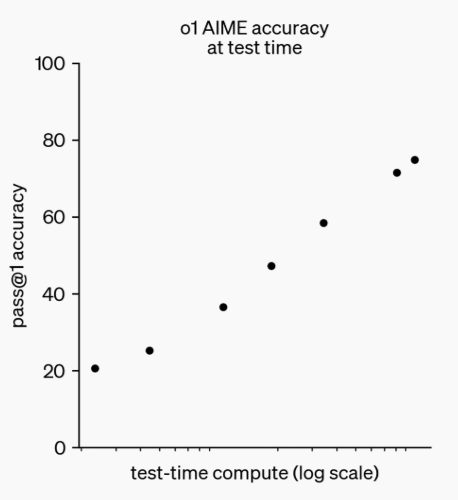}
  \includegraphics[width=0.390\linewidth]{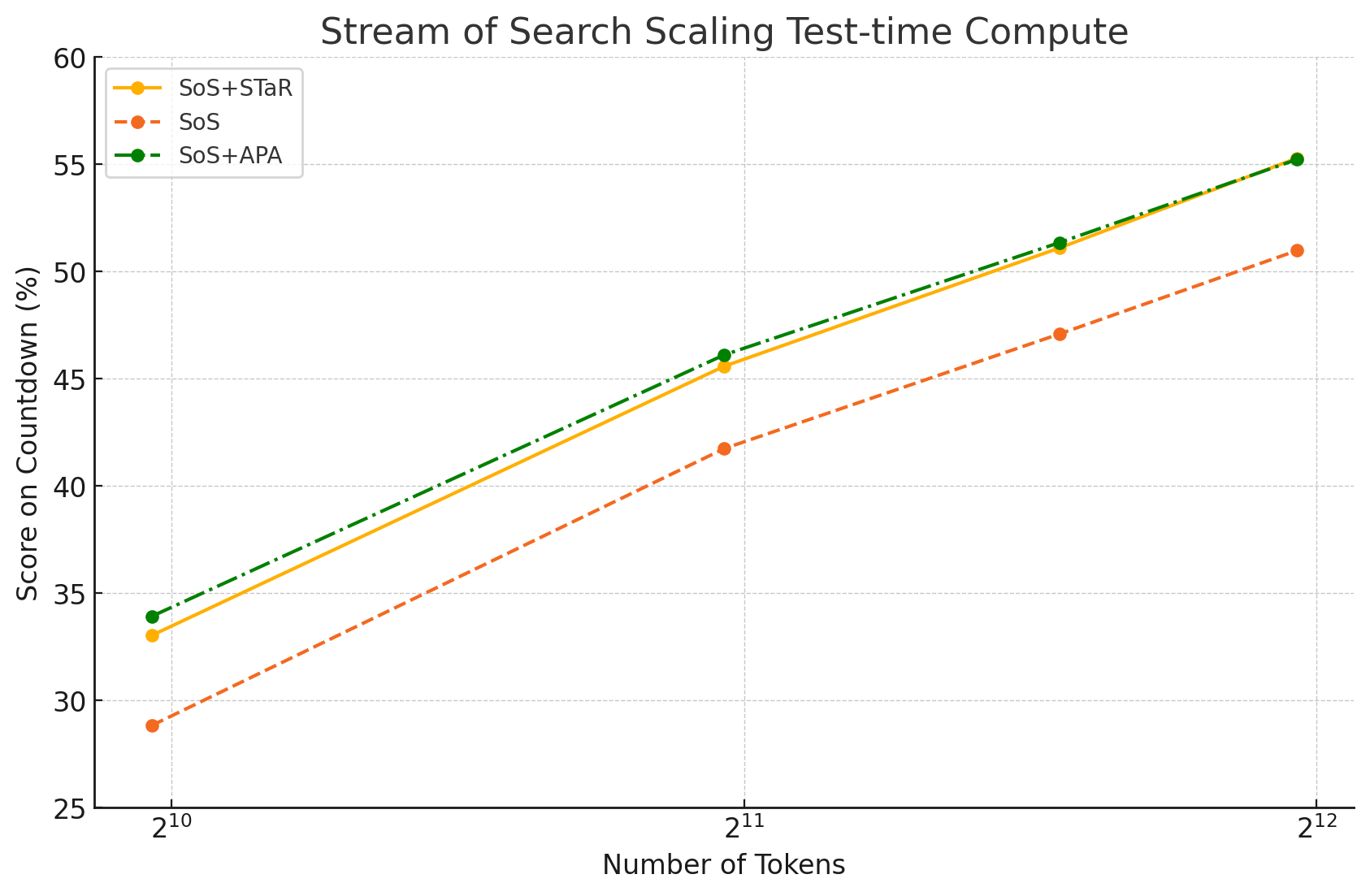}
  \includegraphics[width=0.310\linewidth]{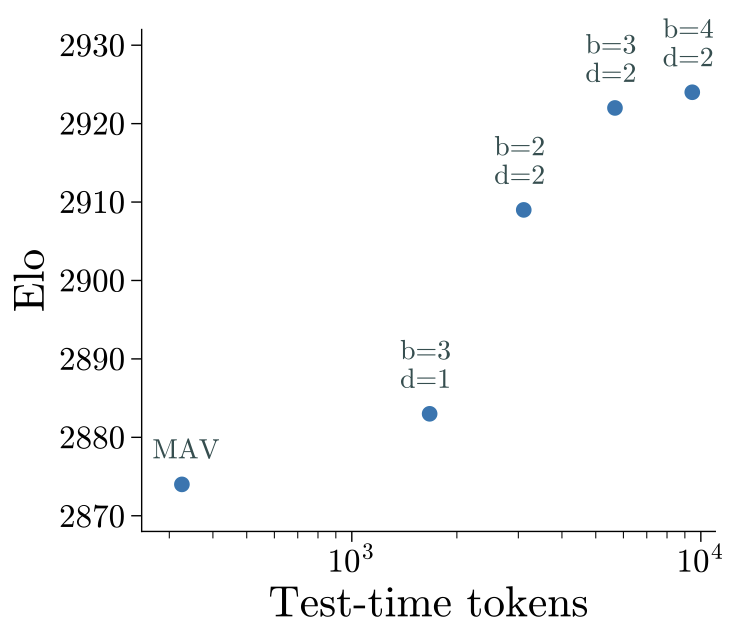}
  \caption{Inference compute scaling relationships for the o1 model (Left, sourced from~\citep{openai2024_o1} on AIME, Stream-of-Search on the Game of 24 (Middle) and MAV-MCTS on Chess (Right, sourced from~\citep{schultz2024masteringboardgamesexternal}). These figures show performance of a single model under different token sampling budgets.}
  \label{fig:SoSInferenceScalling}
\end{figure}

\citet{gandhi2024stream} extend the linearized search idea to a more realistic reasoning task - the Countdown game - which requires the model to predict a sequence of mathematical operations on a given set of numbers to match a target value. While~\citet{gandhi2024stream} use a fixed 250M parameter transformer model and do not explore or discuss the role of model size, training data, and complexity in terms of scaling performance, we obtain additional results in terms of inference-time scaling, shown in Figure~\ref{fig:SoSInferenceScalling}. Our findings demonstrate a consistent log-linear relationship between tokens spent and success rate. Similar results were also observed in recent work by~\citet{schultz2024masteringboardgamesexternal}, who train language models on linearized search traces obtained from MCTS on board game environments. Similar to the work of~\citet{gandhi2024stream}, they find consistent improvements in performance as the model is given additional search budget at test-time (Figure~\ref{fig:SoSInferenceScalling} right). Note that these models demonstrate an inference-time scaling law with the same functional form as the o1 model on difficult mathematics problems~\citep{openai2024_o1}.

\subsubsection{In-context Exploration For LLMs}
\label{sec:revisions}

\begin{figure}[t]
  \begin{center}
    \includegraphics[width=0.95\textwidth]{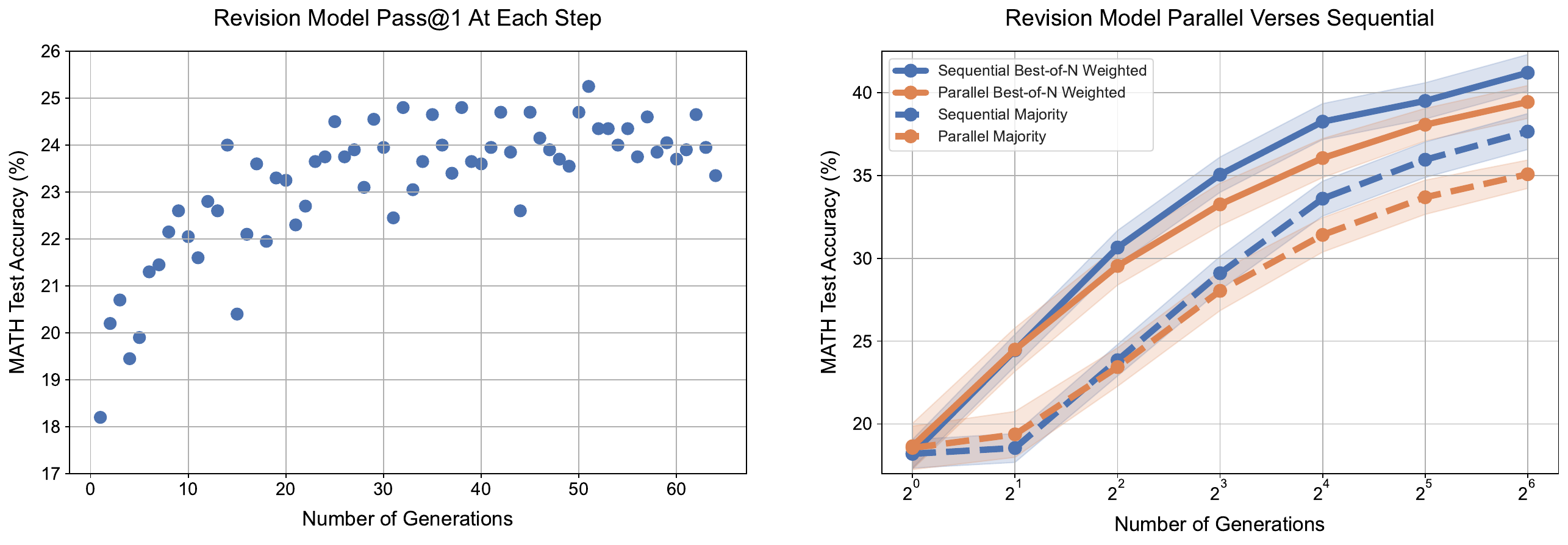}
  \end{center}
  \caption{\textbf{Left}: Pass@1 accuracy of a revision model after the specified number of generations (revisions). \textbf{Right}: Scaling performance of the best-of-N strategy under parallel and auto-regressive (in-context) sampling. The performance gap indicates that the model learns some degree of in-context exploration and self-correction. Source: Figure 6 from~\citep{snell2024scalingllmtesttimecompute}.}
  \label{fig:sequential_sampling}
\end{figure}

While the prior section showed promise in teaching auto-regressive language models to internalize complex search strategies involving exploration and backtracking, it remains unclear whether these results can generalize to realistic language domains. In this section we will overview several recent works, which show promise in internalizing episode-level search. Both~\citet{qu2024recursiveintrospectionteachinglanguage} and~\citet{snell2024scalingllmtesttimecompute} evaluate results using open-source LLMs in the 7B and larger range on problems from the MATH dataset~\citep{hendrycks2021measuringmathematicalproblemsolving}. They pose the problem as sequential sampling - i.e. given a problem $\mathbf{q}$, generating full solutions from the same model auto-regressively as
\begin{equation}\label{eq:autoregressive_sampling}
\mathbf{S}^j\sim \pi_{\theta}(\cdot|\mathbf{S}^{j-1}, \ldots, \mathbf{S}^1, \mathbf{q})
\end{equation}
\noindent where $\mathbf{S}^i$ are full solutions to the problem $\mathbf{q}$. Both works formulate the problem as self-correction, or revisions, during training. The approach generates training data by concatenating a number of incorrect solutions with the correct revision and training on a linearized sequence (although the exact training objective use a particular weighting grounded in RL~\citep{peng2019advantage}). The general objective follows the form
\begin{equation}\label{eq:revision_loss}
    \min_{\theta}\ \mathbb{E}_{\mathbf{S}^{i}\sim\pi_{\text{ref}}(\cdot|\mathbf{q}), \mathbf{q}\sim\mathcal{D}_{\text{train}}}\left[-\log \pi_{\theta}(\mathbf{S}^*|\mathbf{S}^{j-1}, \ldots, \mathbf{S}^1, \mathbf{q})\right]
\end{equation}
\noindent where $j$ is a fixed number of in-context exploration episodes sampled from a fixed distribution $\pi_{\text{ref}}$ (i.e. $\pi_0$) and $\mathbf{S}^*$ is some optimal solution. Essentially, this can be considered a linearization of the Best-Of-N search strategy presented in Section~\ref{sec:best-of-N} with rejection sampling. In this setting, the Meta-CoT represents search in full episodes $\mathbf{Z}=\mathbf{S}^1, \ldots, \mathbf{S}^{j-1}$ and $\mathbf{S}=\mathbf{S}^j$. At test time we can further control the quantity of compute by iteratively sampling from
\begin{equation}
    \mathbf{S}^i\sim \pi_{\theta}(\cdot|\mathbf{S}^{i-1}, \ldots, \mathbf{S}^{i-j}, \mathbf{q}).
\end{equation}
\noindent Representative results for this approach are are shown in Figure~\ref{fig:sequential_sampling}, sourced from~\citep{snell2024scalingllmtesttimecompute}. We see clear improvement in the pass@1 metric with additional amounts of in-context exploration episodes with nearly 6-7\% gain from zero-shot to the level of saturation. At the same time, auto-regressive generation shows clearly better scaling properties than independent parallel sampling (Figure~\ref{fig:sequential_sampling} right). These results indicate that the model learns some degree of in-context exploration and self-correction.

\subsubsection{Using variable Compute}
\begin{wrapfigure}{r}{0.45\textwidth}
    \vspace{-23pt}
    \centering
    \includegraphics[width=0.425\textwidth]{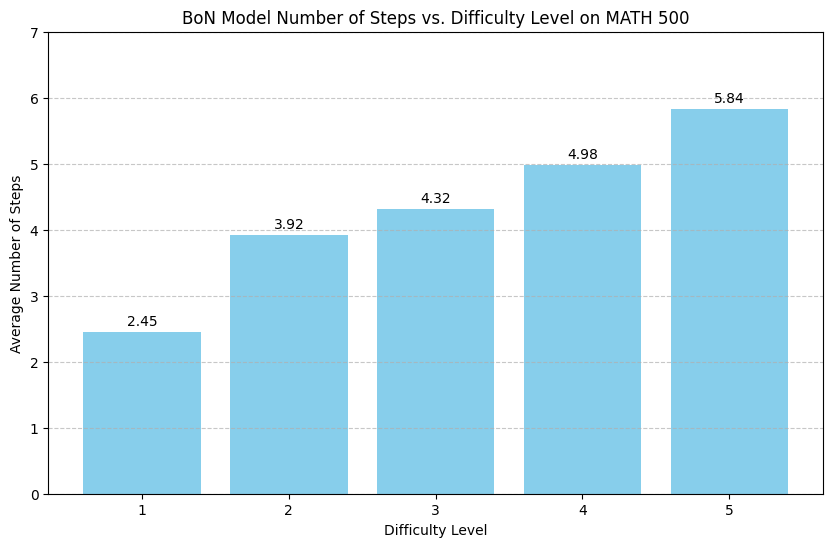}
    \caption{Number of in-context revisions the model attempts grouped by difficulty level.}
    \label{fig:compute_scaling_difficulty}
\end{wrapfigure}

While the above approaches demonstrate promise for the model's capability to carry-out in-context search, they are trained with a fixed number of revisions and use a pre-determined number of revisions at test time. This is not ideal, as ideally the model would be able to use arbitrary amounts of compute until it arrives at a solution with high enough confidence. We repeat the above experiment using a uniform number of in-context solutions during training (ranging between 0-7), allowing the model to generate up to 8 solutions at inference time by optimizing

\begin{equation}
    \min_{\theta}\ \mathbb{E}_{\mathbf{S}^{i}\sim\pi_{\text{ref}}(\cdot|\mathbf{q}), \mathbf{q}\sim\mathcal{D}_{\text{train}}}\left[-\log \pi_{\theta}(\mathbf{S}^*, \textbf{EOS}|\mathbf{S}^{j-1}, \ldots, \mathbf{S}^1, \mathbf{q})\right], j\sim\text{Unif}(1,8)
\end{equation}

This formulation encourages the model to continue revising it's solution until it reaches a solution with high confidence of correctness. Interestingly, our model generates an increasing number of solutions based on question difficulty. Summary statistics by problem difficulty are shown in Figure~\ref{fig:compute_scaling_difficulty} (right), where the model generates an average of 2.45 solutions for Level 1 problems and an average of 5.84 for Level 5 problems, consistent with the behavior shown in Figures~\ref{fig:hardmath_solutions} and~\ref{fig:searchformer_results} (c). Specifically, this demonstrates that the model has internalized the need for extended exploration on complex reasoning tasks. 

Our best performing run achieved an improvement of 2\% over the LLaMa 3.1 8B Instruct model that we initialize our models from. We attribute this to a mismatch between the instruct model's RL post-training, the off-policy SFT fine-tuning we utilize, and the limited amount of training data in the MATH train dataset. Indeed, any regular SFT training we performed on the RL post-trained model actually worsened performance. \textbf{We are currently exploring post-training on pre-RL base models with extended datasets.}

\subsubsection{Backtracking in LLMs}
\label{sec:backtracking}
In the prior sections, we reviewed evidence that auto-regressive models can internalize complex search strategies in simple domains. We also showed that LLMs can learn in-context exploration at the episode-level.
However, whether models can implement complex search strategies (e.g.\ those outlined in Section~\ref{sec:search}) auto-regressively remains an open question in public research. Specifically, we refer to the ability to terminate a reasoning chain prior to completion, and the ability to reset (semantically) to an arbitrary previously visited state in-context.
These two steps can be unified under the concept of backtracking. Here we will review some recent works demonstrating that LLMs can learn to backtrack.

Recent works have demonstrated that training on data with backtracking can improve language models on simple reasoning tasks~\citep{ye2024physicslanguagemodels22, anonymous2024improving} find that language models can sometimes ``recognize'' their errors internally, but do not have the required mechanisms to self-correct. Similar to our motivation with Meta-CoT, their aim is for a single model to both recognize errors and self-correct in-context. In their approach they create training data with incorrect steps directly followed by the correction. The backtracking is signified by a special token, \texttt{[BACK]}, at the end of an incorrect step to allow the model to explicitly state it's belief that an error has occurred. That is: given a dataset $\mathcal{D}_{\text{train}}$ of questions $\mathbf{q}$ and correct reasoning CoT solutions $\mathbf{S}=\mathbf{s}_1, \ldots, \mathbf{s}_n$ the training objective becomes
\begin{equation}\label{eq:backtracking}
    \mathcal{L}_{\text{backtrack}}(\theta) = -\mathbb{E}_{\mathbf{s}_1, \ldots, \mathbf{s}_n\sim \mathcal{D}_{\text{train}}, t\sim\text{Unif}(1,n)}\left[\log \pi_{\theta}(\mathbf{s}_1, \ldots, \mathbf{s}_t^-, \text{[BACK]},\mathbf{s}_t, \ldots, \mathbf{s}_n|\mathbf{q})\right]
\end{equation}
where $t$ is a randomly sampled time step in the solution and $\mathbf{s}_t^-$ is a single incorrect reasoning step. This is in contrast to the standard approach, which only trains on the correct solution chains:
\begin{equation}\label{eq:standard}
    \mathcal{L}_{\text{standard}}(\theta) = -\mathbb{E}_{\mathbf{S}\sim \mathcal{D}_{\text{train}}}\left[\log \pi_{\theta}(\mathbf{S}|\mathbf{q})\right].
\end{equation}
\citet{ye2024physicslanguagemodels22} explore inserting incorrect steps at varying rates (between 1\% and 50\%) and find that high rates of incorrect steps actually leads to improved downstream performance. In particular, they find that a 50\% rate of incorrect steps (objective in Equation~\ref{eq:backtracking}) leads to an increase from 78\% to 94\% accuracy on hard math problems as compared to training on only correct solutions (Equation~\ref{eq:standard}, CoT). While promising, these results are only verified on small models (124M parameters).

In contrast,~\citet{zhang2024backtrackingimprovesgenerationsafety} teach LLMs to backtrack based on safety considerations using the larger Gemma 2B and LLaMa 3 8B models. In particular, following the above notation, given a prompt $\mathbf{q}$ and two possible answers - a safe option $\mathbf{S}^+ = \mathbf{s}_1^+, \ldots, \mathbf{s}_n^+$ and an unsafe option $\mathbf{S}^-=\mathbf{s}_1^-, \ldots, \mathbf{s}_{n'}^-$, where $\mathbf{s}$ here represent individual tokens (unlike before where they stood for logical steps), they optimize the objective:
\begin{align}
    \begin{split}\label{eq:safety_backtracking}
        \mathcal{L}(\theta) = -\mathbb{E}_{(\mathbf{q}, \mathbf{S}^+, \mathbf{S}^-)\sim \mathcal{D}_{\text{train}}, t\sim\text{Unif}(1,n')}[ & \log \pi_{\theta}(\text{[BACK]}, \mathbf{S}^+|\mathbf{S}_t^-, \mathbf{q}) \quad + \\
        & \log \pi_{\theta}(\mathbf{S}^+|\mathbf{q})].
    \end{split}
\end{align}
That is a combination of the Meta-CoT and regular CoT objectives as outlined above. Additionally, notice that this objective masks out the unsafe completion, while the prior work trains on all tokens including the incorrect logical steps. While the approach of~\citet{ye2024physicslanguagemodels22} backtracks for a single logical step (correction) this work always resets the agent to the initial state. SFT training is successful in teaching the model to \textit{backtrack} and improves the safety characteristics over supervised fine-tuning on just the safe answer (only the second term of Equation~\ref{eq:safety_backtracking}). However, these effects appear weak in regular SFT models, but are significantly improved through further downstream RL training, which we will discuss later on.

\subsection{Synthetic Meta-CoT Via Search}
\label{sec:synth_meta_cot}


In the prior sections we argued for an approach to reasoning that teaches an LLM to internalize an auto-regressive search procedure in-context. We also reviewed several recent works showing that small auto-regressive models can carry out in-context exploration at the episode level, and larger models can learn individual step backtracking. In this section, we explore how to construct synthetic data for realistic Meta-CoT that involves full-scale in-context tree search.

\paragraph{Setup.} For demonstrative purposes, we use the math problem presented by~\citet{openai2024_o1} as our benchmark task, where Gemini 1.5 Pro~\citep{reid2024gemini} achieves a Pass@128 score of 6.25\% (8/128 correct) -- notably being the only frontier model (without advanced reasoning) to demonstrate non-zero performance at the time of our experiments. We use the same RL formulations for state and actions as presented in~\ref{sec:general_search}. We explore two principal search algorithms for generating synthetic training data: Monte Carlo Tree Search (MCTS) and A* variants. Both approaches necessitate a heuristic state estimation function, for which we employ pure Monte-Carlo rollouts following the methodology of~\citet{Silver2018AlphaZero}. Specifically, we estimate the value of a partial solution trajectory as
\begin{equation}\label{eq:MC_value}
    v(\mathbf{S}_t, \mathbf{q}) = \mathbf{E}_{\mathbf{S}_{\geq t+1}^j\sim \pi_{\theta}(\mathbf{S}_{\geq t+1}|\mathbf{S}_t, \mathbf{q})}\frac{1}{K}\sum_{j=1}^K r^*([\mathbf{S}^j_{\geq t+1},\mathbf{S}_t], \mathbf{q})
\end{equation}
where $r^*$ is the verifiable ground-truth outcome reward. In our experiments, we sample 128 completions from the partial solution and evaluate the mean success rate under ground-truth outcome supervision. In Appendix~\ref{sec:synth_cot}, the numerical values of the states are listed after each step.

\subsubsection{Monte-Carlo Tree Search}
\label{sec:MCTS_main}

\begin{wrapfigure}{r}{0.50\textwidth}
    \vspace{-13pt}
    \centering
    \includegraphics[width=0.475\textwidth]{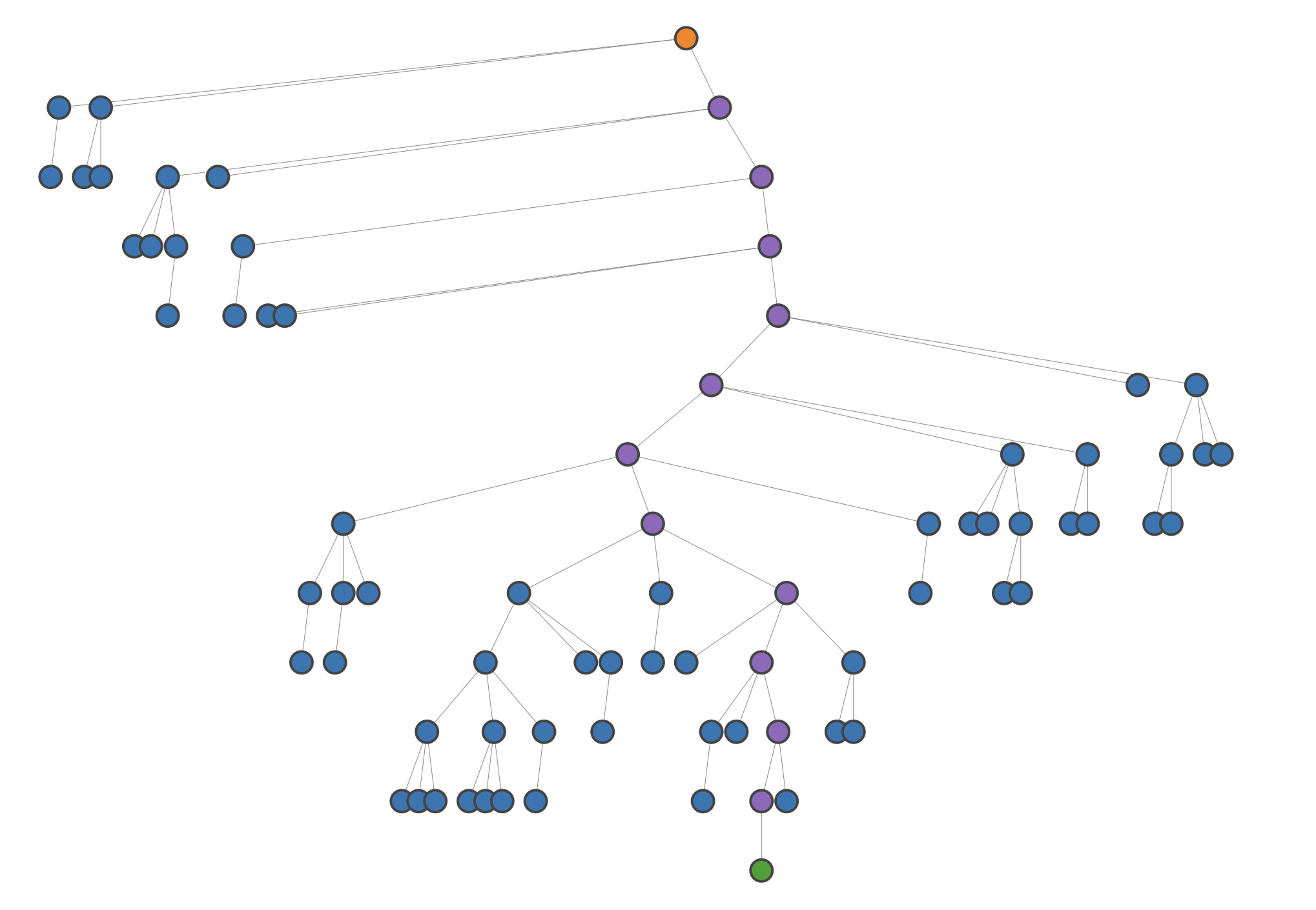}
    \caption{MCTS tree for the math problem presented by~\citet{openai2024_o1}. The red node indicates the solution.}
    \label{fig:MCTSTree}
\end{wrapfigure}
We conduct an example based on Monte-Carlo Tree Search (MCTS), which seeks to balance exploration and exploitation.
The MCTS implementation of~\citet{Silver2018AlphaZero} has been widely applied to the reasoning domain~\citep{tian2024selfimprovementllmsimaginationsearching, feng2024alphazerolike}, and we mostly follow their implementation with some modifications to account for the structure of our search problem (see Appendix~\ref{app:mcts}).

We present the search trace for our example problem - all the actions taken during the search (i.e., the Meta-CoT in a linear format) - in Appendix~\ref{sec:astar_on_openai_blog_problem}. The numbers following each reasoning step represent the value estimates. In our initial MCTS attempt we obtained a trace with an excessive number of backtracks and repetitions, including from high-value states (as high as 1.0) with the resulting exploration tree is shown in Figure~\ref{fig:MCTSTree}. We believe these effects are due to the exploration bonus in MCTS search. We did not carry out extensive ablations on the search parameters due to speed and costs. Since we use pure MC rollouts ("simulations") for state value estimation, a single tree uses up to 20 million tokens inference (a cost of $\sim$\$100 per tree). Moreover the process can take up to half an hour due to API limits. Because of these issues we also evaluate a more efficient best-first exploration strategy, which we present below.

\subsubsection{A* search}

\begin{wrapfigure}{r}{0.50\textwidth}
    \centering
    \includegraphics[width=0.475\textwidth]{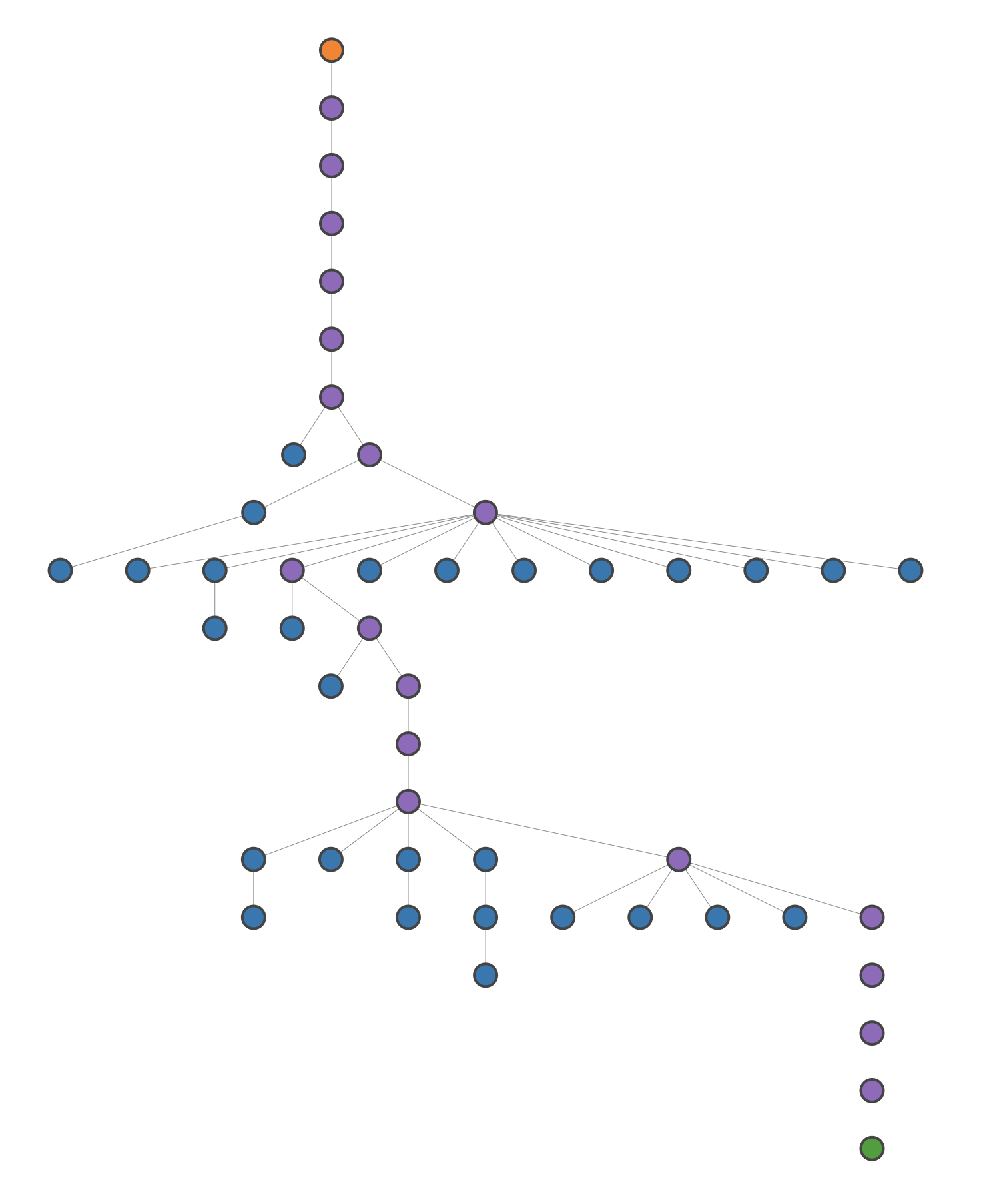}
    \caption{Resulting A* search tree on the math problem from \cite{openai2024_o1}. This trace presents more of a best-first approach with fewer backtracks, concentrated around key steps, as compared to the one produced by MCTS in Figure~\ref{fig:MCTSTree}.}
    \label{fig:AStarTree}
    \vspace{-54pt}
\end{wrapfigure}

We begin with an exploration of a type of best-first search based on the work by~\citet{koh2024tree}, which itself loosely follows an A* approach. The search procedure maintains a frontier $\mathcal{F}$ of states, which is implemented as a max priority queue. Similarly to the MCTS approach, each state $\mathbf{S}_t$ consists of the question $\mathbf{q}$ and a partial solution consisting of generated reasoning steps $(\mathbf{s}_1,\ldots,\mathbf{s}_t)$. 
At each iteration, the state $\mathbf{S}_p \leftarrow \text{pop}(\mathcal{F})$ with the highest value $v_p = v(\mathbf{S}_p, \mathbf{q})$ is selected, where $v_p\in [0,1]$ is the value of the partial solution $\mathbf{S}_p$ including current and previous reasoning steps. At each node the policy $\pi_{\phi}$ proposes $b$ candidate next steps, each of which is evaluated by $v$ and added to $\mathcal{F}$ if the depth of the tree $|(\mathbf{s}_0,\ldots,\mathbf{s}_p)|$ has not reached the maximum depth search limit $d$. For the purpose of generating synthetic data, we run the search until we find a solution that is correct using the ground-truth verifier. The resulting tree is shown in Figure~\ref{fig:AStarTree}. It shows more consistent flow of the
reasoning steps, with less backtracking concentrated around a few key steps. 


\subsection{Do Advanced Reasoning Systems Implement In-Context Search?}
\label{sec:reasoning_model_study}

In this section we will investigate whether advanced reasoning systems, such as OpenAI's O1~\citep{openai2024_o1}, DeepSeek R1~\citep{deepseek2024r1} and Gemini 2.0 Flash Thinking Mode~\footnote{\url{https://ai.google.dev/gemini-api/docs/thinking-mode}} and the Qwen QwQ \cite{qwq-32b-preview} implement in-context search. We provide successful reasoning traces for the same math problem in Appendix~\ref{mcts_on_openai_blog_problem}.

Starting with OpenAI's o1 model, by carefully examining the provided mathematical reasoning trace, we observe:

\begin{enumerate}
    \item Inconsistent flow of thought - consecutive steps do not logically continue the prior state. 
    \item Backtracking - the model carries out "semantic backtracking" - frequently returning to the same logical points.
    \item Repetition - the model often repeats logical steps.
\end{enumerate}

\begin{figure}[!htbp]
    \includegraphics[width=0.92\linewidth]{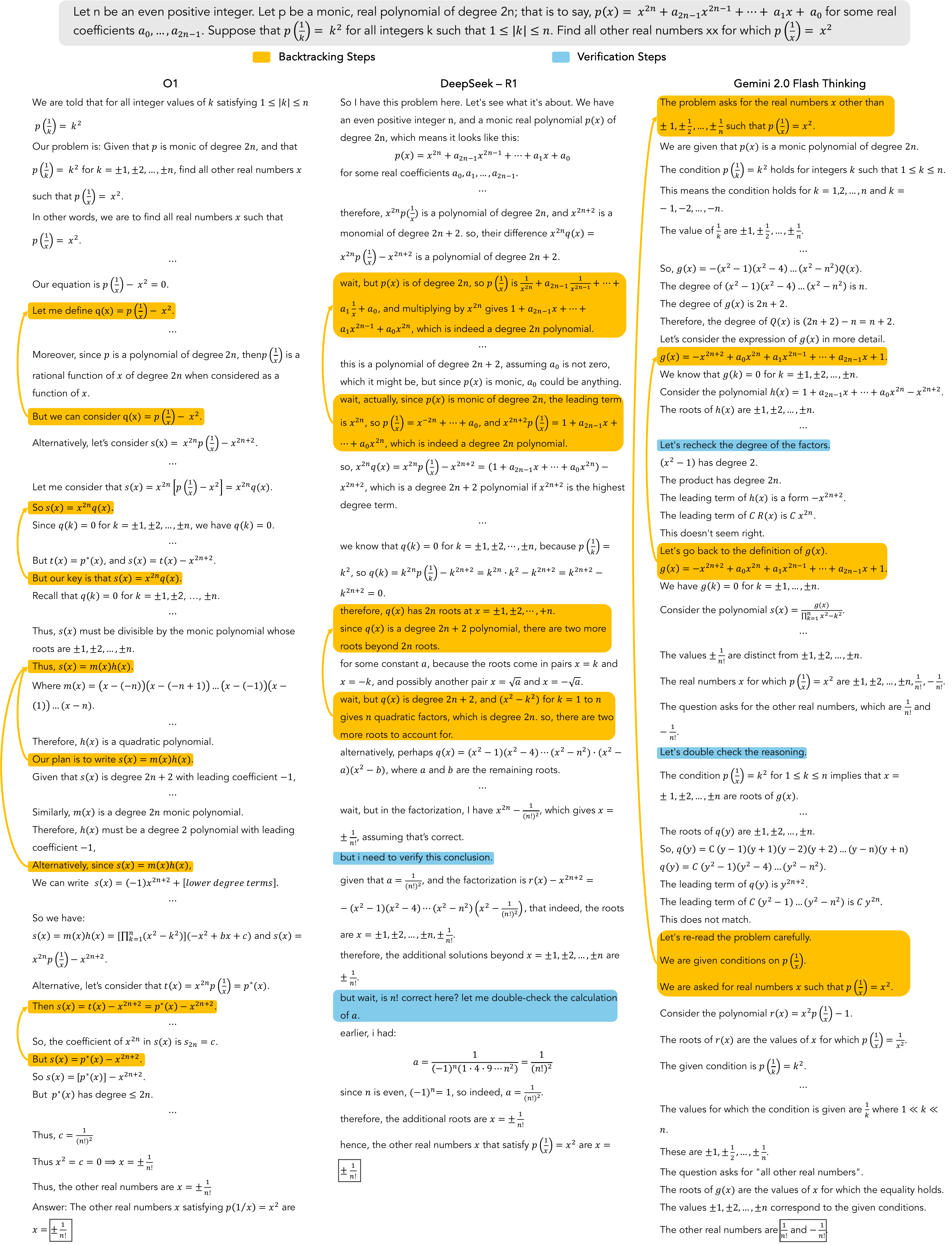}
    \caption{Examples of intermediate traces from 
    o1~\citep{openai2024_o1}, 
    DeepSeek-R1~\citep{deepseek2024r1}, 
    and Gemini 2.0 Flash Thinking Mode. We highlight two types of steps: \textcolor{brownishyellow}{Backtracking}, where the model visits a bad state and returns to a previously visited step, and \textcolor{cyan}{Verification}, where the model assesses the correctness of the previous output. Inconsistent logical flow and repetition are present in all three traces. DeepSeek-R1 and Gemini 2.0 Flash Thinking Mode both exhibit generative verification before reaching an answer, while Gemini makes an incorrect verification and returns to the initial state. Full search traces can be found in 
    Appendix~\ref{sec:synth_cot}.
    }
    \label{fig:backtrack_examples_1}
\end{figure}

\begin{figure}
    \centering
    \includegraphics[width=0.99\linewidth]{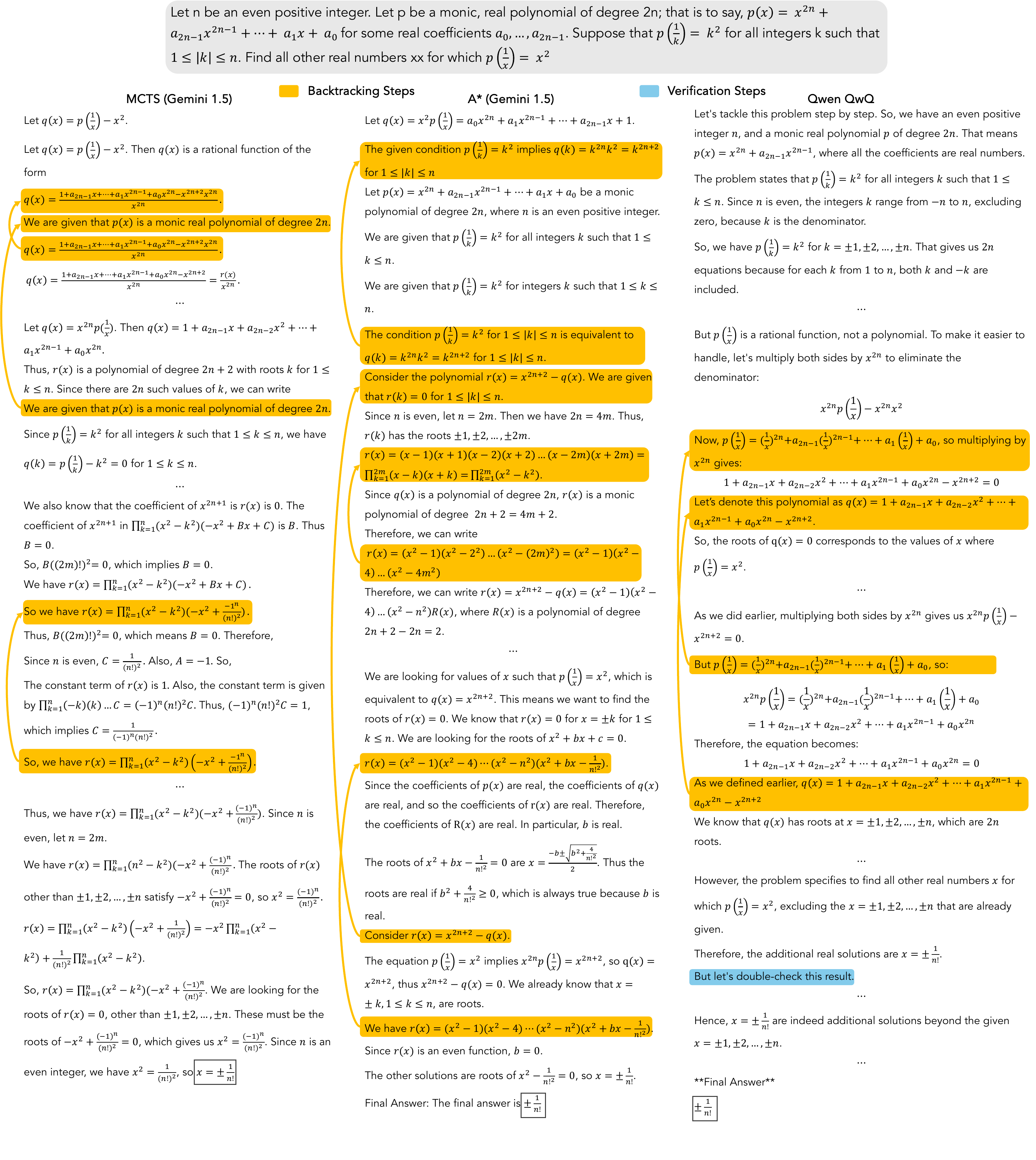}
    \caption{Example synthetic search traces from Gemini 1.5 with MCTS and $A^{*}$. We highlight \textcolor{brownishyellow}{backtracking} steps, where the model visits a bad state and returns to a previously visited step. Inconsistent logical flow and repetition are present in both traces, but stands out particularly in MCTS trace. Full search traces can be found in Appendix~\ref{mcts_on_openai_blog_problem}.} 
    \label{fig:backtrack_examples_2}
\end{figure}

The qualitative behaviors observed in o1 (Figure~\ref{fig:backtrack_examples_1} left) are similar to those in the example synthetic trace (Figure~\ref{fig:backtrack_examples_2}) generated by Gemini 1.5 with and MCTS-like search processes. In particular, there are abrupt changes in logical flow of the (Meta) CoT, which is natural as the model backtracks between branches of the tree. Moreover, the model may explore multiple child nodes of the same parent which are different strings, but can also be very semantically similar leading to repetitive logic. This is clear in the provided trace, as the model repeats logical statement and goes over the same derivations multiple times. 
Note also that we do not claim the model is implementing tree search at \textbf{test time}, but rather that as much as the model's output are expected to resemble it's training data, we hypothesize that examples of search were used during training (likely model initialization). We will specifically address the need and effects of RL training in Section~\ref{sec:meta_RL}. 

The DeepSeek R1 model \cite{deepseek2024r1} also exhibits similar behaviors, as shown in Figure~\ref{fig:backtrack_examples_1}, however, it also carries out a significant amount of self-evaluation steps. This could be achieved by integrating a form of self-criticism~\citep{madaan2023selfrefineiterativerefinementselffeedback, shinn2023reflexionlanguageagentsverbal} or a generative verifier~\citep{zhang2024generativeverifiersrewardmodeling} in the search trace. The LATS framework~\citep{zhou2024languageagenttreesearch} uses a similar approach, combining MCTS search with self-criticism and shows empirical improvements from self-reflection. Another alternative for synthetic data generation is the "Iteration-Of-Thought" approach \cite{radha2024iterationthoughtleveraginginner} which also interleaves generation with inner dialogue. This would explain the rather smooth logical flow of the R1 model, which does not exhibit as much abrupt back-tracking, as compared to O1. As mentioned earlier, in order to adequately model the search process the model must internalize an evaluation mechanism. However, providing an explicit CoT verification may be able to expand the model computational capacity and improve self-verification. This is an empirical question, which is currently unclear in open research.

Gemini 2.0 Flash Thinking Mode appears to implement a somewhat different structure. Specifically, the flow of reasoning qualitatively appears smoother with fewer logically inconsistent steps. Moreover, it backtracks less frequently and often returns to the initial state. In fact in the provided example the model solves the problem correctly and then fully re-generates a new solution from scratch (backtracks from the final state to the initial one). It's behavior seems to be to attempt a full solution, which may be terminated early based on some search heuristic. In cases where the solution attempt is unsuccessful, the model attempts a different solution approach, rather than branch at the step-level in a tree search structure. This seems more consistent with a revision-based strategy as reflected in past works~\citep{qu2024recursiveintrospectionteachinglanguage, anonymous2024improving, kumar2024traininglanguagemodelsselfcorrect}. The Qwen QwQ model \cite{qwq-32b-preview} shows similar behavior, generating multiple solutions in-context, as also pointed out by \cite{chen2024think23overthinkingo1like}.

\section{Process Supervision}
\label{sec:process_supervision}
A key component of the search approaches presented in prior sections is the evaluation function $v(\mathbf{q}, \mathbf{S}_t)$, which scores intermediate states in a reasoning chain. These evaluation functions have become widely known as Process Reward Models (PRM). By incorporating process supervision, the search mechanism gains the flexibility to backtrack to earlier promising states when suboptimal paths are encountered, thereby enabling more effective exploration. However, the question of how to efficiently access such capabilities remains an open question. In Section~\ref{sec:synth_meta_cot} we showed
examples of using outcome-based verification with MCTS in combination with Monte-Carlo rollouts. However, this approach can only be used during training due to the necessity for ground-truth answers, and moreover it is \textbf{extremely} computationally inefficient. As mentioned earlier, a single training example requires up to 20 million inference tokens, costing up to hundreds of dollars. It is significantly more efficient to amortize the evaluation procedure into a single parameterized model, and we will outline strategies for building such process guidance models below. 

\subsection{Learning Process Reward Models}
Parameterized PRMs are built on top of pre-trained models, either using a linear head or the logits of specific tokens. The model takes the question $\mathbf{q}$ and a partial solution $\mathbf{S}_t$ as input and outputs a single scalar value $v_{\theta}(\mathbf{q}, \mathbf{S}_t)\to[0,1]$. Given a dataset $\mathcal{D}_{\text{train}}$ of partial solutions $\mathbf{S}_t$ and corresponding value targets $y_{\mathbf{S}_t}$ the model is generally optimized with a standard cross-entropy classification loss.
A central question for training PRMs is: where do the supervision labels $y_{\mathbf{S}_t}$ come from? One approach is to have human annotators provide step-by-step level evaluation of reasoning problems, as done by~\citet{lightman2023let}. While their work showed promise in terms of empirical results, this method is challenging to scale due to the high annotation time and cost, especially as evaluating hard reasoning problems requires high-caliber experts. An alternative approach presented by~\citet{wang2024math} only relies on access to outcome verification - i.e. problems with a ground truth answer. The proposed approach is to \textit{amortize} the Monte Carlo state-value estimation into a parameterized function. Essentially, this method fits an empirical value function of the reference rollout policy where the targets $y_{\mathbf{S}_t}$ are represented by Equation~\ref{eq:MC_value}. This idea has been widely adopted in follow-up works~\citep{snell2024scalingllmtesttimecompute, anonymous2024improving} and further extended~\citep{setlur2024rewardingprogressscalingautomated}.

\subsection{PRM Quality And Its Effect On Search}
The performance and efficiency of search at test-time depends on the quality of the PRM~\citep{setlur2024rewarding, anonymous2024improving}.~\citet{setlur2024rewarding} demonstrate effective scaling (in both training data size and label quality) of a specific variant of PRMs that estimate values based on the improvement in likelihood of the correct answer after a step. The accuracy of test-time search improves log-linearly with training data size, and the quality of learned value labels improve with more Monte Carlo estimates.~\citet{anonymous2024improving} show that oracle verifier-enabled search is orders of magnitude more efficient than a learned PRM with noisy value estimates. 

In this section we conduct an experiment demonstrating the scaling characteristics of a PRM. To train our PRM, we first need to generate diverse solution trajectories where each solution step is annotated with a ground truth value. To do so, we use the method from~\citet{wang2024math} to obtain ground truth values, performing 16 Monte Carlo (MC) rollouts for each step of a seed solution. We generate the seed solutions and step-level MC rollouts from a supervised finetuned (SFT) Llama3.1-8B using the PRM800K~\citep{lightman2023let} dataset. The PRM training data uses 7,086 unique questions - each with seed solutions - and after removing duplicate seed solutions results in 97,000 trajectories in the training data. To evaluate the scaling performance with increasing data, we split the small set of data into three subsets: one with 500 unique questions, one with 3,000 unique questions, and one with all 7,086 unique questions. We create an evaluation set using the MATH-500 dataset~\citep{hendrycks2021measuringmathematicalproblemsolving, lightman2023let} by generating step-by-step solutions from the SFTed model and step-level ground truth values from 128 MC rollouts. 
\begin{figure}[t]
    \centering
    \includegraphics[width=0.98\linewidth]{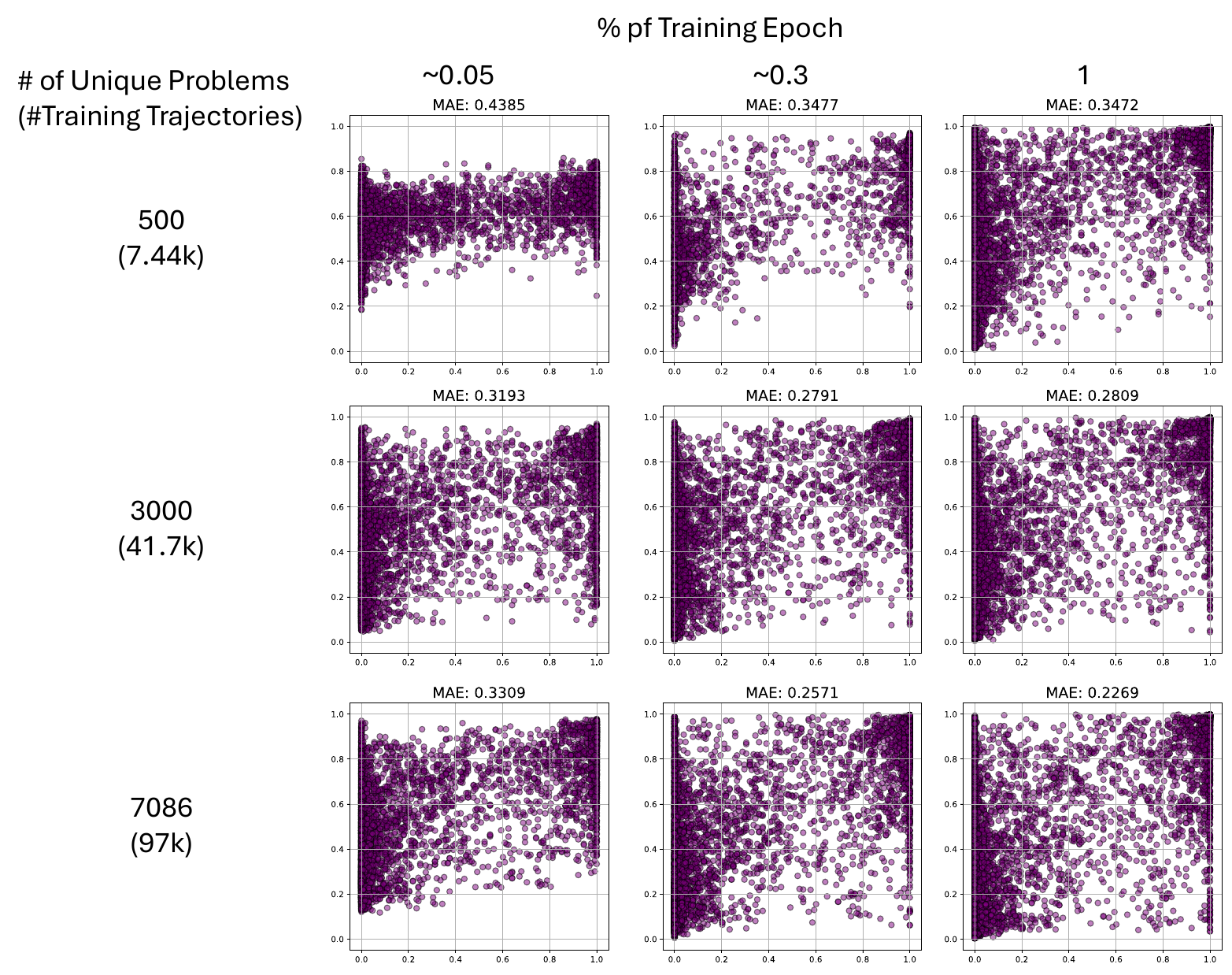}    
    \caption{Distribution of a learned PRM's predicted values for each state vs. ground truth (computed using 128 MC samples) as unique training questions increase. Mean absolute error (MAE) decreases as the PRM is trained with more questions. When the quantity of training data is small the performance on the test set converges early ($\sim 30\%$ of an epoch in training).}
    \label{fig:prm_values}
\end{figure}

\begin{figure}[t]
    \centering
    \includegraphics[width=0.45\linewidth]{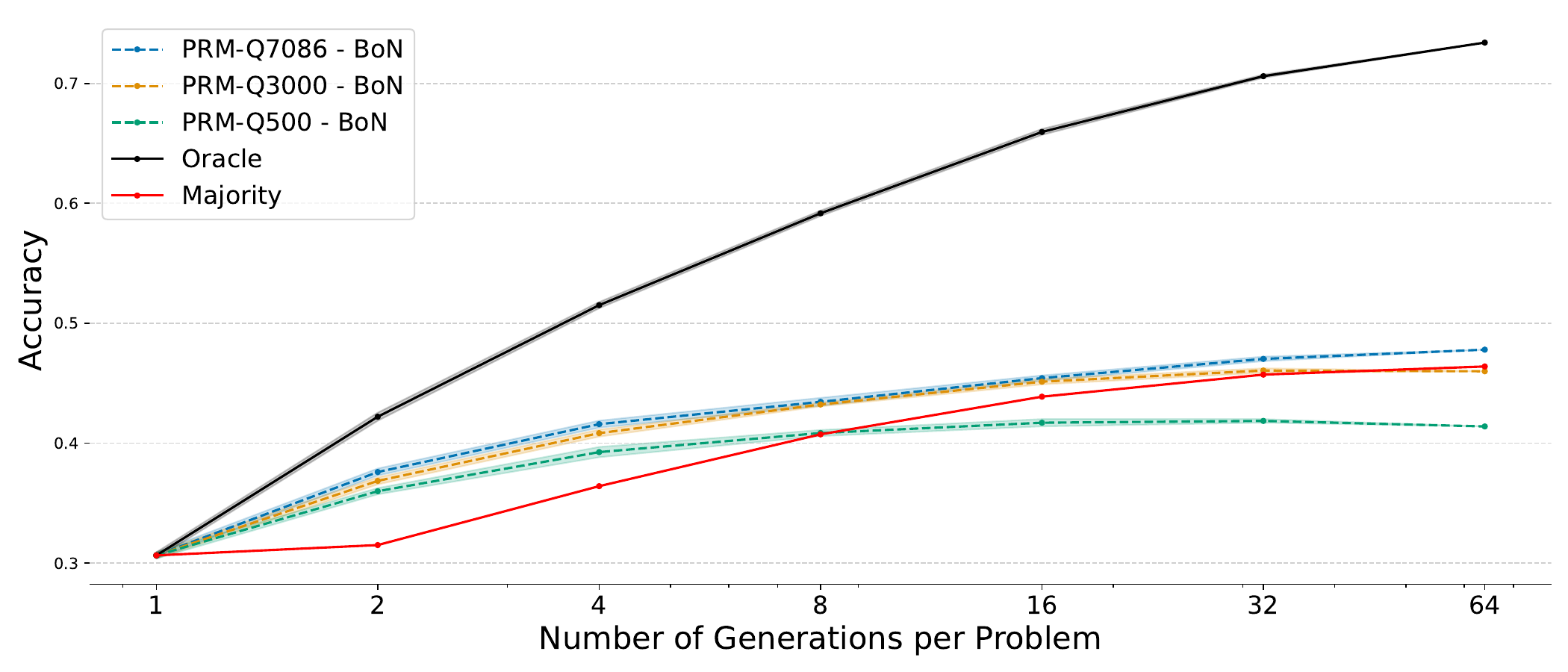}    
    \includegraphics[width=0.48\linewidth]{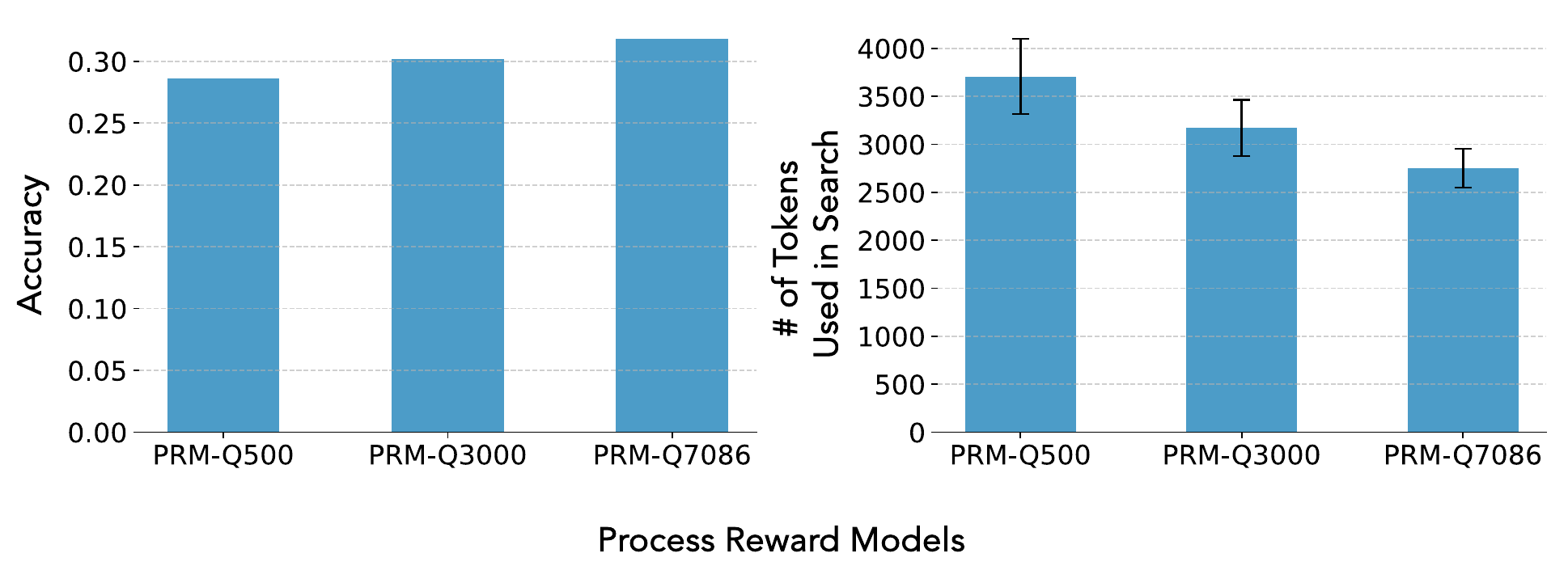}        
    \caption{\textbf{Left:} Scaling curves for Best-of-N (BoN) using PRMs trained with different number of questions with oracle and majority vote. \textbf{Right:} Beam search (N=5, beam width=4) accuracy and number of tokens used during search with the same PRMs. With more training data, the PRM's ability to verify at outcome-level and process-level improves.}
    \label{fig:prm_scaling_curve}
\end{figure}

With this trained PRM, we find a reduction in the absolute error of predicted values when comparing PRMs that are trained across datasets of different sizes, as well as a selection of intermediate checkpoints in Figure~\ref{fig:prm_values}. We observe that: 1) the prediction error decreases as the size of the training data increases, and 2) when the size of the dataset is small, improvement converges early during training (around 30\% of an epoch for Qs=500 and Qs=3000). Although these findings are based on small-scale experiments, we anticipate continued improvement in prediction errors with larger datasets and more extensive training, suggesting significant potential in further refining and scaling PRMs. Additionally, we evaluate the performance of the three fully-trained PRMs as outcome verifiers when performing a Best-of-N search during inference time. Figure~\ref{fig:prm_scaling_curve} \emph{left} shows that the PRM's ability to verify full solutions improves as they are trained with more data, yet there exists a remarkable gap between the trained PRMs and an oracle PRM. Additionally, we observe that the PRM's ability to guide the search process towards the correct answer with a more efficient path also improves as the increased accuracy and reduced number of tokens used in the search process are both observed in Figure~\ref{fig:prm_scaling_curve} \emph{right}. One interesting remaining question is: what is the scaling law for these process supervision models?

\subsection{Verifiable Versus Open-Ended Problems}
Training a value function with MC rollouts is scalable \textit{with infrastructure and inference}, but is fundamentally limited to problems with verifiable solutions. This excludes proof problems and scientific derivations which are often more important than the numerical answer itself. While automated proof assistance is an established area of research in mathematics~\citep{ThemathlibCommunity2020}, this is rather limiting. First of all, these methods are largely limited to math and do not transfer to other domains such as science or more general problem-solving scenarios. In those domains, training a PRM based on human evaluations of valid reasoning steps could yield a general verifier, which can be used for assuring the validity of the proof/solution chain. This would explain the need for human annotators and verification.

\section{Meta Reinforcement Learning - Learning \textbf{How} To Think}\label{sec:meta_RL}
\begin{figure}[t]
     \centering
     \includegraphics[width=0.975\linewidth]{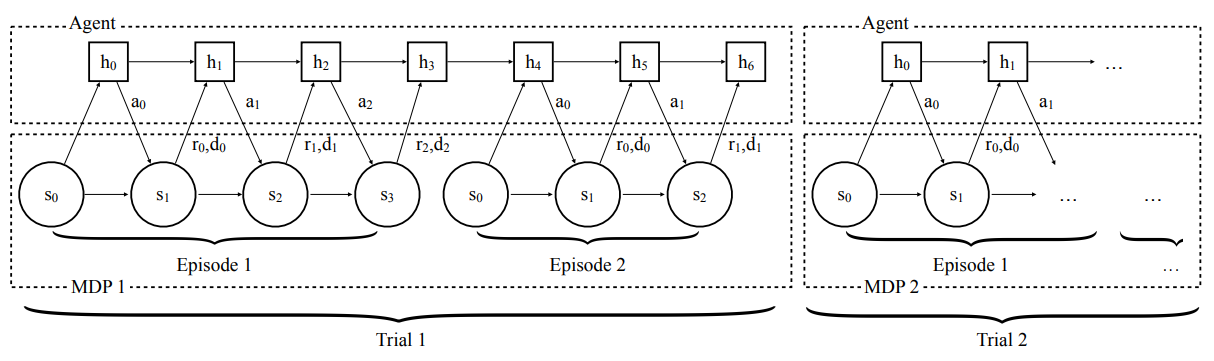}
     \caption{\textbf{Right}: The paradigm of the RL$^2$ formulation of meta-reinforcement learning. For each new task (prompt), the agent explores its environment over multiple episodes, keeping all the experience in context and maximizes rewards over the entire interaction. Source: Figure 1 in~\citep{duan2016rl2fastreinforcementlearning}.} 
     \label{fig:enter-label}
\end{figure}

In this section we will build out an interpretation of the reasoning problem and Meta-CoT from the perspectives of meta-learning and meta-RL. In Section~\ref{subsec:linear_search} we motivated the need for in-context search through the paradigm of computational complexity and the generator-verifier gap. In this section, we build an alternative formulation which will help us formalize empirical results of RL training. In particular, we consider the search problem in the deterministic MDP formulation from Section~\ref{sec:general_search}, however, in this section we assume the reward function $r(\mathbf{S}, \mathbf{q})\to \{0, 1\}$ is a deterministic (but a-priori unknown) function of the prompt $\mathbf{q}$, which accepts only a particular set of solutions. At \textbf{test time} under a new prompt, this creates \textbf{epistemic uncertainty} of the reward function i.e. a-priori we do not know the full set of accepted or rejected solutions for this task (prompt question). This process turns the MDP formulation we previously outlined in \ref{sec:general_search} into a Partially Observable MDP (POMDP), a view formalized in classical RL by~\citet{ghosh2021generalizationrldifficultepistemic}. Their work proves the following general remark:

\begin{tcolorbox}[colback=blue!5!white,colframe=blue!75!black]
\textbf{Remark 5.2 from~\citet{ghosh2021generalizationrldifficultepistemic}:} The expected test-time return of policies that are learned by maximizing reward in any MDP from the posterior, as standard RL methods do, may be arbitrarily low compared to that of the Bayes-optimal behavior.
\end{tcolorbox}

\begin{wrapfigure}{r}{0.50\textwidth}
    \centering
    \includegraphics[width=0.475\textwidth]{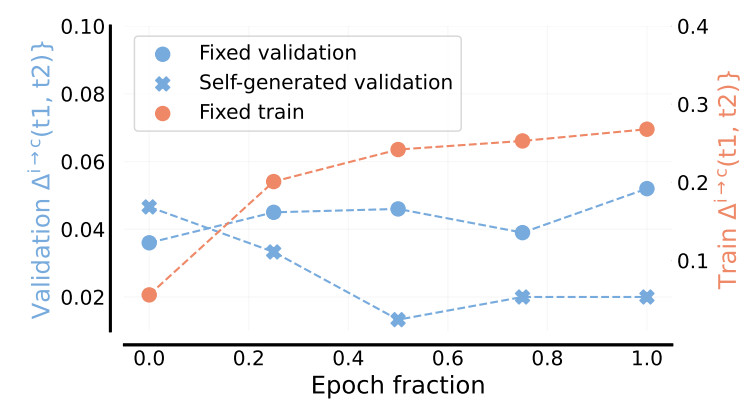}
    \caption{\textbf{Self-correction performance} from self-generated vs. fixed first steps. As training progresses (from left to right) the model becomes more capable of correcting errors in solutions from the reference distribution, but less capable of correcting its own induced errors. Source: Figure 5 from ~\citep{kumar2024traininglanguagemodelsselfcorrect}.}
    \vspace{-20px}
    \label{fig:SCoRe_dist_shift}
\end{wrapfigure}

In other words, regular RL-trained policies can, in theory, have arbitrarily bad performance on new reasoning domains ~\citet{setlur2025opt}. This statement is expanded on and made mathematically precise by~\citet{ghosh2021generalizationrldifficultepistemic} and follow-up works, however that formalization is beyond the scope of the current report. In this work, we stipulate that the reasoning problem, as a POMDP, is better suited to a meta-RL framework~\citep{humplik2019metareinforcementlearningtask, rakelly2019efficientoffpolicymetareinforcementlearning} rather than the regular RL setting, as described above. In the meta-learning setting we are given a distribution of tasks, which in our case would be individual reasoning problems $\mathbf{q}\sim\mathcal{D}_{\text{train}}$. The meta-learning objective seeks a policy $\pi_\theta$ and an adaptation procedure $U$ such that, for a sampled task $\mathbf{q}$, the adapted policy $\pi_{U(\theta)}$ performs well with minimal additional samples. The overall optimization objective is
\[
\min_{\theta} \mathbb{E}_{\mathbf{q}\sim\mathcal{D}_{\text{train}}} \mathbb{E}_{\pi_{U(\theta)}} \left[L_{\mathbf{q}}(\theta)\right],
\]
where $L_{\textbf{q}}$ represents the loss associated with task $\textbf{q}$. For example, in the revision formulation of Section~\ref{sec:revisions} the objective is
\begin{equation}\label{eq:meta_objective}
    \min_{\theta} \mathbb{E}_{\mathbf{q}\sim\mathcal{D}_{\text{train}}} \mathbb{E}_{\mathbf{S}^i\sim \pi_{\text{ref}}(\cdot|\mathbf{q})} \left[-\log\pi_{\theta}(\mathbf{S}^*|\mathbf{S}^j, \ldots, \mathbf{S}^1, \mathbf{q})\right],
\end{equation}
where $\mathbf{S}^*$ is the optimal solution to the problem $\mathbf{q}$ and $\mathbf{S}^i, i=1, \ldots, j$ are solutions provided by some reference policy (usually $\pi_{\theta_0}$). Here the adaptation procedure is represented by the operator $\pi_{U(\theta)}(\cdot|\mathbf{q})\to\pi_{\theta}(\cdot|\mathbf{S}^j, \ldots, \mathbf{S}^1, \mathbf{q})$, which is reminiscent of meta-learning with memory networks \cite{santoro2016oneshotlearningmemoryaugmentedneural}. The issue with this approach is that at inference time, given a new test problem $\mathbf{q}$, we sample solutions auto-regressively from the current iteration of the model $\pi_{\theta}(\cdot|\mathbf{q})$ rather than $\pi_{\theta_0}$ which generated our training data, which generates a train-test distribution shift.
Indeed,~\citet{kumar2024traininglanguagemodelsselfcorrect} noted a continuous shift during training even with $j=1$ where the model successfully corrects wrong solutions from the reference training data, but it's capability to self-correct (correct wrong solutions sampled from the current policy) diminishes - refer to Figure~\ref{fig:SCoRe_dist_shift} (sourced from~\citet{kumar2024traininglanguagemodelsselfcorrect}) for
empirical results. If the reference model $\pi_{\text{ref}}$ does not generate data with sufficiently high coverage, then this distribution shift \textbf{fundamentally necessitates the use of on-policy reinforcement learning approaches}. Unlike conventional reinforcement learning, where the objective is to optimize for immediate rewards, meta-RL emphasizes training agents to quickly explore a new environment and adapt to the task at hand. This requires optimizing the sampling process during meta-training to ensure the adaptation process $U$ maximizes the agent's final performance. 

If we modify the meta-learning objective in Equation~\ref{eq:meta_objective} with on-policy sampling, we can essentially recover a formulation of the RL$^2$~\citep{duan2016rl2fastreinforcementlearning} approach, which has strong synergies with LLMs due to their in-context learning capabilities. In this setting the agent, represented as a recurrent policy, encounters a series of tasks $\textbf{q}$ and interacts with them for several episodes with persistent intra-episode memory for the particular task. Here, the goal of the agent is to maximize the accumulated reward over $K$ episodes:
\begin{equation}\label{eq:RL2}
\max_{\pi_{\theta}}\mathbb{E}_{\mathbf{q}\sim\mathcal{D}_{\text{train}}}\mathbb{E}_{\mathbf{S}^{j}\sim\pi_{\theta}(\cdot|\mathbf{S}^{j-1}, \ldots, \mathbf{S}^{1}, \mathbf{q})}\left[\sum_{j=1}^Kr(\mathbf{S}^j, \mathbf{q})\right]
\end{equation}
across the distribution of tasks, also notice that here the expectation is taken over the current policy iterate $\pi_{\theta}$, removing the issue of distribution shift. We will draw some additional connections between Equation~\ref{eq:meta_objective} and Equation~\ref{eq:RL2} in the next section.  This objective be optimized through standard reinforcement learning algorithms such as REINFORCE~\citep{williams1992reinforce} and PPO~\citep{schulman2017proximal}. While successful in classical meta-RL tasks, this approach can lead to policy collapse on locally greedy behavior - i.e. it may not be able to fully explore new environments but instead collapse to common behaviors. Addressing this issue,~\citet{stadie2019considerationslearningexploremetareinforcement} propose a simple modification, E-RL$^2$, which considers the objective
\begin{equation}\label{eq:ERL2}
\max_{\pi_{\theta}}\mathbb{E}_{\mathbf{q}\sim\mathcal{D}_{\text{train}}}\mathbb{E}_{\mathbf{S}^{j}\sim\pi_{\theta}(\cdot|\mathbf{S}^{j-1}, \ldots, \mathbf{S}^{1}, \mathbf{q})}\left[r(\mathbf{S}^K, \mathbf{q})\right]
\end{equation}
aiming to maximize the return over the \textit{final} episode only. This allows the policy to explore without reward supervision for $K-1$ episodes, providing wider coverage of the environment which allows the policy to maximize rewards in the final episode. This slight modification mostly prevents the collapse to greedy behaviors seen in the standard RL$^2$ algorithm. Since these early works there has been significant follow-up literature and we refer the reader to the survey of~\citet{beck2024surveymetareinforcementlearning} for an overview. 

The above discussion uses the standard RL$^2$ episodic formulation as it was studied in classical MDPs, however, this is not strictly necessary in the language setting outlined in Section~\ref{sec:internal_reasoning}. If we equip the agent with the capability to
\begin{enumerate}
    \item terminate an episode early - i.e. achieve the information objective before it reaches the final solution
    \item reset to an arbitrary state in context rather than restart the solution from the first step
\end{enumerate}
\noindent then the meta-RL formulation remains valid for any in-context exploration strategy, including general tree search approaches. In purely language domains, such as mathematical reasoning, this is feasible and we can directly apply the E-RL$^2$ algorithm outlined above to models such as those by~\citet{yang2022chainthoughtimitationprocedure} ,~\citet{lehnert2024abetterplanningtransformers}, or~\citet{gandhi2024stream}. \textbf{However, in domains where the policy does not have full control of the environment, such as code or agentic formulations, different search structures may be required.}

\subsection{Meta-RL In Small Domains}

\begin{figure}
    \centering
    \includegraphics[width=0.3625\linewidth]{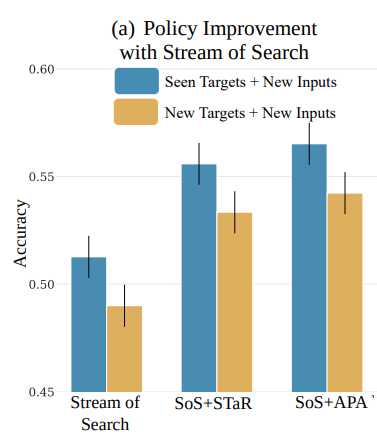}
    \includegraphics[width=0.3000\linewidth]{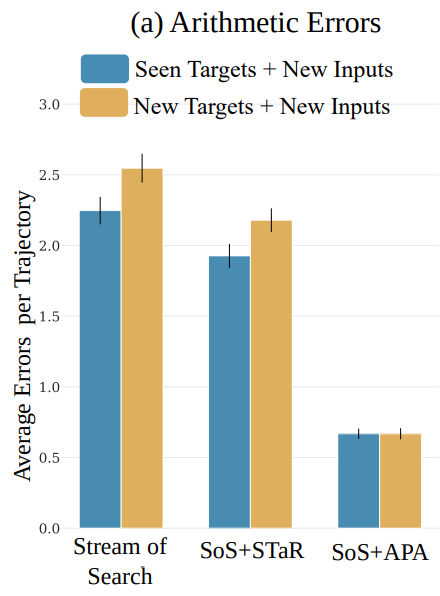}
    \includegraphics[width=0.3125\linewidth]{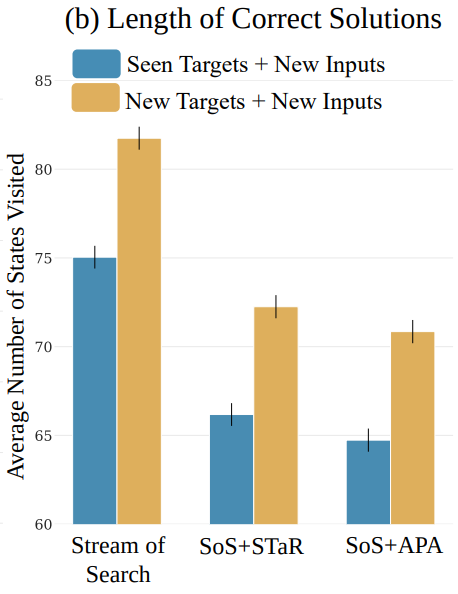}
    \caption{\textbf{The benefits of reinforcement learning for langauge model reasoning.} When comparing Expert Iteration (SoS+STaR) vs. the RL-based SoS+APA, we see that the use of RL leads to improved policy performance (left), with fewer arithmetic errors (center), and improved efficiency (right). Source: (left to right) Figures 4a, 6a, and 6b from~\citep{gandhi2024stream}.}
    \label{fig:sos_results}
\end{figure}
In the prior section we argued for on-policy RL in reasoning systems to handle train-test distribution shifts, but whether pure instruction-tuning (without RL) is enough to induce capabilities in the model is still debated. In general, the use of RL in post-training improves model performance. Indeed, in Section~\ref{sec:internal_reasoning} we outlined the capability to discover improved exploration (adaptation $U$) algorithms through RL as a major advantage, but the degree to which this occurs remains unclear. Specifically, we want to answer the question: can an in-context search algorithm, post-trained with reinforcement learning, outperform a modular search system? One such example exists where, as shown in Figure~\ref{fig:sos_results}, RL post-training improves overall performance in terms of accuracy, reduces the number of logical mistakes, and makes the search more efficient. However, the overall performance is still \textbf{only comparable} to the modular (symbolic) search paradigm - i.e. while RL significantly improves performance over the pure SFT (instruction-tuned) model, RL is not currently able to discover \textbf{new search algorithms}. Whether this is a fundamental limitation of the environment, pre-training data, or scaling is currently an open question in the literature.


\subsection{Meta-RL In Language Model Reasoning}
\begin{figure}
    \centering
    \includegraphics[width=0.95\linewidth]{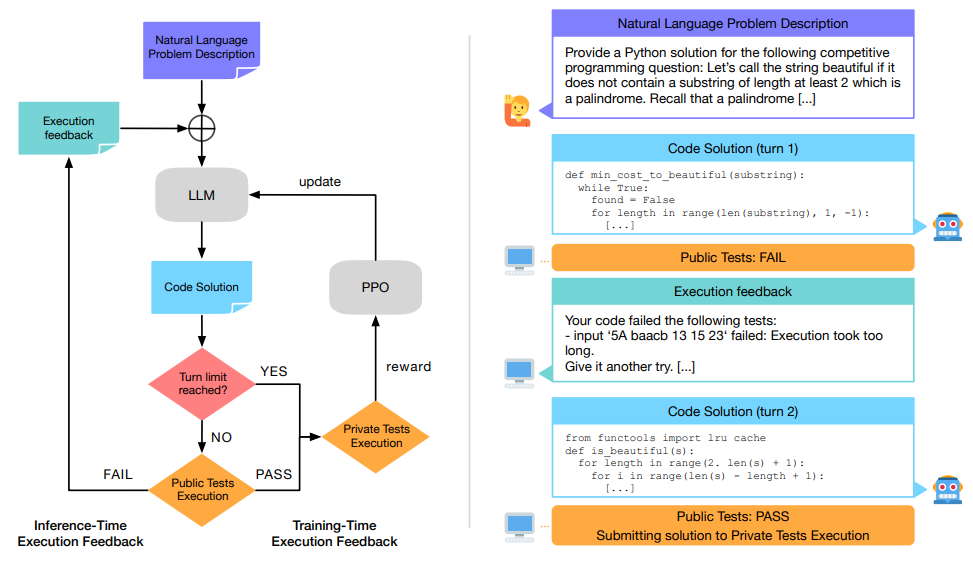}
    \caption{Overview of Reinforcement Learning with Execution Feedback. This training routine directly maps to the E-RL$^2$ framework~\citep{stadie2019considerationslearningexploremetareinforcement}. Source: Figure 2 in~\citep{gehring2024rlefgroundingcodellms}.}
    \label{fig:rlef}
\end{figure}

\begin{wraptable}{r}{0.5\textwidth}
    \vspace{-23pt}
    \centering
    \begin{tabular}{l l r r}
        \toprule
        \multirow{2}{*}{Model} & Training & \multirow{2}{*}{Valid} & \multirow{2}{*}{Test} \\
        & Method & & \\
        \midrule
        \multirow{4}{*}{\begin{tabular}{l} Llama 3.1 \\ 8B Instruct \end{tabular}} & – & 8.9 & 10.5 \\
        & Few-Shot & 8.5 & 8.5 \\
        & SFT & 10.3 & 10.0 \\
        & RLEF & 17.2 & 16.0 \\
        \midrule
        \multirow{4}{*}{\begin{tabular}{l} Llama 3.1 \\ 70B Instruct \end{tabular}} & – & 25.9 & 27.5 \\
        & Few-Shot & 22.5 & 20.3 \\
        & SFT & 27.7 & 27.2 \\
        & RLEF & 37.5 & 40.1 \\
        \bottomrule
    \end{tabular}
    \caption{Comparison of different training methods for 8B and 70B Instruct models on validation and test datasets. ``--'' directly evaluates the instruct model. Source: Table 3 in~\citep{gehring2024rlefgroundingcodellms}.}
    \label{tab:sft-vs-rl}
\end{wraptable}

Two works that have scaled the multi-turn formulation of meta-RL for reasoning tasks are~\citet{kumar2024traininglanguagemodelsselfcorrect} and~\citet{gehring2024rlefgroundingcodellms}. While Reinforcement Learning with Execution Feedback (RLEF)~\citep{gehring2024rlefgroundingcodellms} does not use the meta-RL or search formulations, they do frame their approach as "learning from feedback", and it fits the Meta-CoT framework. Specifically, RLEF implements the E-RL$^2$ objective from equation~\ref{eq:ERL2} with an additional distributional constraint following standard RLHF practices~\citep{stiennon2022learning, ouyang2022training}. First, the model runs several iterations of exploration (episodes, which represent a full code solution), receiving compiler feedback from public test cases until it passes or reaches an exploration limit. Next, the model proposes a final solution (evaluation episode) and receives a reward based on hidden private test cases, which is used for RL training. See Figure~\ref{fig:rlef} for an overview of the process, along with an example of model outputs.~\citep{gehring2024rlefgroundingcodellms} shows a number of interesting empirical findings. The first finding is the relative effect of SFT versus RL training (results shown in Table~\ref{tab:sft-vs-rl}). The clear trend is that SFT training does not induce any in-context exploration capability and the does not lead to improved performance. This finding is consistent with the results from~\citet{kumar2024traininglanguagemodelsselfcorrect}, which carry out a similar analysis over multi-turn revisions of math problem solutions. 

\begin{figure}
    \centering
    \includegraphics[width=0.975\linewidth]{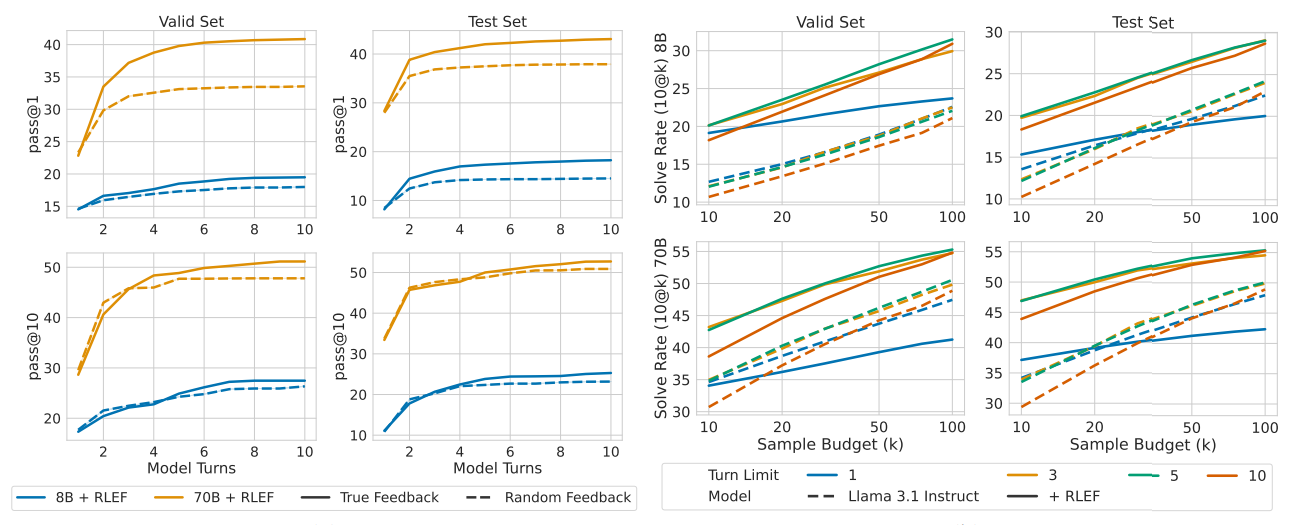}
    \caption{Scaling results for Reinforcement Learning with Execution Feedback. \textbf{Left}: Pass@1 and pass@10 for 8 and 70B models when given either ground truth feedback or random execution feedback. \textbf{Right}: Model solve rates at various turn limits (1, 3, 5, and 10) and sample budgets. Source: Figure 4 in~\citep{gehring2024rlefgroundingcodellms}.}
    \label{fig:rlef_scaling}
\end{figure}

Next, as shown in Figure~\ref{fig:rlef_scaling}, it is clear from the pass@1 results that the model is able to explore and internalize environment feedback to refine the code over multiple turns. However, more interestingly, it is able to do so \textit{without reliable feedback}. In particular the "random feedback" results replaces the compiler output with output from an unrelated problem. This likely creates significant issues for the model, since the feedback could throw off the LLM's grounding~\citep{mirzadeh2024gsmsymbolicunderstandinglimitationsmathematical}. However, we see that with RL tuning performance continues to improve with additional test-time revisions, although a gap with ground-truth feedback still exists. This is consistent with the findings from~\citet{kumar2024traininglanguagemodelsselfcorrect}, which demonstrate the capability to \textit{self-correct without external feedback} in the mathematical reasoning domain. Furthermore, these results are also consistent with our formulation of exploration in the epistemic POMDP from Section~\ref{sec:meta_RL}.

\subsection{Efficiency Or Super-Intelligence?}
As outlined in the beginning of Section~\ref{sec:internal_reasoning} two main reasons to internalize a search procedure within a single auto-regressive model are: (1) improved efficiency of the search procedure, and (2) the emergence of ``super''-intelligence.

\begin{figure}
    \centering
    \includegraphics[width=0.675\textwidth]{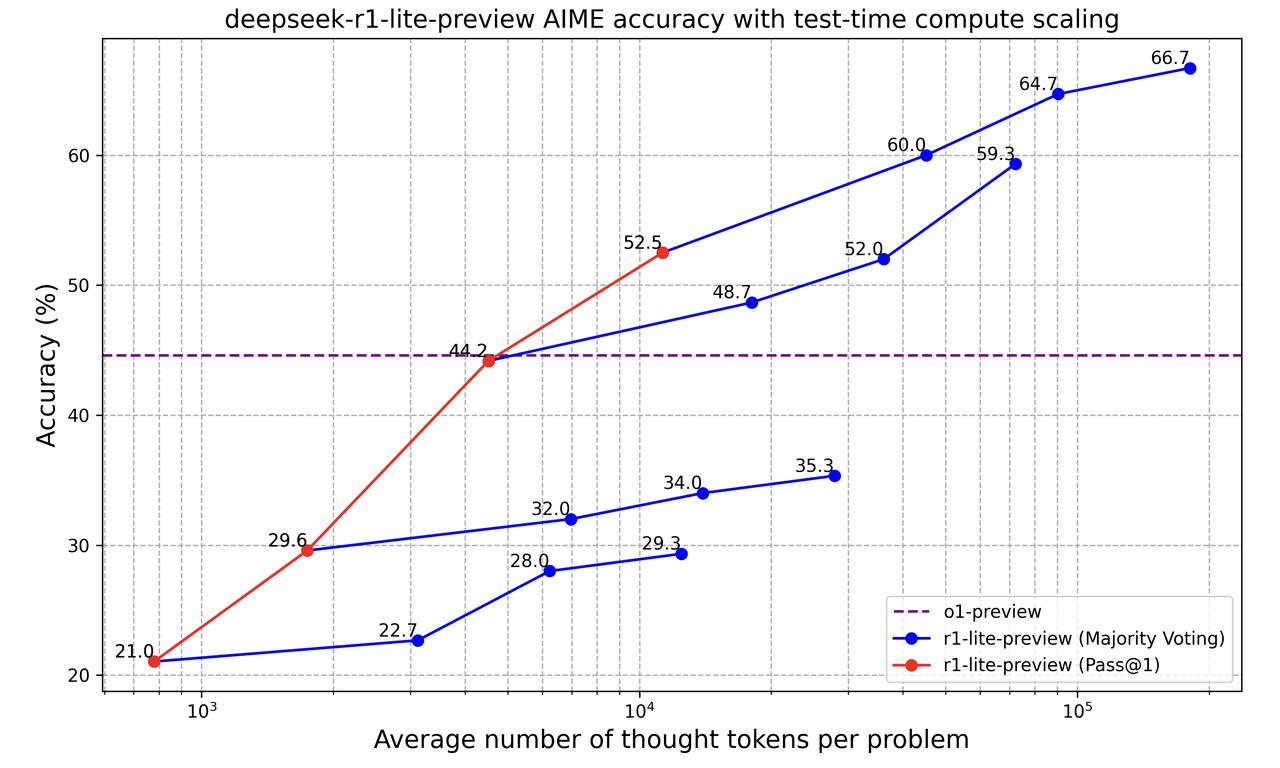}
    \includegraphics[width=0.315\textwidth]{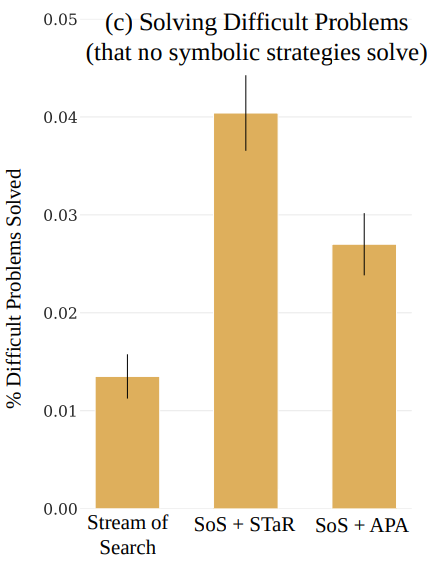}
    \caption{\textbf{Left}: Scaling laws of the R1 model. \textcolor{red}{Red:} different training checkpoints. \textcolor{blue}{Blue:} inference time scaling curves from each checkpoint. We see that the model uses inference budget more efficiently than parallel sampling with majority vote, but does not outperform large-scale parallel sampling at higher token budgets. Source:~\citep{deepseek2024r1}. \textbf{Right}: The percentage of problems which the SoS model solves on Countdown, but symbolic strategies do not. The biggest gain comes from RL tuning via STaR. Source: Figure 5c in~\citep{gandhi2024stream}.}
    \label{fig:SoS_difficult_problems}
\end{figure}

Results in Figure~\ref{fig:sos_results} show significant improvement from RL post-training in the limited domain of the Countdown game, specifically, success rates improve while also using smaller search budgets. However, even after RL post-training \textbf{the SoS model does not out-perform the success rate of the symbolic approach} which generated the instruction-tuning data. Similar findings have also been reported by~\citet{lehnert2024abetterplanningtransformers}, where the model achieves significant improvement in token efficiency, but not in success rates based on some simple post-training procedure.
Currently, it is unclear whether continued RL training can lead to a stronger model that substantially out-performs the modular search approach. An interesting analysis by~\citet{gandhi2024stream} (Figure~\ref{fig:SoS_difficult_problems}) shows the performance of various methods on difficult problems (i.e., problems that no modular symbolic search approach solves). The base SFT-tuned SoS model solves about 1\% of these problems, with RL-tuned models reaching up to 4\% (STaR in particular).

In Section~\ref{sec:meta_RL} we outlined the Meta-RL formulation, which stipulates that through RL post-training, we're essentially searching over algorithms, i.e. in-context adaptation procedures $U(\theta)$, rather than pure policies. In-theory, this process could discover novel reasoning strategies that unlock capabilities beyond the training data or manually designed reasoning approaches. 
Within the simple domain of Countdown, this does allow us to solve some complex problems that standard search methods do not, but this effect appears weak, even in this limited setting. As outlined earlier, it is unclear whether further RL training will allow us to discover \textit{novel} reasoning algorithms that allow us to solve \textbf{new classes} of problems. 
At larger scales, similar results seem to hold in the math~\citep{snell2024scalingllmtesttimecompute} and code-generation domains (with ground-truth environment feedback)~\citep{gehring2024rlefgroundingcodellms}. As shown in Figures~\ref{fig:sequential_sampling} and~\ref{fig:rlef_scaling} the E-RL$^2$ training clearly improves the search performance in terms of pass rate versus search budgets. However, it is unclear whether the RL post-trained model actually solves classes of problems that are unsolved by the base model under \textbf{increased search budget}. Indeed, for code generation, at the 70B scale, with $k=100$ the performance of the RL-tuned model is only about 5\% above the base model. At the same time, performance of the RL model appears to be saturating, while the base model demonstrates continued scaling. 
One potential counterpoint is the scaling law presented by~\citet{deepseek2024r1}, as shown in Figure~\ref{fig:SoS_difficult_problems} (left). In particular, we see that at low token limits the model demonstrates significantly better scaling with additional in-context search rather than parallel majority voting. On the other hand, at the higher end of token limits, majority voting can out-perform the model with in-context search, but at a higher inference cost. We should note that this approach uses solution-level majority voting, rather than a separate verifier, which has proven to be a much stronger search strategy~\citep{cobbe2021trainingverifierssolvemath, lightman2023let}.

\begin{tcolorbox}[colback=blue!5!white,colframe=blue!75!black]
In conclusion, based on public evidence, the proposed framework of in-context search can significantly improve the efficiency of the search procedure over base models and "symbolic" approaches or search strategies. However, there is only weak evidence that these models can discover novel reasoning methods that can solve classes of problems that were not solvable previously under some search budget.
\end{tcolorbox}

That is, under current empirical evidence the benefits outlined in point (1) above appear clear. However, evidence of emergent "super"-reasoning under point (2) appears weak.

\subsection{Can System 2 Reasoning Emerge From Pure RL?}
\label{sec:reasoning_emergence}
There is an opinion in the open-research community that the current generation of advanced reasoning models are artifacts of continual reinforcement learning~\footnote{https://www.interconnects.ai/p/openais-o1-using-search-was-a-psyop}. For both theoretical and practical reasons, we believe this is unlikely with the current generation of language models. First, outside of the OpenAI o1, DeepSeek R1, and Qwen QwQ models, which have undisclosed training routines, ``standard'' models do not exhibit such strong reasoning behaviors, despite the fact that they have already been extensively post-trained with instruction-tuning and reinforcement learning on reasoning tasks. From a more theoretical point-of-view, \textbf{meta-RL does not arise from standard RL}. Indeed, this is the main point of~\citet{ghosh2021generalizationrldifficultepistemic} and the remark in Section~\ref{sec:meta_RL} - models trained with RL on standard CoT formulation can exhibit arbitrarily bad performance on new problems.

We see this empirically in Figure~\ref{fig:rlef_scaling} (right). Note that setting the turn limit to 1 (solid blue line) corresponds to standard RL training (no in-context exploration episodes). In this setting the RL post-trained model \textbf{performs worse than the base model} (dotted blue line) \textbf{at higher levels of inference compute} (Sample Budget) on a held-out test set. Furthermore, models trained with even a turn limit of 3 show significantly better inference scaling over the base model, continuing well beyond their training budget.
Similar results were obtained by~\citet{kumar2024traininglanguagemodelsselfcorrect}, showing that a naive application of the RL$^2$ objective in Equation~\ref{eq:RL2} leads the model to collapse onto a greedy policy which does not perform in-context exploration (consistent with the findings of the original work~\citep{stadie2019considerationslearningexploremetareinforcement}).
On the first step, the two-episode formulation of SCoRe (method proposed by~\citet{kumar2024traininglanguagemodelsselfcorrect}) is identical to standard RL, which prevents the model from meaningful in-context exploration. To alleviate this issue they include an additional training stage using the E-RL$^2$ objective, similar to~\citet{gehring2024rlefgroundingcodellms}, before annealing into an RL$^2$ approach. These empirical observations are in line with prior results from meta-RL theory.
However, in the pure language setting, such as mathematical reasoning, the model uses an auto-regressive architecture with memory. Moreover, it essentially has full control of the environment and, in theory, can induce meta-behaviors such as backtracking and branching. In practical terms, there is no theoretical reason that the LLM cannot produce a complex sequence of tokens, such as the Meta-CoTs described earlier. We will investigate the presence of such artifacts next.

\begin{figure}[!htbp]
\centering
\includegraphics[width=0.99\linewidth]{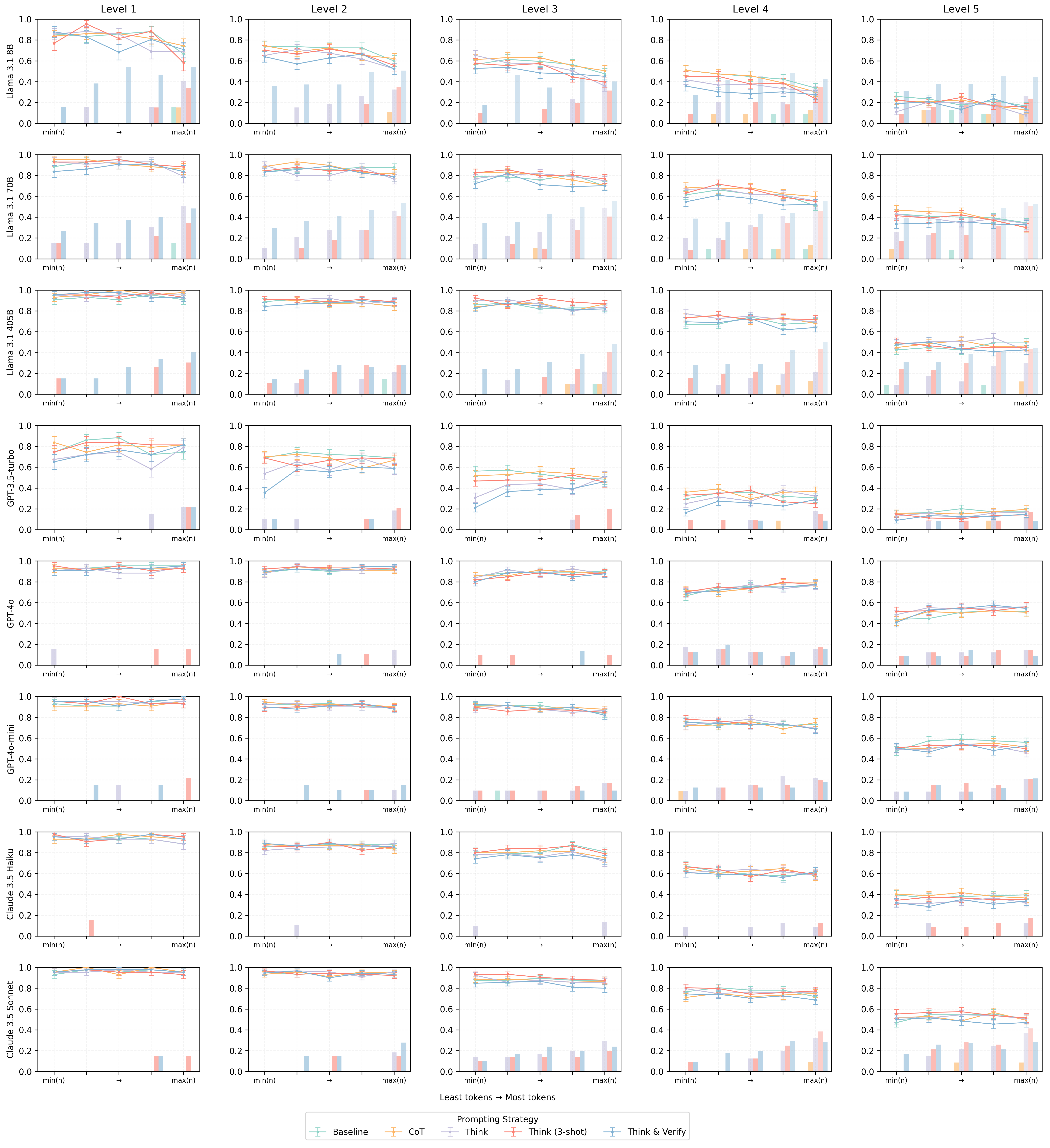}
\caption{Per-problem, token-ordered attempts (n=5) analysis of solution accuracy and self-correction behavior. Lines indicate accuracy trajectories; bars represent frequency of explicit error recognition.
}
\label{fig:error_analysis}
\end{figure}


\begin{figure}
\includegraphics[width=0.99\linewidth]{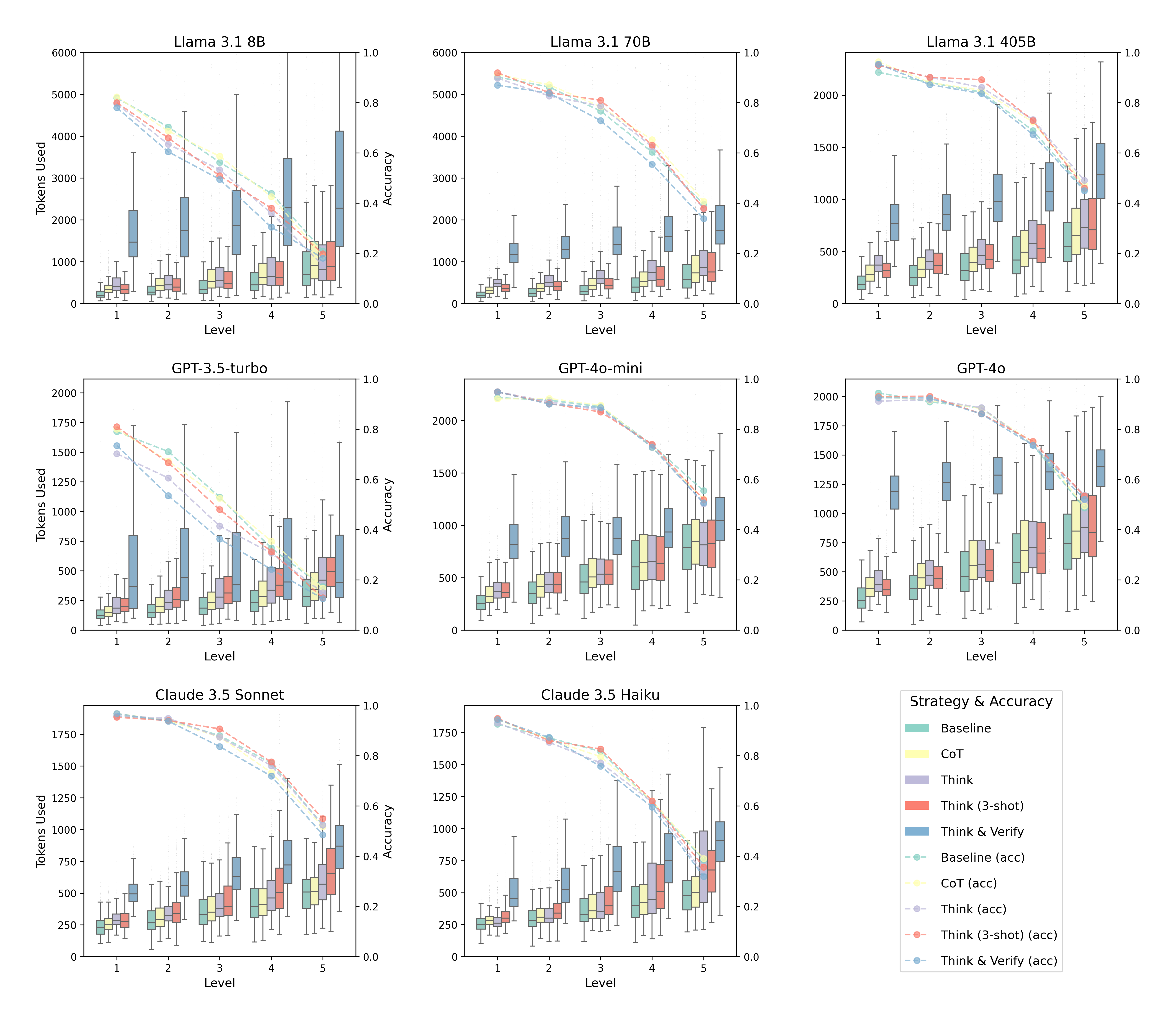}
\caption{\textbf{Boxplots of completion tokens generated and scatter plots of accuracy across MATH difficulty levels}, broken down by model and prompting strategy. Higher difficulty problems generally elicit longer generations. Box plots represent token distributions while dashed lines track accuracy. Although complex strategies (e.g. Think \& Verify) spend the most total tokens, the baseline prompt consistently scales at faster percentage gains from level-to-level while accuracy lines follow similar curves. 
}
\label{fig:token_analysis}
\end{figure}

\subsubsection{Inducing Meta-Reasoning In LLMs}\label{sec:in_context_meta_reasoning}

Prior works have shown that complex exploration and reasoning behaviors can be induced in LLMs through in-context demonstrations~\citep{sel2024algorithmthoughtsenhancingexploration, gandhi2023strategicreasoninglanguagemodels, nie2024evolveevaluatingoptimizingllms}. However, the degree to which such prompting can induce genuine meta-reasoning capabilities, particularly for complex reasoning, remains an open question. In this section, we investigate meta-reasoning induction through carefully constructed prompting strategies. We evaluate model performance on the MATH test set~\citep{hendrycks2021measuringmathematicalproblemsolving}. Our analysis examines three key dimensions: token generation patterns, expressions of meta-cognitive behaviors (such as explicit error recognition), and mathematical problem-solving accuracy. We evaluate five prompting strategies with progressively increasing complexity:

\begin{enumerate}
    \item The \textbf{Baseline} strategy implements minimal instruction, directing the model to assume mathematical expertise while providing basic formatting requirements.
    \item The \textbf{CoT} strategy extends the Baseline prompt by requesting explicit step-by-step reasoning and chain of thought verbalization. 
    \item The \textbf{Think} strategy implements a distinctive approach to mathematical reasoning by requesting \textit{stream-of-consciousness meta-cognition} within explicit structural constraints. This strategy aims to elicit authentic search by placing an inner monologue inside of a private ``thinking'' header, using natural language patterns (``Hmm'', ``Let me see...'', ``Because of this...'') that request self-verification and can demand explicit backtracking on identified errors. These design choices mimic the non-linear reasoning patterns of mathematicians approaching new problems. The three-shot variant (\textbf{Think 3-shot}) augments this with examples, using a static set of in-context demonstrations to illustrate desired behavior patterns.
    \item The \textbf{Think \& Verify} prompt introduces structured iteration bounds and verification requirements through dual constraints. The prompt mandates a second-pass verification for confident solutions while maintaining a soft maximum of six distinct solution attempts. Upon reaching a high-confidence solution, the protocol demands one additional verification attempt. Successful verification permits termination, though the model retains autonomy to continue exploration within the maximum bound should it self-report uncertainty. This adaptive termination criterion functions as an in-context best-of-N approach with self-consistency based early-stopping, optimistically balancing comprehensive solution space exploration with self-verification.

\end{enumerate}


We analyze expressions of regret and self-correction across models. We define regret as explicit acknowledgments of errors or uncertainty through phrases like "I made a mistake", "oops", "let me reconsider", or similar language. A detailed breakdown of regret statistics across different models and prompting strategies is provided in Table~\ref{tab:regret-statistics} in Appendix~\ref{app:regret_analysis}.

The regret analysis reveals a clear distinction between basic prompting approaches and those explicitly designed to encourage meta-cognitive behaviors. Under baseline and standard CoT prompting, which provide no explicit instruction for self-verification, models rarely express regret or acknowledge errors (<0.5\% of solutions across all models), showing that behaviors like self-verification and backtracking rarely emerge natively with standard prompting. In contrast, the Think strategy, which explicitly instructs models that they can "think, reflect, revise, backtrack, and verify during responding when it considers doing so would lead to a better solution," produces substantially higher rates of error recognition and correction, particularly in larger models like Llama 3.1 70B (12.65\%). This effect is further amplified when combining such instruction with structured examples (Think 3-shot) and explicit verification requirements (Think \& Verify), leading to significantly increased rates of regret expression across most models, peaking at 25.67\% for Llama 3.1 70B under the Think \& Verify strategy.

Interestingly, model scale correlates with willingness to express regret, but not monotonically. While larger Llama models show high rates of regret expression (15-25\% under Think \& Verify), more advanced models like GPT-4o and Claude 3.5 Sonnet exhibit notably lower rates (1-4\%).

Our main set of results is shown in Figure~\ref{fig:token_analysis}. First, to further understand how model behavior varies across difficulty levels and prompting strategies, we analyze the relationship between token usage and accuracy. The box plots show token distributions while overlaid dashed lines track accuracy across difficulty levels. Notably, more complex prompting strategies incur higher token overhead but do not consistently translate this additional computation into proportional accuracy gains, particularly in higher difficulty regimes.
For smaller models like Llama 3.1 8B, we observe significantly increased token usage in higher difficulty levels that corresponds with a marked decline in accuracy. In contrast, larger models maintain more consistent token usage patterns across difficulty levels.
The Think \& Verify approach consistently produces longer solutions across all models, reflecting the inherent overhead of verification steps. However, this increased verbosity does not translate to improved accuracy—particularly for smaller models where longer solutions may indicate joint struggles with problem-solving and verification. Moreover, we see significant increases in verbosity on all problem difficulties.
Of particular interest, we find that the Think \& Verify strategy leads to equivalent accuracies on the lowest difficulty problems (for all models), while often requiring more than double the token budget. \textbf{This observation indicates that models seek to match the reasoning "style" rather than substance and may be even faking mistakes to match the desired in-context behaviors}~\citep{gudibande2023falsepromiseimitatingproprietary}.

\begin{tcolorbox}[colback=blue!5!white,colframe=blue!75!black]
In conclusion, while sophisticated prompts successfully elicit reasoning-like behaviors absent from baseline approaches, these behaviors - including recognizing mistakes and backtracking - do not consistently yield performance benefits. In fact, when models engage in self-correction and backtracking, the final answers are more likely to be incorrect. This pattern, combined with the observation that token generation increases with MATH difficulty level across all strategies while accuracy trajectories remain similar, suggests fundamental limitations in using explicit meta-cognitive instruction to induce robust reasoning capabilities.
\end{tcolorbox}

\section{Putting It All Together - A Pipeline for System 2 Reasoning}
\label{sec:reasoning_pipeline}
So far, we have presented a theory of advanced reasoning capabilities based around search, as well as some early empirical findings. In this section we suggest an overall approach to training advanced reasoning models. Our proposal follows the overall structure of modern post-training, consisting of instruction-tuning and RL training~\citep{stiennon2022learning,ouyang2022training}.

\subsection{Instruction Tuning}
\label{sec:inst_tuning_pipeline}
In Section~\ref{sec:in_context_meta_reasoning}, we presented evidence that the current generation of models cannot induce effective meta-reasoning through in-context prompting, and may even exhibit misleading behaviors. Instead, we propose to begin the process through instruction-tuning with synthetic in-context search data, in the manner outlined in Section~\ref{sec:synth_meta_cot}.
We showed in Section~\ref{sec:in_context_meta_reasoning} that even advanced models rarely demonstrate meta-reasoning capabilities, such as expressing regret or backtracking.
Hence, we believe an instruction-tuning stage is critical to endow the model with such backtracking and branching capabilities, which are not frequently present in the pre-training corpus or general purpose instruction-tuning data. Following the synthetic data approach in Section~\ref{sec:synth_meta_cot}, we construct a training dataset $\mathcal{D}_{\text{train}} = \{\mathbf{q}^{(i)}, \mathbf{Z}^{(i)}, \mathbf{S}^{(i)}\}_{i=1}^N$ which represents the Meta-CoT $\mathbf{Z}=\mathbf{z}_1, \ldots, \mathbf{z}_K$, followed by verifiable solution $\mathbf{S}=\mathbf{s}, \ldots, \mathbf{s}_n$. For this stage of training, multiple training objectives can be considered, which we outline in Appendix~\ref{app:inst_tuning_objectives}. To what degree each of these different objectives (and combinations thereof) yield qualitatively different behaviors is an open empirical question. However, based on previous results at the scale of modern LLMs, we hypothesize that \textbf{the exact format of the pre-training stage is not crucial, rather the key performance contributor is RL post-training}~\citep{ye2024physicslanguagemodels22,kumar2024traininglanguagemodelsselfcorrect, gehring2024rlefgroundingcodellms}.

\subsection{Post-Training With RL}\label{sec:RL_training}
Prior works have shown strong empirical results with multi-turn RL training~\citep{kumar2024traininglanguagemodelsselfcorrect, gehring2024rlefgroundingcodellms}. We propose using a similar objective based on the E-RL$^2$ approach outlined in Equation~\ref{eq:ERL2} with the standard additional distributional constraints:
\begin{equation}\label{eq:rl_post_training}
    \max_{\theta}\mathbb{E}_{\mathbf{S}, \mathbf{Z}\sim \pi_{\theta}(\cdot|\mathbf{q}), \mathbf{q}\sim \mathcal{D}_{\text{train}}}\left[r^*(\mathbf{S, \mathbf{q}}) - \beta\sum_{t}\mathbb{D}_{KL}[\pi_{\theta}(\mathbf{z}_{t+1}|\mathbf{Z}_{t}, \mathbf{q})||\pi_{\text{ref}}(\mathbf{z}_{t+1}|\mathbf{Z}_{t}, \mathbf{q})]\right]
\end{equation}
where $r^*$ is the verifiable reward from  the solution and $\pi_{\text{ref}}$ is a reference policy (usually the instruction-tuned model). Similar objectives have been considered in agentic applications where the "Meta-CoT" represents an actual search over a web interface~\citep{nakano2022webgpt, putta2024agent}. Note that the reference constraint here is not strictly necessary, but likely required to keep the chain stable and interpretable. Keeping the RL process stable over long horizons and learning robust credit assignment is likely a significant challenge. There are a number of possible options for dealing with these challenges:

\begin{enumerate}
    \item The step-wise branching structure of MCTS training presents one possible resolution. For example, by using an objective similar to that proposed by~\citet{feng2024alphazerolike}, who perform an SFT policy distillation approach using MCTS. Although in general SFT-based policy optimization tends to be inefficient in language model settings~\citep{tajwar2024preference}.
    \item A number of step-level DPO objectives have been proposed, specifically for reasoning applications, which can be combined with tree-search based exploration~\citep{xie2024montecarlotreesearch, setlur2024rlincorrectsyntheticdata,lai2024stepdpostepwisepreferenceoptimization}. We believe this might be a promising direction, as DPO-based approaches also tend to be more robust to off-policy data~\citep{noukhovitch2024asynchronousrlhffasterefficient} than comparable policy-gradient based approaches~\citep{shao2024deepseekmathpushinglimitsmathematical}, which allows for massive asynchronous RL scaling. 
    \item A branching version of on-policy methods could also be used, such as a step-level version of RLOO~\citep{ahmadian2024back} or VinePPO~\citep{havrilla2024teachinglargelanguagemodels,kazemnejad2024vineppounlockingrlpotential}, although these methods might not be as robust to off-policy data as prior options and hence less scalable~\citep{noukhovitch2024asynchronousrlhffasterefficient}.
\end{enumerate}

The exact correlations between instruction-tuning, pre-training, and RL are unknown in the public literature. Smaller scale studies have found SFT tuning to be effective in endowing transformers with meta-reasoning behaviors~\citep{yang2022chainthoughtimitationprocedure, lehnert2024abetterplanningtransformers, gandhi2024stream, ye2024physicslanguagemodels22}, while larger scale studies indicate that purely training with next-token prediction on synthetic traces might not be enough due to train-test distribution shifts and find that on-policy RL is the key component for improving performance~\citep{zhang2024backtrackingimprovesgenerationsafety, kumar2024traininglanguagemodelsselfcorrect, gehring2024rlefgroundingcodellms}.

\subsubsection{Q* or q-STaR?}
\begin{figure}
    \centering
    \includegraphics[width=0.485\linewidth]{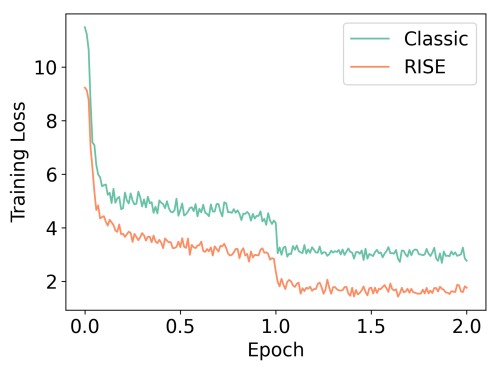}
    \includegraphics[width=0.485\linewidth]{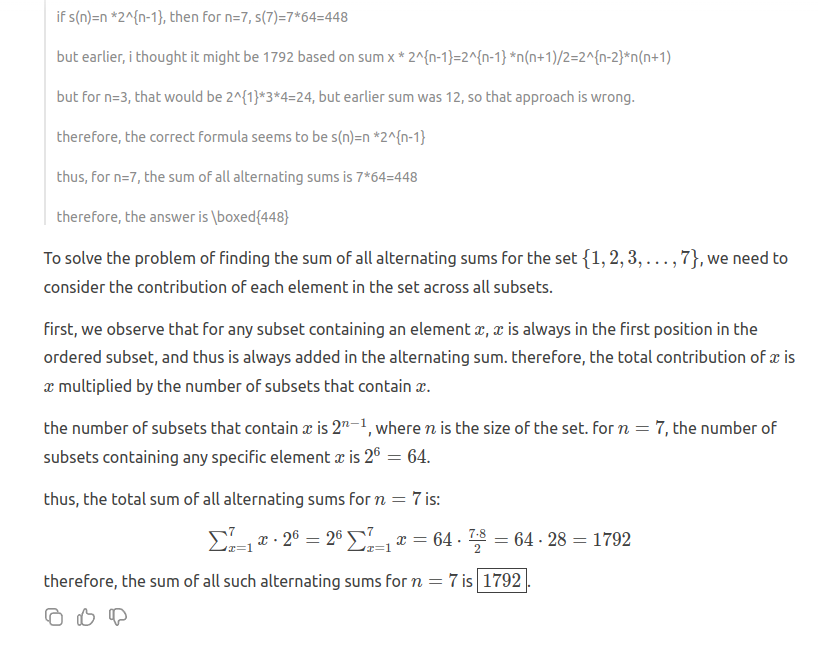}
    \caption{\textbf{Left}: The training loss of fitting only the oracle answer or a sequence of answers with the revision objective in Equation~\ref{eq:revision_loss}. The model that uses in-context exploration is better able to model the correct solution than directly outputting it after the question, consistent with our complexity considerations in Section \ref{sec:meta_cot}. Source: Figure 5 in \cite{qu2024recursiveintrospectionteachinglanguage}. \textbf{Right}: An example in which the DeepSeek R1 model \cite{deepseek2024r1} reaches the correct final solution (448) in its Meta-CoT $\mathbf{Z}\sim\pi_{\theta}(\cdot|\mathbf{q})$, but the summary model $\mathbf{S}\sim\pi_{\theta}(\cdot|\mathbf{Z}, \mathbf{q})$ fails and produces the wrong final solution (1792).}
    \label{fig:perplexity-loss}
\end{figure}
In the prior section, we introduced the RL training objective in Equation~\ref{eq:rl_post_training} using a verifier reward $r^*(\mathbf{S}, \mathbf{q})$. Here we will derive another RL formulation with a self-training type of RL objective without the need for an external verifier. 

Under the (Meta-)CoT formulation of Section \ref{sec:meta_cot}, we can consider the Meta-CoT process $\mathbf{Z}$ as a general latent variable model for the whole problem solution $\mathbf{S}$. We can then apply a widely known formulation of the STaR algorithm as latent inference \cite{singh2024humandatascalingselftraining}:
\begin{align}
    \log\pi_{\text{data}}(\mathbf{S}|\mathbf{q}) = \log\int\pi(\mathbf{S}|\mathbf{Z}, \mathbf{q})\pi(\mathbf{Z}|\mathbf{q})d\mathbf{Z} \geq \\
    \max_{q(\mathbf{Z}|\mathbf{q})}\mathbb{E}_{q(\mathbf{Z}|\mathbf{q})}[\log\pi(\mathbf{S}|\mathbf{Z}, \mathbf{q})] + \mathbb{D}_{KL}[q(\mathbf{Z}|\mathbf{q})||\pi(\mathbf{Z}|\mathbf{q})]
\end{align}
where $q(\mathbf{Z}|\mathbf{q})$ is a variational inference function \cite{kingma2013auto}, or in our case a latent reasoner. Since all components here are auto-regressive transformers (LLMs), we can amortize them inside a single model. Now, if we set $\pi(\mathbf{Z}|\mathbf{q})$ to be the prior model $\pi_{\text{ref}}$, which is initialized as the instruction-tuned base model ($\pi_{\theta_0}$) from Section \ref{sec:inst_tuning_pipeline} and amortize the reasoning inference model $q(\mathbf{Z}|\mathbf{q})$ and the decoder model $\pi(\mathbf{S}|\mathbf{Z}, \mathbf{q})$ into a single LLM $\pi_{\theta}$ we get the objective:
\begin{equation}
    \max_{\theta}\mathbb{E}_{\mathbf{Z}\sim \pi_{\theta}(\cdot|\mathbf{q}), \mathbf{S},\mathbf{q}\sim \mathcal{D}_{\text{train}}}\left[\log\pi_{\theta}(\mathbf{S}|\mathbf{Z}, \mathbf{q}) - \beta\mathbb{D}_{KL}[\pi_{\theta}(\mathbf{Z}, |\mathbf{q})||\pi_{\text{ref}}(\mathbf{Z}| \mathbf{q})]\right]
\end{equation}
where we used the $\beta-$VAE formulation \cite{higgins2017beta}. Unlike standard VAEs though, the parametric models here are auto-regressive transformers, which sample discrete tokens, hence we cannot use the reparameterization trick to compute gradients of the above distribution and must result to RL optimization. This objective indeed looks similar to the  main RL objective from the previous section, as formulated in Equation~\ref{eq:rl_post_training}, but with a few key differences. First, the reward function is represented as \begin{equation}\label{eq:qSTaR_reward}
r(\mathbf{S}, \mathbf{q})=\log\pi_{\theta}(\mathbf{S}|\mathbf{Z}, \mathbf{q})
\end{equation}
where the solution $\mathbf{S}$ is no longer sampled on-policy from the model, but from a pre-existing question-solution training dataset. Hence the sampling expectation from the policy is \textbf{only over the latent Meta-CoT}. Notice also that the reward itself is a function of the model parameters $\theta$, which then requires a modification to the standard policy gradient approach. With some simple differentiation-by-parts calculus, the above objective can be represented as:

\begin{align}\label{eq:rl_q_STaR}
    & \max_{\theta}\mathbb{E}_{\mathbf{Z}\sim \pi_{\theta}(\cdot|\mathbf{q}), \mathbf{S},\mathbf{q}\sim \mathcal{D}_{\text{train}}}\left[\textbf{sg}(\log\pi_{\theta}(\mathbf{S}|\mathbf{Z}, \mathbf{q})) - \beta\mathbb{D}_{KL}[\pi_{\theta}(\mathbf{Z}, |\mathbf{q})||\pi_{\text{ref}}(\mathbf{Z}| \mathbf{q})]\right] + \\
    & \max_{\theta}\mathbb{E}_{\mathbf{Z}\sim \textbf{sg}(\pi_{\theta}(\cdot|\mathbf{q})), \mathbf{S},\mathbf{q}\sim \mathcal{D}_{\text{train}}}[\log\pi_{\theta}(\mathbf{S}|\mathbf{Z}, \mathbf{q})]\nonumber
\end{align}

Here the first equation is now a standard RL problem with reward as defined in Equation~\ref{eq:qSTaR_reward} which can be optimized with the standard methods, outlined in Section \ref{sec:RL_training}. The second part is a maximum likelihood training of the Meta-CoT "summarization" model on the ground-truth solution. Essentially the goal of this training objective is to make the conditional model $\pi_{\theta}(\mathbf{S}|\mathbf{Z}, \mathbf{q})$ more expressive than the standard training objective $\pi_{\theta}(\mathbf{S}|\mathbf{q})$. Some empirical evidence from \cite{qu2024recursiveintrospectionteachinglanguage} supports this hypothesis as shown in Figure~\ref{fig:perplexity-loss} (left) which shows training curves for the revision objective in Equation~\ref{eq:revision_loss}. Indeed, even with standard SFT using in-context exploration the model achieves significantly lower loss than directly predicting the correct answer. This is similar in spirit to the goal of Quiet-STaR \cite{zelikman2024quiet}, with the difference being that we optimize and marginalize over the entire latent meta-reasoning process and final solution with arbitrary complexity. In comparison Quiet-STaR applies the above objective at a token-level with fixed token budget for the latent process, thus essentially making the transformer computation graph deeper, similarly to Universal Transformer \cite{dehghani2019universaltransformers}, rather than optimizing meaningful semantic reasoning. This line of reasoning is also consistent with the complexity arguments outlined in Section \ref{sec:meta_cot}.

Finally, we should note that while this approach does not utilize verifiers for RL training, it still needs to bootstrap the latent reasoning processes $\mathbf{Z}$, which still requires verifiable outcomes to generate high-quality synthetic training data. 

The objective in Equation~\ref{eq:rl_q_STaR} has one major advantage that we no longer require verification of the final solution, since the training objective only requires a dataset of question-solution pairs. This allows us to train on open-ended and hard to verify problems, such as proofs and general reasoning. The downside is now that the model $\pi_{\theta}(\mathbf{S}|\mathbf{Z}, \mathbf{q})$ is only essentially trained with supervised fine-tuning and does not get on-policy reward feedback, which may be sub-optimal. If the Meta-CoT process can find a solution with high certainty, then perhaps an SFT training objective for the summarization model is sufficient, which is an empirical question. Anecdotally, we have observed instances of the DeepSeek R1 model \cite{deepseek2024r1} finds the correct answer in its Meta-CoT but actually outputs the wrong final solution as shown in Figure~\ref{fig:perplexity-loss} (right). 

\subsubsection{Discount Rates}
\begin{figure}
\centering
\includegraphics[width=0.650\linewidth]{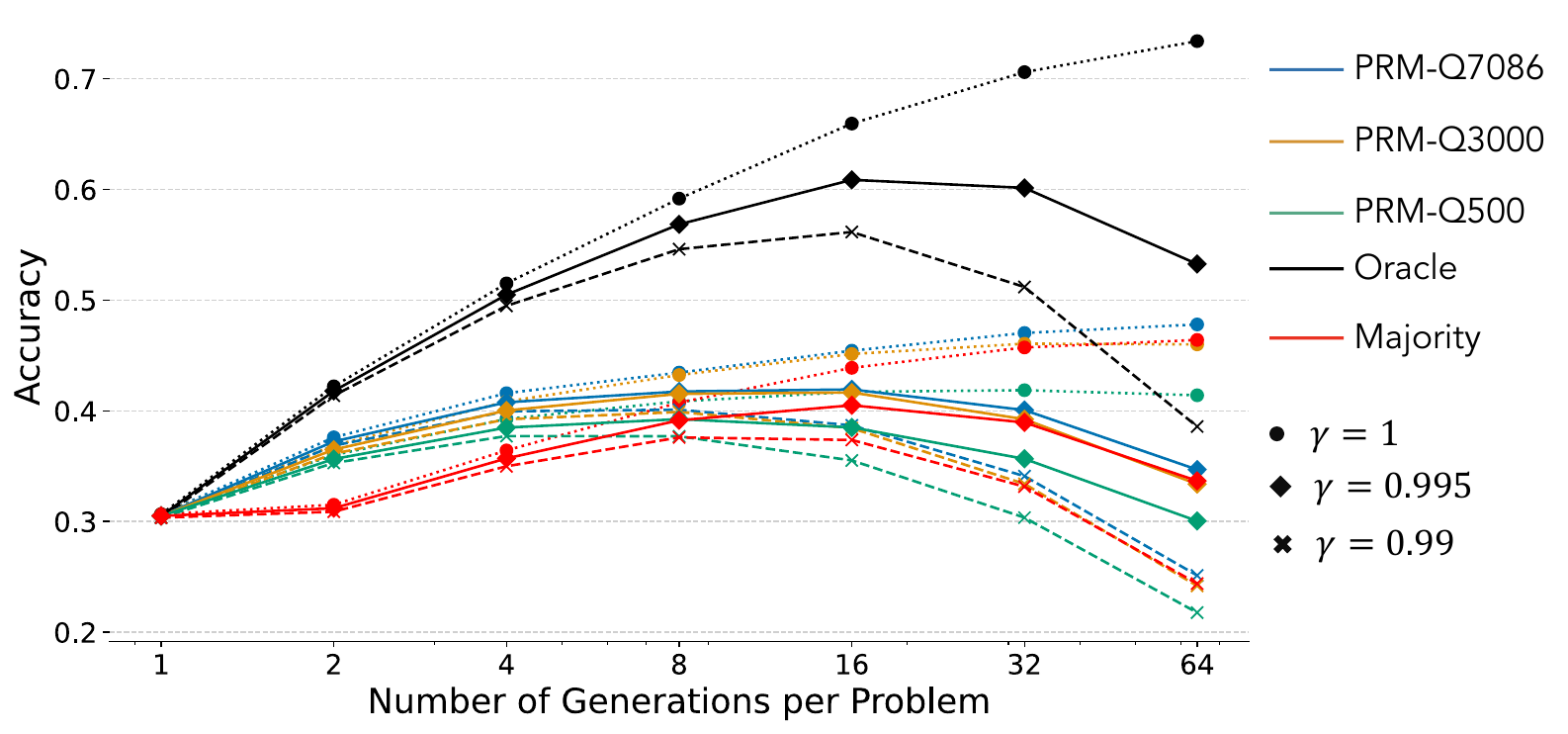}
\includegraphics[width=0.325\textwidth]{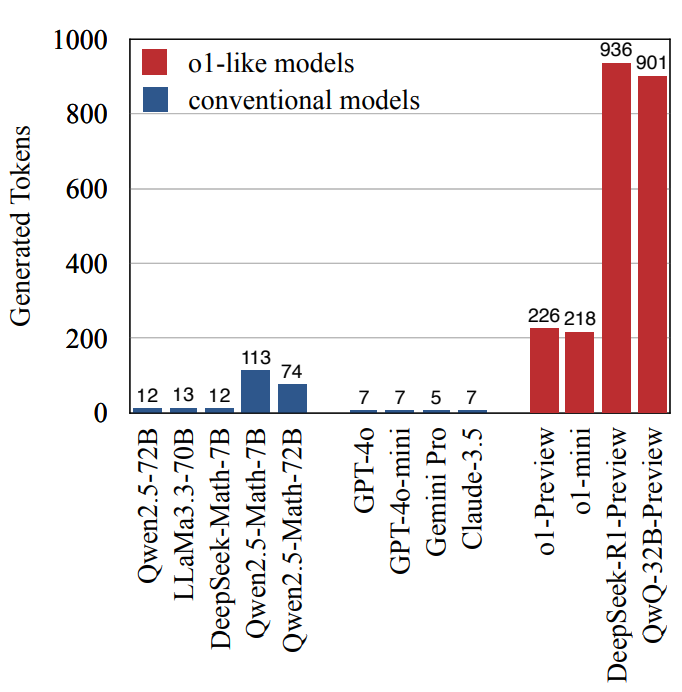} 
\caption{\textbf{Left}: Figure \ref{fig:prm_scaling_curve} with discounted objective. The undiscounted version of Best-Of-N sees continued improvement with additional sampling, while the discounted objective exhibits a hump-shaped frontier. \textbf{Right}: Token lengths for answers of "what is 2+3=?" by different models. Source: Figure 1 in \cite{chen2024think23overthinkingo1like}.}
  \label{fig:reasoning_max_token_lengt}
\end{figure}
While standard RLHF pipelines have not used discount rates in the past, they may be required in reasoning applications. Recent works have discovered that small biases in preferences for longer answers are routinely exploited by reward models in RL pipelines, which has yielded significantly more verbose models~\citep{singhal2023long, park2024disentangling}. These issues are present even in strong systems, such as GPT-4~\citep{gpt4} and require explicit regularization. As demonstrated in prior chapters, when increased sampling correlates with higher accuracy then, without explicit regularization, the model can choose to continue generating/searching for solutions or collapse on some majority voting approach with potentially unlimited inference targets. Indeed, recent work \cite{chen2024think23overthinkingo1like} found that advanced reasoning models can generate significantly longer sequences even for simple problems ("What is 2+3=") as shown in Figure \ref{fig:reasoning_max_token_lengt}. The Qwen QwQ model \citep{qwq-32b-preview} generates up to 13 solutions in context before providing a final answer.
To mitigate this issue, we might require a modification of the RL objective from Equation~\ref{eq:rl_post_training}, to use a discounted objective
\begin{equation}\label{eq:discounted_rl_post_training}
    \max_{\theta}\mathbb{E}_{\mathbf{S}, \mathbf{Z}\sim \pi_{\theta}(\cdot|\mathbf{q}), \mathbf{q}\sim \mathcal{D}_{\text{train}}}\left[\gamma^{|\mathbf{Z}|}r^*(\mathbf{S}, \mathbf{q}) - \beta\sum_{t}\mathbb{D}_{KL}[\pi_{\theta}(\mathbf{z}_{t+1}|\mathbf{Z}_{t}, \mathbf{q})||\pi_{\text{ref}}(\mathbf{z}_{t+1}|\mathbf{Z}_{t}, \mathbf{q})]\right]
\end{equation}
fiwhere $\gamma<1$. This objective would (in theory) force the model to trade-off compute versus certainty in its response. This effect is demonstrated in Figure~\ref{fig:reasoning_max_token_lengt} (left), which shows a discounted version of the best-of-N objective. Without a particular penalty on the generation budget, performance increases steadily with additional compute, which may push the model to continuously increase the inference budget (as compared to base models) as shown on the right. If we consider the discounted reward (accuracy) instead, we see a hump-shaped objective as there is a stronger trade-off between verifier confidence and the generation budget.

This also raises an interesting trade-off on controlling model behaviors in terms of inference/accuracy. We might be interested in forcing a particular trade-off depending on problem difficulty. Consider then a distribution over discount rates $\gamma_1< \ldots < \gamma_m$. We can also associated an individual language prompts $\mathbf{p}_{\gamma}$ with each discount rate such as "Work as quickly as possible" for high discount, to "Take as much time as needed" for a low discount rate. Then, at train time we can optimize the objective
\begin{equation*}\label{eq:rl_post_training_discounted_prompt}
    \max_{\theta}\mathbb{E}_{\mathbf{S}, \mathbf{Z}\sim \pi_{\theta}(\cdot|\mathbf{q}, \mathbf{p}_{\gamma_i}), \mathbf{q}\sim \mathcal{D}_{\text{train}}, i\sim 1:m}\left[\gamma_i^{|\mathbf{Z}|}r^*(\mathbf{S}, \mathbf{q}) - \beta\sum_{t}\mathbb{D}_{KL}[\pi_{\theta}(\mathbf{z}_{t+1}|\mathbf{Z}_{t}, \mathbf{q}, \mathbf{p}_{\gamma_i})||\pi_{\text{ref}}(\mathbf{z}_{t+1}|\mathbf{Z}_{t}, \mathbf{q})]\right]
\end{equation*}
Then, at inference time we can control the qualitative model behavior through conditioning on the prompt to trade-off inference compute versus accuracy.~\citet{schultz2024masteringboardgamesexternal} showed that a if we train a model on MCTS search traces with different parameters, at inference time we can condition the model on a particular parameter configuration and recreate the qualitative behaviors(as shown in Figure~\ref{fig:SoSInferenceScalling}). As discussed earlier, simple supervised fine-tuning may not be sufficient to induce advanced reasoning capabilities (and likely even more difficult to do so in a controllable way), however, these results demonstrate promise in inducing controllable behaviors into the model through the conditional prompting objective above.

\section{Going Forward}
Public open-research/source progress on reasoning models is currently bottle-necked by three main issues:

\begin{enumerate}
    \item \textbf{Access to resources} in terms of both \textit{data and compute} as these algorithms require significant amount of both.
    \item \textbf{Open-source infrastructure} for large scale inference and training is currently lackluster.
    \item \textbf{Algorithmic exploration} - we have a lot of avenues to explore and only limited people and resources actively working on the right directions. 
\end{enumerate}

We expand on these issues in this section.

\subsection{The "Big MATH" Project}
\label{sec:dataset}

While compute bottlenecks are a persistent issue in open research, we found the lack of open datasets with verifiable reasoning problems to be an even bigger challenge. Prior works have used the GSM8k~\citep{cobbe2021trainingverifierssolvemath} and MATH~\citep{hendrycks2021measuringmathematicalproblemsolving} datasets, but the first has largely been saturated by the current generation of models and the second is quite limited, with only 12,000 problems. To overcome these challenges, we have put significant work into the "Big MATH" project - an effort to aggregate over 1,000,000 high-quality, high-confidence, and diverse verifiable math problems. We combine existing datasets with significant post-processing, as well as efforts to acquire additional data from novel sources. We outline this below. 

Our proposed training pipeline requires a large-scale corpus of challenging prompts with verifiable answers, but developing such datasets presents fundamental constraints in automated verification and assessment of reasoning capabilities.
In many domains, solutions are too unstructured or nuanced for automated verification: a single objective answer may not exist, correct solutions can appear in equally valid but textually distinct forms, and certain tasks (e.g.\ complex proofs) resist reduction to deterministically verifiable outputs with a single canonical representation. These issues make automated grading infeasible. Even in contexts where problem-answer pairs can be constructed and easily verified, these formats may reduce to factual retrieval rather than exercises that require reasoning capabilities.

As evidenced by the wide array of literature discussed in previous sections, mathematics offers a stable ground truth for correctness, naturally programmatic answer grading, and existing public datasets in problem-solution-answer format. While we recognize significant potential in many domains (e.g., programming, diagnostics, finance, analysis) where structured reasoning paths can be validated, the substantial resources required to construct and verify such datasets, particularly without access to proprietary data, makes them impractical for our immediate objectives. While the availability of existing datasets ultimately drove our domain selection, even these openly available resources are constrained in both scale and distribution. We face the critical challenge of distinguishing between computational and pattern proficiency and conceptual understanding when assessing model capabilities - a distinction that fundamentally shapes our dataset requirements and evaluation metrics.


To guide our data construction, we define three core criteria: 
\begin{enumerate}
    \item the existence of \textbf{uniquely verifiable solutions}, meaning that problems must admit a single correct answer that can be reliably verified;
    \item \textbf{open-ended problem formulations}, ensuring that tasks cannot be easily solved by guessing (as might occur in multiple-choice formats) and instead require nontrivial reasoning steps; and
    \item \textbf{closed-form solutions}, such that the final answer must be expressible in a closed form (e.g., a scalar or formula, not a proof), thereby enabling automated evaluation.
\end{enumerate}


These criteria reveal significant limitations in existing datasets. Those that meet our structural requirements remain severely limited in scale, containing orders of magnitude fewer examples than we estimate necessary for model training. Within the structured collections, we observe a concerning scarcity of non-trivial problems that effectively challenge reasoning capabilities, with many examples simply testing computational abilities or following predictable patterns. Perhaps most problematic is the non-negligible proportion of incorrect problem-solution labels, even in widely-used datasets, introducing substantial complexity to data cleaning.

Preliminary analysis of one of the largest available datasets, NuminaMath~\citep{numina_math_datasets}, exemplifies these issues. Basic verification shows that out of roughly 860,000 entries, more than 42,500 are duplicates ($\sim$5\% of the dataset). Further, while 89.7\% of entries contain exactly one clearly boxed solution suitable for automated verification, 2.6\% contain no boxed solution, and 7.7\% include multiple boxed solutions. Problems like this underscore the state of open and publicly available data and suggest deeper quality concerns.

\subsubsection{Data Sourcing}
We consider multiple established mathematical problem datasets that are commonly used in the literature (Table~\ref{tab:data_sources}).
First, we include the Human Annotated Reasoning Problems (HARP) dataset~\citep{yue2024harp}, containing nearly 4,800 competition-level short answer problems with programmatically-checkable answers.
Next, we use the NuminaMath~\citep{numina_math_datasets} dataset, which is composed of roughly 860,000 problems from a variety of benchmarks and sources: Chinese high school math exercises, math olympiad-style competition problems, the art of problem solving forum, MATH~\citep{hendrycks2021measuringmathematicalproblemsolving}, and GSM8k~\citep{cobbe2021trainingverifierssolvemath}.
NuminaMath further incorporates synthetic data from the synthetically generated dataset Orca-Math~\citep{mitra2024orcamathunlockingpotentialslms} as well as further generating a significant amount of synthetic data based on the MATH dataset and a subset of the math competition problems.
Additionally, while NuminaMath uses the original split of the MATH dataset (7,000 training problems, 5,500 test problems), we choose to use version with 12,000 training problems and 500 test problems, as originally proposed by~\citep{lightman2023let}.
We incorporate the Omni-MATH dataset~\citep{gao2024omnimathuniversalolympiadlevel}, contributing almost 4,500 olympiad-level problems curated from 39 different competition websites. Notably, this dataset employs professional annotators and verifiers to maintain solution-answer quality.
Finally, we include OpenMathInstruct-2~\citep{toshniwal2024openmath2}, composed of about 607,000 synthetically generated problems.~\citet{toshniwal2024openmath2} use Llama3.1-405B~\citep{dubey2024llama3herdmodels} in a multi-step data augmentation pipeline which they seed using the GSM8k~\citep{cobbe2021trainingverifierssolvemath} and MATH~\citep{hendrycks2021measuringmathematicalproblemsolving} datasets.

Beyond drawing from these established datasets, we maintain an active data collection initiative that extends our corpus through manual curation of competition-level problems. This effort has already contributed thousands of additional olympiad-style problems and continues to grow. Our manual collection process focuses on acquiring mathematically rigorous content from competition archives, olympiad materials, and historical sources not present in existing datasets.


\subsubsection{Data Filtering}
To create datasets with which we can train a model in our training pipeline, we next clean and filter the data from each source using a combination of bespoke and common strategies~\citep{albalak2024surveydataselectionlanguage}.
Specifically for our two-stage pipeline, we create two versions of the dataset. First, we have the \textbf{base filter}, where the data will be used during SFT for the model to learn (1) an approximation of the distribution of math problems and (2) to follow the desired chain-of-thought format. Next, the \textbf{strict filter} is intended to be used for the Meta-RL training stage. Thus, the strictly filtered data should contain only problems that closely follow our three desired properties: open-ended, verifiable, closed-form problem-solution pairs.

\begin{table}[t]
\centering
\begin{tabular}{|l|r|r|r|}
\hline
 \textbf{Data Source}& \textbf{Original} & \textbf{Base Filter} & \textbf{Strict Filter} \\
\hline
HARP~\citep{yue2024harp} & 4,780 & 3,691 & 2,996 \\
NuminaMath~\citep{numina_math_datasets} & 859,608 & 452,820 & 231,887 \\
Omni-MATH~\citep{gao2024omnimathuniversalolympiadlevel} & 4,428 & 3,660 & 2,478 \\
OpenMathInstruct-2~\citep{toshniwal2024openmath2} & 607,324 & 600,191 & 496,331 \\
\hline
\textbf{Total} & 1,476,140 & 1,060,362 & 733,692 \\
\hline
\end{tabular}
\caption{Comparison of Problems by Data Source and Filter Type}
\label{tab:data_sources}
\end{table}

The \textbf{Base Filter} performs some operations that are unique to each subset, and some operations that are shared across the full collection of data.
First, in the HARP dataset~\citep{yue2024harp}, we find many problems that contain figures in the Asymptote\footnote{\url{https://asymptote.sourceforge.io/}} vector graphics language, which we filter out of the dataset.
Next, we filter the NuminaMath dataset~\citep{numina_math_datasets}. This dataset contains a significant amount of synthetic data, which is often difficult to verify for accuracy and correctness. However, the Orca-Math~\citep{mitra2024orcamathunlockingpotentialslms} subset has empirically been proven to improve performance on supervised fine-tuning, and for this reason, we elect to maintain only the Orca-Math data, while discarding the other synthetic subsets. Next, NuminaMath does not explicitly contain answers to each problem, so we extract answers to problems by searching for boxed solutions (``\textbackslash boxed\{\}'' in LaTeX). Any problems whose solution does not contain exactly 1 boxed answer is filtered out.
When exploring Omni-MATH~\citep{gao2024omnimathuniversalolympiadlevel}, we found many problems containing author attributions (e.g.\ a person's name in parenthesis) and removed the attributions from the problem.
OpenMathInstruct-2 originally contains multiple solutions and final answers to each problem~\citep{toshniwal2024openmath2}. The first step we take is to group all matching problems together, removing those which have mismatched expected answers. Next, we found that the data still contains evidence of the synthetic data generation process, such as asking ``do you want to solve it?'' or ``here's the new problem'' and remove any problems with similar phrases.

Finally, after running each of the described filters over the individual subsets, we perform four filtering operations across the full collection. First, we use exact matching to find and remove duplicate problems. Then, we use a FastText language identifier~\citep{joulin2016bag, joulin2016fasttext, grave2018learning} and remove any problems where English is not the primary language. Next, we remove problems containing a hyperlink, as this suggests that a model may not have the full resources required to solve the problem. Lastly, we decontaminate the dataset by removing any examples of the MATH500 test set~\citep{lightman2023let}.

The \textbf{Strict Filter} further reduces the base filtered data to problem-answer pairs which are more likely to be open-ended, verifiable, and closed-form.
The strict filters are applied equally to all subsets of the dataset through a mix of rule-based and model-based filtering.
First, we remove questions containing multiple parts as these can be challenging to evaluate. For a similar reason, we also remove questions that ask for a proof as these problems are difficult to evaluate. Next, we choose to remove multiple choice problems as models have a high probability of selecting the correct answer without producing an accurate reasoning chain. For the same reason, we also remove Yes/No and True/False problems, as these may give a poor learning signal during Meta-RL training. Finally, we use the SemDeDup algorithm~\citep{abbas2023semdedupdataefficientlearningwebscale} with the model at \href{https://huggingface.co/sentence-transformers/all-MiniLM-L6-v2}{sentence-transformers/all-MiniLM-L6-v2} and remove problems with a cosine similarity over 0.5.

\subsection{Infrastructure}
For the proposed family of algorithms we need RL training infrastructure that can:

\begin{enumerate}
    \item Scale to multiple nodes in a straightforward way
    \item Allow for high-performance inference throughput
    \item Allow for interleaving inference and training efficiently for online RL algorithms
\end{enumerate}

Here we outline our recent progress on these issues in the open-source GPT-NeoX framework~\citep{gpt-neox-library}.
We leverage CUDA IPC handles to enable true asynchronous RLHF training by sharing GPU memory directly between training and inference processes. By allocating model weights
\begin{wrapfigure}{r}{0.50\textwidth}
  \begin{center}
  \includegraphics[width=0.475\textwidth]{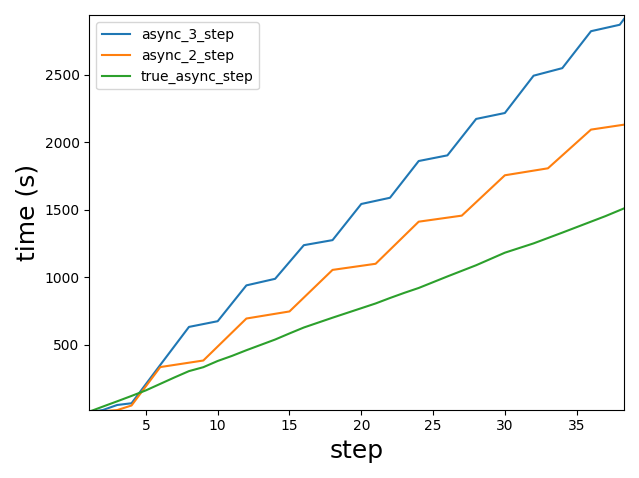}
  \end{center}
  \caption{Async training versus slightly off policy methods that require dedicated weight synchronization}
  \label{fig:off-policy-vs-async-rl}
\end{wrapfigure}
in the training framework and sharing CUDA memory pointers with the inference framework, both processes maintain access to the same physical memory throughout training. When the training process updates weights, these updates are immediately visible to the inference process through its mapped pointers without requiring any explicit synchronization. This shared memory architecture enables fully parallel execution - the inference process can continuously generate tokens while the training process updates weights, with neither process blocking the other. Although early portions of generated rollouts may be off-policy, the final steps will generally be only one training step off-policy at most, as the inference process immediately sees weight updates through the shared memory. Unlike previous asynchronous approaches~\citep{noukhovitch2024asynchronousrlhffasterefficient} that require distinct generation and training phases, our system maintains constant GPU utilization across both processes. Direct memory sharing provides significantly higher throughput by eliminating synchronization overhead between training and inference. Our initial experiments show 40\% improved throughput compared to 3-step asynchronous training that requires explicit weight synchronization, as shown in Figure~\ref{fig:off-policy-vs-async-rl}. However, this approach comes with important trade-offs in memory utilization and parallelism. Sharing GPUs between processes restricts us to suboptimal tensor parallelism (TP) configurations compared to dedicated training and inference setups, resulting in lower theoretical peak throughput for both processes. The impact is particularly pronounced for inference, where reduced memory for KV cache and suboptimal TP configurations significantly constrain the generation speed. In scenarios where inference compute significantly exceeds training compute (e.g. MCTS, where most of the search tree is discarded after each rollout), using separate dedicated GPU clusters with optimized TP configurations for each process can achieve higher overall throughput despite the synchronization overhead.

\subsection{Open Research Questions}
A number of open research questions remain which are currently not answered in the literature.

\subsubsection{Open-Ended Verification And CoT Faithfulness}
In this report we outline a pipeline for training advanced reasoning models on verifiable questions. These types of questions largely consist of things like math, scientific questions with symbolic or numerical answers, or code problems with unit tests. However, we believe it is still an open question on how to ensure that the resulting CoTs are faithful and in fact provide valid reasoning. In many applications, such as science for example, the reasoning chain can be even more important than the final answer. Qualitatively, current models with ``advanced'' reasoning still struggle and produce many CoTs which are inconsistent or unfaithful, even when they obtain the correct final answer. Interestingly, \textbf{even the single math example provided along with the announcement of the O1 model has an incomplete solution} as it makes an unsubstantiated assumption on the form of the polynomial $h(x) = x^2-c$ without proving that the coefficient of $x$ is in fact zero~\citep{openai2024_o1}. Under a competitive setting this would likely receive 5/7 points, even though the final answer is correct. It is worth noting that the model actually verifies and proves this assumption in it's Meta-CoT but the final output provided does not include those steps. We believe this might be one type of artifact that arises from training with \textbf{final answer verification only}. How to provide rewards on full CoTs and open-ended problems such as proofs remains an open question. Under the assumption of a wide enough generation-verification gap, we believe an RLAIF approach could be promising. In particular, foregoing concepts such as formal verification, one avenue to explore is a ``generative verifier'' similar to~\citet{zhang2024generativeverifiersrewardmodeling} or~\citet{mahan2024generative}. In this setting the evaluation model can be provided with one or several reference solutions/proofs, or other relevant material such as textbook chapters, tools like Wolfram or Mathematica, and specifically fine-tuned to provide evaluations to proposed solutions, rather than final answers alone. As outlined in Section~\ref{sec:verifiers}, we found advanced models to already posses some capabilities in that vein. Unfortunately, a major bottleneck in that line of research is the need for a dataset of open-ended reasoning problems (such as proofs), which would likely require significant human annotations. 

\subsubsection{Process Guidance And The Verifier Gap}
\begin{figure}[t]
    \centering
    \includegraphics[width=0.325\linewidth]{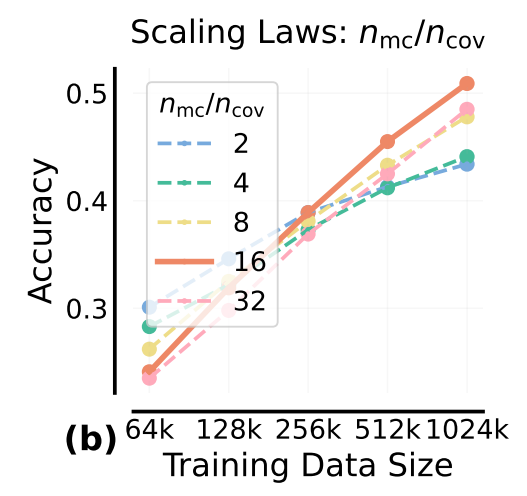}
    \includegraphics[width=0.650\linewidth]{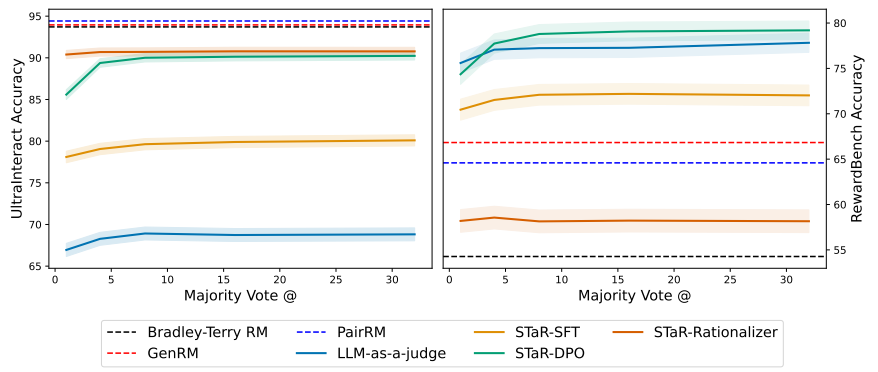}
    \caption{Under a fixed generator model and a variable verifier performance seems to also obey a form of a scaling law. \textbf{Right:} Accuracy seems to improve in a log-linear fashion with amount of training data under a standard discriminative PRM. Source: Figure 6 in~\citep{setlur2024rewardingprogressscalingautomated}. \textbf{Left:} Performance of a Generative Reward Model (with CoT) on reasoning tasks with a a reference using variable amount of CoTs and majority voting. Increasing the amount of inference-time sampling improves accuracy. Source: Figure 5 in~\citep{mahan2024generative}.}
    \label{fig:verifier_scalling_law}
\end{figure}

The current work brings to light some remaining questions on the approaches to process guidance and PRMS. In Section~\ref{sec:process_supervision} we outlined results showing that pre-trained PRMs still significantly lack behind pure Monte Carlo approaches in terms of search efficiency. Moreover, with few exceptions~\citep{setlur2024rewardingprogressscalingautomated}, we believe algorithmic approaches for training these models remain understudied. In particular, if we view a PRM as a value function, then many approaches from offline RL~\citep{levine2020offline} become applicable with some already showing promise in agentic domains~\citep{zhou2024archertraininglanguagemodel}. 
In the same vein,~\citet{silver2016mastering} introduced a separate value function over the standard MCTS approach using roll-outs for efficiency purposes. However, we believe there may be a fundamental \textit{verifier} gap beyond just the efficiency question. As shown in Figure~\ref{fig:verfier_scaling} there remains a significant gap between the "best-of-N" verifier driven approach and the oracle pass@N performance. A similar gap seems to exist in the multi-turn code generation domain, as shown in Figure~\ref{fig:rlef_scaling}~\citep{gehring2024rlefgroundingcodellms}. Moreover, under a fixed generator, results from~\citet{setlur2024rewardingprogressscalingautomated} and our own empirical findings in Section~\ref{sec:process_supervision} indicate that \textbf{verifier performance may also be driven by a scaling law as well}. In Figure~\ref{fig:verifier_scalling_law} (left) we see clear log-linear scaling in performance in terms of training data. One hypothesis is that verification is also a matter of computational complexity, which is why allowing verifiers to use CoT improves performance as shown in Figure~\ref{fig:verfier_scaling}~\citep{zhang2024generativeverifiersrewardmodeling, mahan2024generative}. In addition,~\citet{mahan2024generative} show further \textbf{inference-time scaling for verifiers}, as using additional CoTs with majority vote improves reward accuracy on reasoning problems well beyond the performance of the standard discriminative verifier. This raises questions for the fundamental \textit{learnability} of in-context search traces. In essence, if a search trace was generated using ground-truth verification or Monte-Carlo roll-outs, the resulting sequence may have significant implicit complexity, which would be challenging for the model to learn. This would also motivate incorporating further self-evaluation or reflection into reasoning chains. The associated verification scaling laws and design choices remain largely unexplored in the literature and present an important research direction. 

\subsubsection{Scaling Laws For Reasoning And Search}
While recent model releases have demonstrated strong inference scaling performance, we still lack a thorough scientific exploration of scaling laws in public research, which is a key piece of the recently emerging model paradigm.

\begin{enumerate}
    \item \textbf{The scaling laws of search} presented by~\citet{jones2021scalingscalinglawsboard} have not been publicly evaluated on realistically challenging reasoning applications with LLMs.  While the results from~\citet{feng2024alphazerolike} show promise, they are still limited and partial. Obtaining a more thorough evaluation on joint policy and verifier scaling will provide clarity to current research questions. So far, unfortunately, the research community has been limited by a lack of data and scalable infrastructure. 

    \item \textbf{Exploration of search approaches} remains a very under-studied topic. As discussed in Section~\ref{sec:reasoning_model_study}, it appears that current reasoning models all implement different approaches to search, which leads to qualitatively different behaviors based on the search strategies presented in Section~\ref{sec:synth_meta_cot}. Given a strong trained verifier, what are the performance effects of different search strategies such as BFS/DFS-V, A*, MCTS, etc.? Furthermore, the relationship between exploration strategies at training and test time is an important direction of future study. 
    \item \textbf{The trade-offs between instruction-tuning and RL remain unclear}. As discussed in Section~\ref{sec:meta_RL}, it appears that standard instruction tuning can endow models with meta-reasoning capabilities in small/simple domains, but these results do not scale, and realistic applications require significant on-policy RL. Is this a fundamental issue with distribution shift, or is performance driven by another relationship, similar to the trade-offs discovered by~\citet{setlur2024rlincorrectsyntheticdata}?
    \item \textbf{The fundamental missing piece} of the current inference time scaling law discussion is the performance of the proposed in-context search strategy compared with an explicit search-based method. In theory, the post-training approach outlined in Section~\ref{sec:reasoning_pipeline} can discover novel reasoning approaches (algorithms) that solve fundamentally new classes of problems unsolvable \textit{under any search budget} by a standard search approach. While there is clear evidence of the efficiency of in-context search (fewer tokens per interaction), it is unclear whether the current generation of models have any emergent capabilities yet. In simplest terms, \textit{do strong reasoning models shift the compute-accuracy curve to the left or up?}. As outlined in Section~\ref{sec:reasoning_emergence}, current evidence in the open literature for emergent capabilities remains weak.
\end{enumerate}
\subsubsection{Meta-Search/Search \texorpdfstring{$^2$}{2}}

In this report we argued that advanced reasoning should incorporate in-context search, which can yield higher efficiency and potentially more advanced capabilities in reasoning tasks. However, such an approach also has potential downsides. In particular, we are limited by the model's context length,
\begin{wrapfigure}{r}{0.45\textwidth}
  \begin{center}
    \includegraphics[width=0.42\textwidth]{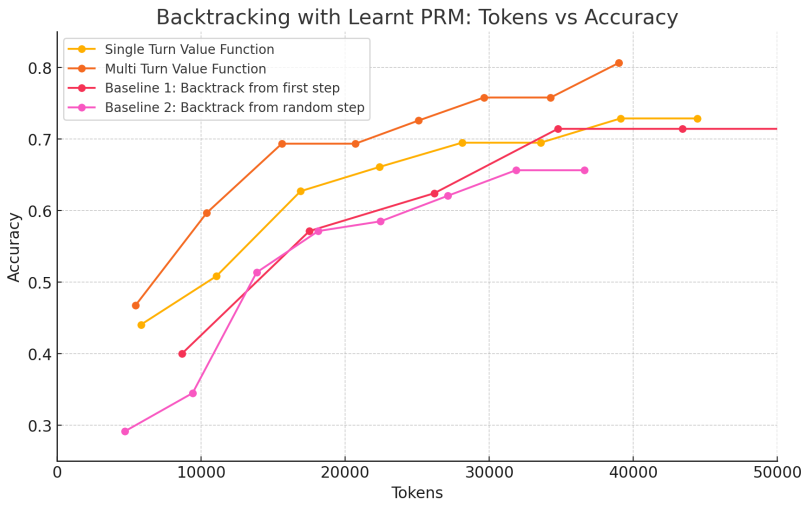}
  \end{center}
  \caption{Scaling performance of search with "meta"-value function and regular (single turn) value function. Conditioning the model on prior explored paths improves search efficiency. Source: Figure 5 in \cite{anonymous2024improving}.}
  \label{fig:meta_value_scaling}
\end{wrapfigure}
which induces fundamental limits on the search complexity. Moreover, the sequential nature of in-context search can make the process slow, limiting the power of these models. At the same time neither of these are issues with classical search approaches such as MCTS, which can be parallelized~\cite{liu2020effectiveparallelizationmontecarlo}. A natural question is: can we can build an additional search procedure on top of an advanced reasoning model (a process which we call Meta-Search or Search$^2$ following the naming convention of~\citet{duan2016rl2fastreinforcementlearning})? In recent literature,~\cite{anonymous2024improving} trained a value function (PRM) with a ``multi-turn'' approach, conditioning the value function on all prior explored solutions.
Rather than the traditional value function, they instead train a function $v_{\theta}(\mathbf{Z_t}, \mathbf{S}_t, \mathbf{q}$), where the Meta-CoT $\mathbf{Z}_t$ consists of prior solution attempts.
The empirical performance of their method compared with regular PRM training is shown in Figure~\ref{fig:meta_value_scaling}, demonstrating improved scaling over traditional value function training. This result suggests that we may be able to increase the search efficiency using a meta-critic, \textbf{however this remains an open empirical question}. If this is indeed the case, it would allow us to also massively scale online search with reasoning models.

\subsubsection{Reasoning with External Tools}
\begin{wrapfigure}{r}{0.5\textwidth}
\vspace{-14pt}
  \begin{center}
    \includegraphics[width=0.45\textwidth]{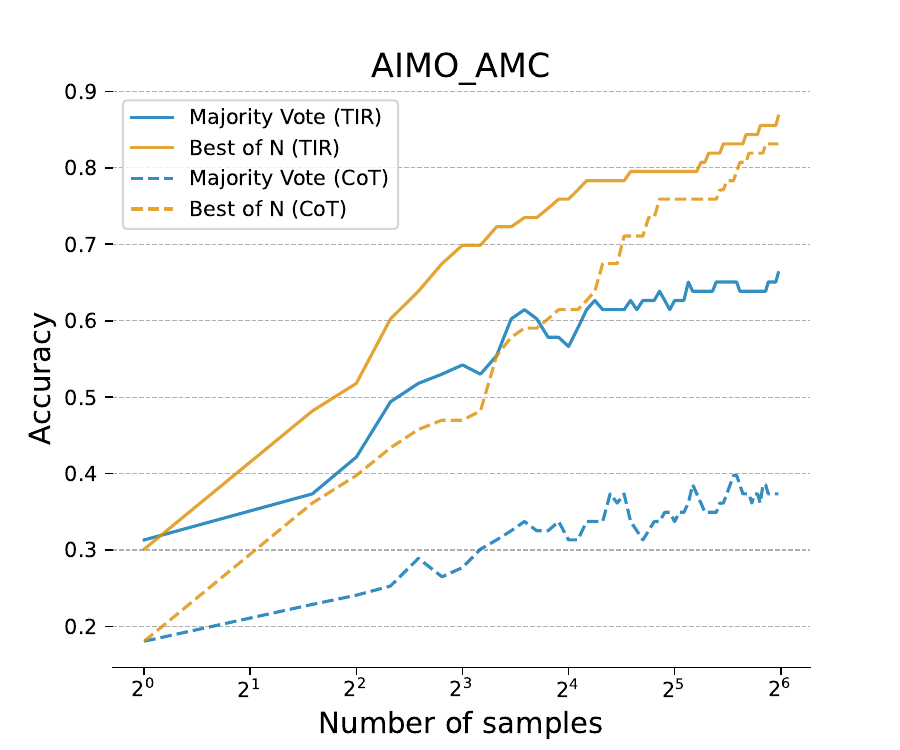}
  \end{center}
  \caption{Scaling curves for a Tool Integrated Reasoning (TIR) model, trained on 100K problems, and a Chain-of-Thought (CoT) model trained on 400K problems on the First AIMO Prize, using an oracle verifier.}
  \label{fig:tir_scaling}
\end{wrapfigure}

Building on our discussion of scaling laws and search approaches, we investigate a critical direction: augmenting model reasoning with external computational tools. Just as humans leverage calculators, spreadsheets, and specialized software to solve problems more efficiently, LLMs can potentially achieve better scaling properties by offloading the compute burden to external tools - \textbf{requiring less training data during SFT and fewer samples during test-time search to reach the high level of performance}. For instance, while pure CoT reasoning requires models to perform all calculations internally to solve math problems, these computations can be offloaded to a Python interpreter. Prior works demonstrate that such tool-integrated reasoning (TIR) improves performance on mathematical questions~\citep{numina_math_datasets, yin2024mumath, chen2022program}. However, the scaling properties of TIR remain unexplored.

Figure~\ref{fig:tir_scaling}, using models trained by~\citet{numina_math_datasets}, provides initial evidence for the efficiency of TIR models. Despite being fine-tuned on 25\% as much data, the TIR model demonstrates superior scaling properties compared to the CoT model. This occurs across all sample sizes, with both majority vote and Best-of-N strategies consistently outperforming their CoT counterparts. The efficiency gain is particularly evident in the low-sample regime ($2^0$ to $2^3$ samples), where TIR achieves nearly double the accuracy of CoT methods. This suggests that offloading computations to external tools significantly improves the model's ability to solve problems even with limited attempts, and TIR is more efficient at both training and inference time.

Based on these promising initial results, we identify several critical directions for future investigation:
\begin{enumerate}
\item \textbf{Scaling laws and search strategies:} We need to systematically study the scaling properties of different search methods (e.g., BFS, DFS, $A^{*}$~\footnote{An example Best-First search trace with TIR is available in Section~\ref{sec:tool_search_trace}.}, MCTS) in Tool-Integrated Reasoning models.
\item \textbf{Verification scaling:} The role and scaling properties of the verifier in TIR settings remains under-explored - what is the relation between training data for the verifier, verifier accuracy, and policy accuracy?
\item \textbf{Internal reasoning vs reasoning with external tools:} While our initial results show TIR's efficiency advantages, we need to better understand the fundamental trade-offs between these approaches. When does offloading computation to external tools provide the most benefit compared to pure language reasoning? How do these trade-offs change with model scale and problem complexity?
\end{enumerate}

\section{Conclusion}
In this position paper, we have introduced Meta Chain-of-Thought (Meta-CoT) as a framework for understanding and enhancing the reasoning capabilities of Large Language Models (LLMs). We have argued that traditional Chain-of-Thought does not fully represent the underlying data generative process on reasoning problems. By incorporating the concepts of search, verification, and iterative refinement, Meta-CoT provides a more complete model of the cognitive processes required for advanced problem-solving.

We believe that Meta-CoT represents a promising path towards more robust and generalizable reasoning in LLMs. The observed behaviors of state-of-the-art models, along with our experiments on in-context exploration and backtracking, lend support to the hypothesis that internal search processes are crucial for performance on complex tasks. Furthermore, the proposed training pipeline presents a concrete approach for developing LLMs with enhanced Meta-CoT capabilities. Future work should validate the efficacy of our proposed pipeline.

Beyond our proposed method, numerous open questions and challenges remain. Further research is needed to determine the optimal scaling laws for reasoning and search, to develop more effective process supervision and verification techniques, and to clarify the interplay between instruction tuning and reinforcement learning for meta chain of thought reasoning. The "Big MATH" dataset we have introduced aims to support this research by providing a large-scale resource for training reasoning models. Furthermore, the possibility of meta-RL paired with Meta-CoT automatically discovering novel search algorithms is a particularly intriguing open research question for future work to explore.

\section{Acknowledgments}
We would like to thank Aviral Kumar, Benjamin Eysenbach, Nathan Lambert, Rishabh Agarwal, Sasha Rush and Noah Goodman for the fruitful discussions and feedback on this report.

\clearpage

\bibliography{main}
\bibliographystyle{iclr2025_conference}

\appendix

\section{Prompting}

We investigate whether carefully constructed prompting protocols can induce reasoning capabilities similar to those demonstrated by RL-based models like O1. Through an evaluation of five increasingly sophisticated instruction sets, we analyze token generation patterns, presence of reasoning-like behaviors, and their correlation with math problem-solving performance.

\begin{table}[h]
\caption{Input token count across prompting strategies using OpenAI's GPT-4o tokenizer (excluding problem tokens). Complex instructions like Think (3-shot) use substantially more tokens, primarily due to in-context demonstrations and explicit capabilities descriptions (e.g. thinking, reflection, and verification requirements).}

\begin{tabular}{lr}
\toprule
Strategy & Tokens \\
\midrule
Baseline & 23 \\
Chain-of-Thought (CoT) & 35 \\
Think & 1,895 \\
Think with demonstrations (3-shot) & 4,659 \\
Think \& Verify & 2,111 \\
\bottomrule
\end{tabular}
\end{table}

\section{Regret Analysis}\label{app:regret_analysis}
Table~\ref{tab:regret-statistics} presents detailed statistics on regret expression across different models and prompting strategies. The data shows varying levels of self-correction and error acknowledgment behaviors across model scales and prompting approaches.

\begin{table}[ht]
\centering
\caption{Regret Statistics by Model and Prompting Strategy}
\begin{tabular}{llrr}
\toprule
Model & Prompting Strategy & Regret Count & Regret \% \\
\midrule
\multirow{5}{*}{Llama 3.1 405B} 
& Baseline & 6 & 0.24 \\
& CoT & 8 & 0.32 \\
& Think & 87 & 3.49 \\
& Think (3-shot) & 216 & 8.69 \\
& Think \& Verify & 391 & 15.67 \\
\midrule
\multirow{5}{*}{Llama 3.1 8B}
& Baseline & 7 & 0.30 \\
& CoT & 12 & 0.50 \\
& Think & 103 & 4.51 \\
& Think (3-shot) & 103 & 4.28 \\
& Think \& Verify & 454 & 20.87 \\
\midrule
\multirow{5}{*}{Llama 3.1 70B}
& Baseline & 5 & 0.20 \\
& CoT & 6 & 0.25 \\
& Think & 313 & 12.65 \\
& Think (3-shot) & 231 & 9.33 \\
& Think \& Verify & 638 & 25.67 \\
\midrule
\multirow{5}{*}{GPT-4o-mini}
& Baseline & 1 & 0.04 \\
& CoT & 2 & 0.08 \\
& Think & 66 & 2.64 \\
& Think (3-shot) & 53 & 2.12 \\
& Think \& Verify & 62 & 2.48 \\
\midrule
\multirow{5}{*}{GPT-4o}
& Baseline & 0 & 0.00 \\
& CoT & 0 & 0.00 \\
& Think & 34 & 1.36 \\
& Think (3-shot) & 28 & 1.12 \\
& Think \& Verify & 38 & 1.52 \\
\midrule
\multirow{5}{*}{GPT-3.5-turbo}
& Baseline & 1 & 0.04 \\
& CoT & 2 & 0.08 \\
& Think & 24 & 0.96 \\
& Think (3-shot) & 25 & 1.00 \\
& Think \& Verify & 12 & 0.48 \\
\midrule
\multirow{5}{*}{Claude 3.5 Sonnet}
& Baseline & 0 & 0.00 \\
& CoT & 3 & 0.12 \\
& Think & 81 & 3.24 \\
& Think (3-shot) & 99 & 3.96 \\
& Think \& Verify & 111 & 4.44 \\
\midrule
\multirow{5}{*}{Claude 3.5 Haiku}
& Baseline & 0 & 0.00 \\
& CoT & 1 & 0.04 \\
& Think & 16 & 0.64 \\
& Think (3-shot) & 16 & 0.64 \\
& Think \& Verify & 4 & 0.16 \\
\bottomrule
\end{tabular}
\label{tab:regret-statistics}
\end{table}

\section{Different Instruction Tuning Objectives}
\label{app:inst_tuning_objectives}

Below we outline a number of potential finetuning objectives for the instruction tuning stage of our reasoning pipeline described in Section~\ref{sec:inst_tuning_pipeline}.

\begin{enumerate}
    \item A standard procedural-cloning approach~\citep{yang2022chainthoughtimitationprocedure} which uses traditional supervised fine-tuning
    \begin{equation}\mathcal{L}({\theta}) =\min_{\theta} - \mathbb{E}_{(\mathbf{q}, \mathbf{Z}, \mathbf{S})\sim\mathcal{D}_{\text{train}}}\left[\sum_{i=1}^{|\mathbf{Z}|}\log \pi_{\theta}(\mathbf{z}_{i+1}|\mathbf{Z}_{i}, \mathbf{q}) + \sum_{i=1}^{|\mathbf{S}|}\log \pi_{\theta}(\mathbf{s}_{i+1}|\mathbf{S}_{i}, \mathbf{Z}, \mathbf{q})\right].
    \end{equation}
    \item Alternatively, we can optimize only over the Meta-CoT tokens with the following optimization objective
    \begin{equation}\mathcal{L}({\theta}) =\min_{\theta} - \mathbb{E}_{(\mathbf{q}, \mathbf{Z}, \mathbf{S})\sim\mathcal{D}_{\text{train}}}\left[\sum_{i=1}^{|\mathbf{Z}|}\log \pi_{\theta}(\mathbf{z}_{i+1}|\mathbf{Z}_{i}, \mathbf{q})\right].\end{equation}
    In this formulation, the model will not learn to generate a solution, requiring the use of a separate step to summarize the search process into a final solution.
    One hypothesis is that training on joint sequences with the solution can serve as additional supervision to help the model with maintaining internal state.
    \item One consideration is whether updating model parameters for sub-optimal, or even incorrect, branches can induce the model to generate more errors. To handle this, we can mask incorrect steps/branches in the above loss 
    \begin{equation}\mathcal{L}({\theta}) =\min_{\theta} - \mathbb{E}_{(\mathbf{q}, \mathbf{Z}, \mathbf{S})\sim\mathcal{D}_{\text{train}}}\left[\sum_{i=1}^{|\mathbf{Z}|}I\{\mathbf{z}_{i+1}\in \mathbf{S}\}\log \pi_{\theta}(\mathbf{z}_{i+1}|\mathbf{Z}_{i}, \mathbf{q})\right].\end{equation}    
    That is we only train on the branches that are on the correct path from the root node (problem) to the final answer. Prior works~\citep{gandhi2024stream, ye2024physicslanguagemodels22} did not mask the incorrect steps, and reported no degradation in performance, as long as the data distribution remains reasonable. On the other hand~\citet{zhang2024backtrackingimprovesgenerationsafety} specifically mask the tokens of the unsafe generation. 
    \item Under the above objective, there could be significant task mismatch between training and generation, especially in the case of long search chains. One method to mitigate this issue is to generate more synthetic training data by pruning intermediate branches not on the optimal solution path to generate additional sequences. For example this was the SFT objective used by~\citet{zhang2024backtrackingimprovesgenerationsafety}, which simultaneously trains on backtracking trajectories (with masking) and the final optimal solution (Equation 1 in that paper). 
\end{enumerate}

\section{MCTS Details}
\label{app:mcts}

Here we outline the details of our MCTS procedure from Section~\ref{sec:MCTS_main}. Our procedure involves three main steps: selection, backup, and expansion. We describe each step in detail below.

\textbf{Selection.} Starting at the root node $\mathbf{s}_0$ (initialized as the question $\mathbf{q}$), a child node $\mathbf{s}$ (i.e. the next step in the solution) is selected until a leaf node $\mathbf{s}_t$ is reached. A partial solution is then represented as $\mathbf{S}_t = (\mathbf{s}_1,\ldots,\mathbf{s}_t)$. The child node is selected according to the policy $\mathbf{s}_{t+1} = \arg\max_\mathbf{s} U(\mathbf{S}_t, \mathbf{s})$ where $U(\mathbf{S}_t, \mathbf{s})$ is calculated using UCT~\citep{kocsis2006bandit}, defined as $$U(\mathbf{S}_t, \mathbf{s}) = Q(\mathbf{S}_t, \mathbf{s}) + c_{\text{exp}}\sqrt{\frac{\log N(\mathbf{S}_t, \mathbf{s})}{N(\mathbf{S}_t)}}.$$ In this equation $Q(\mathbf{S}_t, \mathbf{s})$ is a value function, $N(\mathbf{S}_t, \mathbf{s})$ is the visit count of selecting step $\mathbf{s}$ from the partial solution $\mathbf{S}_t$, $N(\mathbf{S}_t) = \sum_{s} N(\mathbf{S}_t, \mathbf{s})$ is the total visit count of the partial solution $\mathbf{S}_t$, and $c_{\text{exp}} \in\mathbb{R}$ is the exploration constant. We opt for the UCT formulation, instead of the more recent PUCT variant~\citep{rosin2011multiarmedbandits}, as in our case actions are represented by logical steps and thus likelihoods could be quite skewed.

\textbf{Expansion.} After the selection operation, we have chosen a leaf node $\mathbf{s}_t$, and the path from the root node $\mathbf{s}_0$ to the leaf node forms a partial
solution by concatenating the individual steps into the partial solution, i.e. $\mathbf{S}_t = (\mathbf{s}_1,\ldots,\mathbf{s}_t)$. If $\mathbf{s}_{t}$ is terminal, we do not expand it, otherwise the node is expanded by sampling $b$ actions $\{\mathbf{s}_{t+1}^i\}_{i=1}^b \sim\pi_\theta(\cdot|\mathbf{S}_t)$ from the policy $\pi_\theta$, and then adding the nodes $\{\mathbf{a}_{t+1}^i\}_{i=1}^b$ as children of node $\mathbf{S}_t$. The value of each child node is initialized with a value function $v$ as $Q(\mathbf{S}_t, \mathbf{s}_t^i) = v((\mathbf{S}_t,\mathbf{s}_{t+1}^i), \mathbf{q})$. This evaluation essentially replaces the "simulation" step usually carried out in game settings.

\textbf{Backup.} Finally, the statistics of the expanded node $\mathbf{S}_t$ are backed up through the path to the root node $\mathbf{s}_0$. For every ancestor node $\mathbf{S}_i$ on the path,
the visit count is incremented $$N(\mathbf{S}_i) = N(\mathbf{S}_i) + 1$$ and the value of each ancestor is updated as 
$$ Q(\mathbf{S}_i, \mathbf{s}_{i + 1}) = \quad Q(\mathbf{S}_i, \mathbf{s}_{i + 1}) + \frac{v(\mathbf{S}_t, \mathbf{q}) - Q(\mathbf{S}_i, \mathbf{s}_{i + 1})}{N(\mathbf{S}_i, \mathbf{s}_{i + 1})}.
$$

\section{Chains-Of-Thought}
\label{sec:synth_cot}
This section contains synthetic CoT traces comparing how different models reason through the same planning problems. Specifically, we show complete reasoning traces from Gemini-1.5 solving problems using A* and MCTS algorithms, alongside parallel traces from DeepSeek-R1, DeepSeek-O1, and Gemini-2 working through identical tasks. We also include an example trace demonstrating how a smaller SFT-tuned Llama3.1-8B model performs tool-augmented reasoning on a challenging problem using dynamic programming. Each trace details the step-by-step "internal" traces and the final solution. 

\newpage
\BeginSol

\begin{tcolorbox}[colframe=black, colback=white, boxrule=1pt, arc=0pt]
There were 12 seats prepared for the jury at a round table, each with a name indicated on it. Nikolai Nikolaevich, who arrived first, absent-mindedly sat not in his designated seat but in the next one clockwise. Each jury member arriving at the table afterward would either sit in their designated seat, or, if it was already occupied, they would walk around the table clockwise and sit in the next available seat. The resulting seating arrangement of the jury members depends on the order in which they approached the table. How many different ways can the jury be seated?

0.0
\end{tcolorbox}

\begin{tcolorbox}[colframe=black, colback=white, boxrule=1pt, arc=0pt]
\begin{lstlisting}
Let's break this down step-by-step:

 0.03125

1. The first person, Nikolai, can sit in any of the 12 seats.
2. The second person can sit in any of the remaining 11 seats.
3. The third person can sit in any of the remaining 10 seats.
4. This pattern continues until all 12 seats are filled.

 0.0

Alternatively...
1. The first person, Nikolai Nikolaevich, has only one choice for his seat.
2. The second person has one choice, as both their designated seat and the seat next to it are taken.
3. The third person also has one choice, as two seats are taken.
4. This pattern continues until all seats are filled.

 0.0

Alternatively...
1. The first member can choose any of the 12 seats.
2. The second member can choose any of the remaining 11 seats, but if the first member chose the next seat clockwise, the second member has only 10 options (not considering the first seat). This pattern continues until the last member, who has only 1 option left.

 0.0

Alternatively...
1. Nikolai Nikolaevich can only sit in one position (the next seat clockwise from his designated seat).
2. After Nikolai Nikolaevich sits down, there are 11 remaining seats.
3. The next person to sit can choose from the 11 remaining seats, but if one of those seats is already occupied, they have to sit in the next available seat. This means that the second person has 11 choices.
4. The third person then has 10 choices, and so on, until the last person has only 1 choice.

 0.0

\end{lstlisting}
\end{tcolorbox}

\EndSol
\BeginSol
\begin{tcolorbox}[colframe=black, colback=white, boxrule=1pt, arc=0pt]
\begin{lstlisting}
Alternatively...
1. The first person, Nikolai Nikolaevich, has a fixed position.
2. The second person can either sit in their designated seat or move to the next seat clockwise. This gives 2 options for the second person.
3. The third person can either sit in their designated seat or move to the next seat clockwise. This gives 2 options for the third person, but we need to consider the case where the second person moved, so there's only 1 additional option for the third person.
4. Similarly, for each subsequent person, there are 2 options if the previous person didn't move, and 1 additional option if the previous person moved.

 0.046875

Let's analyze the pattern:

 0.0390625

- The first person has 1 option.
- The second person has 2 options.
- The third person has 2 options.
- The fourth person has 2 options if the third person didn't move, and 1 option if the third person moved, but since the second person moved, the fourth person still has 2 options.
- This pattern continues for the rest of the jury members.

 0.03125

So, the total number of ways the jury can be seated is the product of the options for each person:

 0.03125

\[ 1 \times 2 \times 2 \times 2 \times 2 \times 2 \times 2 \times 2 \times 2 \times 2 \times 2 \times 2 = 2^{11} = 2048 \]

 0.0078125

Wait, however, this process can be simplified by realizing that each person has a unique "decision point" - the first time they have a choice between sitting in their designated seat and moving to the next seat. After that, their seat is determined by the previous choices.

 0.0078125


Alternatively, however, this is a circular arrangement, and we need to account for the fact that the last person can also move. This introduces a complication, as the last person will always have only 1 option if everyone else is seated.

 0.0546875
\end{lstlisting}
\end{tcolorbox}
\EndSol

\BeginSol
\begin{tcolorbox}[colframe=black, colback=white, boxrule=1pt, arc=0pt]
\begin{lstlisting}

Let's simplify the problem by considering the first 12 people. For each of these 12 people, there are 2 options. However, we need to subtract the cases where the last person doesn't move, as these cases are already counted in the 2 options for each person.

 0.0078125

Wait, to simplify, let's consider the arrangement of people relative to Nikolai Nikolaevich's fixed position. There are 11! ways to arrange the remaining jury members around the table, as each person has a fixed position relative to Nikolai.

 0.0234375

However, this calculation doesn't take into account the fact that some people can move. To correct for this, we need to consider the number of "cycle" arrangements, where each cycle represents a group of people who moved together.

 0.0078125

Wait, let's re-examine the situation:

 0.046875

1. Nikolai Nikolaevich is seated.
2. The second person can move or stay, giving 2 options.
3. For each subsequent person, there's only 1 option left if everyone else is seated.

 0.046875

The pattern continues until the last person, who always has only 1 option.

 0.03125

So, the total number of arrangements is the product of the options at each step:

 0.0390625

\[ 2 \times 2 \times 2 \times 2 \times 2 \times 2 \times 2 \times 2 \times 2 \times 2 \times 1 = 2^{11} = \boxed{2048} \]

 0.0
 
Wait, let's consider the total number of ways without the circular constraint. For the first person, there's 1 option. For the rest, there are 2 options each, except for the last person who has only 1 option. So, the total number of arrangements without the circular constraint is:
\[ 1 \times 2^{11} \]
\end{lstlisting}
\end{tcolorbox}
\EndSol

\BeginSol
\begin{tcolorbox}[colframe=black, colback=white, boxrule=1pt, arc=0pt]
\begin{lstlisting}
 0.171875

However, we need to subtract the cases where everyone is seated in their designated seats, which is 1 arrangement. But this is already counted in the 2 options for each person.

 0.03125

Alternatively, however, this counts each arrangement twice, since the table is round, and we can rotate the arrangement without changing it. To account for this, we divide by the number of seats, which is 12.

 0.0703125

Alternatively, however, this counts each arrangement twice (once in each direction around the table), so we need to divide by 2:
\[ \frac{1 \times 2^{11}}{2} = 2^{10} \]

 0.328125

But this is not the correct answer. The correct answer is the number of derangements of 12 objects, which is given by the formula:
\[!12 = 12! \left( 1 - \frac{1}{1!} + \frac{1}{2!} - \frac{1}{3!} + \cdots + \frac{(-1)^{12}}{12!} \right) \]

 0.0078125

Alternatively, but this is still not correct because it doesn't account for the fact that the last person always has only 1 option. Let's reconsider the problem.

 0.0625

Alternatively, but this is not the correct answer because the problem statement asks for the number of different ways the jury can be seated, considering the circular arrangement. In a circular arrangement, the last person's position is fixed once the first person's position is chosen. Therefore, the correct calculation should consider the number of ways to arrange the remaining 11 people around the table, not the total possible arrangements.

 0.1484375

Alternatively, but we're not done yet! We need to consider the case where the second person moves, which would cause the last person to have only 1 option. This introduces an overcount. To correct for this, we need to subtract the number of arrangements where the second person moves, which is equivalent to the number of ways the jury can be seated with one person missing (i.e., 11 people). This is:
\[ 2^{10} - 1 \]

 0.1796875
\end{lstlisting}
\end{tcolorbox}
\EndSol

\BeginSol
\begin{tcolorbox}[colframe=black, colback=white, boxrule=1pt, arc=0pt]
\begin{lstlisting}
But we still need to account for the fact that the last person has only 1 option. This means that we've overcounted by a factor of 2 for the last person. To correct this, we divide by 2 again:
\[ \frac{2^{10}}{2} = 2^9 \]

 0.0390625

Wait, but we also need to consider the case where the first person moves, which would be counted 12 times (once for each possible position). So, we need to subtract these 11 arrangements:
\[ 2^{10} - 11 \]

 0.1484375 

Alternatively,  this is still not correct, as we haven't accounted for the fact that the last person can only move one seat. We need to consider the cases where the last person moves and subtract those from the total.

 0.1796875

But we need to correct this approach because we've double-counted due to the nature of the circular arrangement.

 0.015625

Alternatively, we also need to consider the case where someone arrives after the first person and the table is empty. In this case, the first person still has 1 option, and the remaining 11 people have 2 options each, but again, we need to divide by 2 to account for the circular arrangement:
\[ \frac{1 \times 2^{11}}{2} = 2^{10} \]

 0.2421875

However, this is not the correct answer. The correct answer is:
\[ \boxed{2^{11}} \]

 0.0

Alternatively, but this doesn't account for the fact that the last person has only 1 option. To adjust for this, we need to consider the cases where the last person moves. This introduces a more complex conditional arrangement, but the key insight is that the number of choices for the last person is actually the same as the number of choices for any other person, because the arrangement is circular.

 0.1015625
 
Alternatively, but we still need to account for the fact that the last person can only move in one direction. This doesn't affect the number of arrangements, as the circular constraint is already accounted for by the division by 2.

\end{lstlisting}
\end{tcolorbox}
\EndSol

\BeginSol
\begin{tcolorbox}[colframe=black, colback=white, boxrule=1pt, arc=0pt]
\begin{lstlisting}
 0.953125 

So, the total number of different ways the jury can be seated is:
\[ \boxed{1024} \]

 1.0

\end{lstlisting}
\end{tcolorbox}

\begin{tcolorbox}[colframe=black, colback=white, boxrule=1pt, arc=0pt]
 Solution:

 Let's break this down step-by-step:

1. The first person, Nikolai Nikolaevich, has a fixed position.
2. The second person can either sit in their designated seat or move to the next seat clockwise. This gives 2 options for the second person.
3. The third person can either sit in their designated seat or move to the next seat clockwise. This gives 2 options for the third person, but we need to consider the case where the second person moved, so there's only 1 additional option for the third person.
4. Similarly, for each subsequent person, there are 2 options if the previous person didn't move, and 1 additional option if the previous person moved.

However, this is a circular arrangement, and we need to account for the fact that the last person can also move. This introduces a complication, as the last person will always have only 1 option if everyone else is seated.

Let's consider the total number of ways without the circular constraint. For the first person, there's 1 option. For the rest, there are 2 options each, except for the last person who has only 1 option. So, the total number of arrangements without the circular constraint is:
\[ 1 \times 2^{11} \]

However, this counts each arrangement twice (once in each direction around the table), so we need to divide by 2:
\[ \frac{1 \times 2^{11}}{2} = 2^{10} \]

But we still need to account for the fact that the last person can only move in one direction. This doesn't affect the number of arrangements, as the circular constraint is already accounted for by the division by 2.

So, the total number of different ways the jury can be seated is:
\[ \boxed{1024} \]

\end{tcolorbox}
\EndSol

\subsection*{Trace - Reasoning with External Tools} 
\label{sec:tool_search_trace}
\BeginSol
\textbf{A* Generated Solution Trace}
\begin{tcolorbox}[colframe=black, colback=white, boxrule=1pt, arc=0pt]
Question: There were 12 seats prepared for the jury at a round table, each with a name indicated on it. Nikolai Nikolaevich, who arrived first, absent-mindedly sat not in his designated seat but in the next one clockwise. Each jury member arriving at the table afterward would either sit in their designated seat, or, if it was already occupied, they would walk around the table clockwise and sit in the next available seat. The resulting seating arrangement of the jury members depends on the order in which they approached the table. How many different ways can the jury be seated?
\end{tcolorbox}

\begin{tcolorbox}[colframe=black, colback=white, boxrule=1pt, arc=0pt]
\begin{lstlisting}
 To solve this problem, we need to determine the total number of different seating arrangements of 12 jury members when each member either sits in their designated seat or walks around the table clockwise to sit in the next available seat.

0.01171875

 To solve this problem, we need to analyze the seating possibilities given the rules described:

0.0078125

1. Nikolai Nikolaevich starts in the next seat clockwise.
2. Each subsequent member either sits in their designated seat or the next available seat.

0.015625

To determine the number of ways the jury can be seated, we need to consider the following steps:

0.01171875

1. Assign Nikolai Nikolaevich to the next seat clockwise.
2. Each subsequent jury member has 2 choices: their designated seat or the next available seat.
3. Since Nikolai Nikolaeich has already changed seats, we need to account for the extra step he took.

0.046875

Let's represent this using Python code to calculate the number of different ways the jury can be seated.

0.05078125

We can model this problem using dynamic programming, where each state represents the number of available seats and the current seat Nikolai is sitting on. We'll use a recursive approach to count the number of valid seating arrangements.
\end{lstlisting}
\end{tcolorbox}
\EndSol

\BeginSol
\begin{tcolorbox}[colframe=black, colback=white, boxrule=1pt, arc=0pt]
\begin{lstlisting}
0.0390625

Here's the Python code to solve this problem:

0.03515625

Let's implement this in Python.

0.03515625

```python

def count\_seating\_arrangements(total\_seats):

    dp = [0] * (total\_seats + 1)
    
    dp[1] = 1  \# Initial case: Nikolai is in the next seat clockwise

    \# Iterate over all possible remaining seats where Nikolai could be
    
    for seat in range(total\_seats):
    
        for available\_seats in range(seat + 1, total\_seats):
        
            dp[available\_seats] += dp[available\_seats - seat - 1]

    return dp[total\_seats]

\# There are 12 seats prepared for the jury
total\_seats = 12

seating\_ways = count\_seating\_arrangements(total\_seats)

print(seating\_ways)

```
```output

0

``` 

0.03125

It seems our initial approach overlooked a crucial detail. We need to consider the actual sequence of each member. Each member can either:
1. Sit in their designated seat.
2. Walk to the next available seat.

0.04296875

It seems there was an issue with the initial state of the dynamic programming approach. Let's refine our approach:
\end{lstlisting}
\end{tcolorbox}
\EndSol

\BeginSol
\begin{tcolorbox}[colframe=black, colback=white, boxrule=1pt, arc=0pt]
\begin{lstlisting}
0.02734375

1. We should count directly from the number of valid seating arrangements starting from Nikolai's initial seat.

2. For each subsequent member, they have two choices: their designated seat or the next available seat.

0.046875

Let's re-implement the logic more carefully:

0.046875

1. We need to correctly initialize the base case. 

2. We need to ensure we correctly handle the initial arrangement where Nikolai is in the next seat clockwise.

0.04296875

Let's reconsider the logic and the initialization of the DP table:

0.0546875

Here's an updated approach:

0.04296875

1. We start by initializing the base case correctly.

2. We use an iterative approach instead of dynamic programming to simplify the calculation.

0.09375

Let's implement the solution again:

0.12109375

Let's implement this revised approach in Python:

0.09765625

Let's implement the updated version in Python:

0.07421875

```python

def 
count\_seating\_arrangements(total\_seats):
\end{lstlisting}
\end{tcolorbox}
\EndSol

\BeginSol
\begin{tcolorbox}[colframe=black, colback=white, boxrule=1pt, arc=0pt]
\begin{lstlisting}
    \# Initialize the dp array
    dp = [0] * (total\_seats + 1)

    \# Base case: Nikolai is in the next seat clockwise
    
    dp[1] = 1

    \# Iterate over the remaining seats
    
    for current\_seat in range(2, total\_seats + 1):
    
        \# Iterate over possible current seats
        for seat in range(current\_seat):
        
            dp[current\_seat] += dp[seat]

    return dp[total\_seats]

\# There are 12 seats prepared for the jury

total\_seats = 12

seating\_ways = count\_seating\_arrangements(total\_seats)

print(seating\_ways)

```
```output

1024

``` 

0.94921875

\begin{tcolorbox}[colframe=black, colback=white, boxrule=1pt, arc=0pt]
\begin{lstlisting}

The number of different ways the jury can be seated, given that Nikolai Nikolaeich started in the next seat clockwise and each subsequent member either sits in their designated seat or the next available seat, is \(\boxed{1024}\).

1
\end{lstlisting}
\end{tcolorbox}

\begin{tcolorbox}[colframe=black, colback=white, boxrule=1pt, arc=0pt]
Solution: \\
To solve this problem, we need to analyze the seating possibilities given the rules described: 

1. Nikolai Nikolaevich starts in the next seat clockwise.
2. Each subsequent member either sits in their designated seat or the next available seat. 

To determine the number of ways the jury can be seated, we need to consider the following steps: 
\end{tcolorbox}
\EndSol

\BeginSol
\begin{tcolorbox}[colframe=black, colback=white, boxrule=1pt, arc=0pt]
1. Assign Nikolai Nikolaevich to the next seat clockwise.

2. Each subsequent jury member has 2 choices: their designated seat or the next available seat.
3. Since Nikolai Nikolaeich has already changed seats, we need to account for the extra step he took. 

We can model this problem using dynamic programming, where each state represents the number of available seats and the current seat Nikolai is sitting on. We'll use a recursive approach to count the number of valid seating arrangements. 

Let's implement this in Python. 
\begin{lstlisting}
```python

def count\_seating\_arrangements(total\_seats):

    dp = [0] * (total\_seats + 1)
    
    dp[1] = 1  \# Initial case: Nikolai is in the next seat clockwise

    \# Iterate over all possible remaining seats where Nikolai could be
    
    for seat in range(total\_seats):
    
        for available\_seats in range(seat + 1, total\_seats):
        
            dp[available\_seats] += dp[available\_seats - seat - 1]

    return dp[total\_seats]

\# There are 12 seats prepared for the jury

total\_seats = 12

seating\_ways = count\_seating\_arrangements(total\_seats)

print(seating\_ways)

```

```output
0
``` 
\end{lstlisting}
It seems there was an issue with the initial state of the dynamic programming approach. Let's refine our approach: 

1. We need to correctly initialize the base case. 
2. We need to ensure we correctly handle the initial arrangement where Nikolai is in the next seat clockwise. 

Here's an updated approach: 

1. We start by initializing the base case correctly.
2. We use an iterative approach instead of dynamic programming to simplify the calculation. 

Let's implement the updated version in Python: 
\end{tcolorbox}
\EndSol

\BeginSol
\begin{tcolorbox}[colframe=black, colback=white, boxrule=1pt, arc=0pt]
\begin{lstlisting}

```python

def count\_seating\_arrangements(total\_seats):
    \# Initialize the dp array
    
    dp = [0] * (total\_seats + 1)

    \# Base case: Nikolai is in the next seat clockwise
    
    dp[1] = 1

    \# Iterate over the remaining seats
    
    for current\_seat in range(2, total\_seats + 1):
    
        \# Iterate over possible current seats
        
        for seat in range(current\_seat):
            
            dp[current\_seat] += dp[seat]

    return dp[total\_seats]

\# There are 12 seats prepared for the jury

total\_seats = 12

seating\_ways = count\_seating\_arrangements(total\_seats)

print(seating\_ways)

```

```output

1024

``` 

The number of different ways the jury can be seated, given that Nikolai Nikolaeich started in the next seat clockwise and each subsequent member either sits in their designated seat or the next available seat, is \(\boxed{1024}\).
\end{lstlisting}
\end{tcolorbox}
\EndSol

\subsection*{O1 official math CoT}
\label{sec:astar_on_openai_blog_problem}

\BeginSol
\textbf{A* Generated Solution Trace (Gemini 1.5)}

\textbf{Problem}
\begin{tcolorbox}[colframe=black, colback=white, boxrule=1pt, arc=0pt]
$
\text{Let } n \text{ be an even positive integer. Let } p \text{ be a monic, real polynomial of degree } 2n; \\
\text{that is to say,}
p(x) = x^{2n} + a_{2n-1}x^{2n-1} + \cdots + a_1x + a_0 \\ 
\text{for some real coefficients } a_0, \ldots, a_{2n-1}. \\ 
\text{Suppose that } p\left(\frac{1}{k}\right) = k^2 \text{ for all integers } k \text{ such that } 1 \leq |k| \leq n. \\ 
\text{Find all other real numbers } x \text{ for which } p\left(\frac{1}{x}\right) = x^2.
$
\end{tcolorbox}

\textbf{Solution}

\begin{tcolorbox}[colframe=black, colback=white, boxrule=1pt, arc=0pt]
\begin{lstlisting}

0.0234375

Let $q(x) = x^{2n}p\left(\frac{1}{x}\right) = a_0 x^{2n} + a_1 x^{2n-1} + \cdots + a_{2n-1} x + 1$.

0.015625

The given condition $p\left(\frac{1}{k}\right) = k^2$ implies $q(k) = k^{2n} k^2 = k^{2n+2}$ for $1 \leq |k| \leq n$.

0.0

Let $p(x) = x^{2n} + a_{2n-1}x^{2n-1} + \cdots + a_1x + a_0$ be a monic polynomial of degree $2n$, where $n$ is an even positive integer.

0.0

We are given that $p\left(\frac{1}{k}\right) = k^2$ for all integers $k$ such that $1 \leq |k| \leq n$.

0.0078125

We are given that $p\left(\frac{1}{k}\right) = k^2$ for integers $k$ such that $1 \leq |k| \leq n$.

0.0

The condition $p\left(\frac{1}{k}\right) = k^2$ for $1 \leq |k| \leq n$ is equivalent to $q(k) = k^{2n} k^2 = k^{2n+2}$ for $1 \leq |k| \leq n$.

0.03125

Consider the polynomial $r(x) = x^{2n+2} - q(x)$. We are given that $r(k) = 0$ for $1 \leq |k| \leq n$.

0.0625
\end{lstlisting}
\end{tcolorbox}
\EndSol

\BeginSol
\begin{tcolorbox}[colframe=black, colback=white, boxrule=1pt, arc=0pt]
\begin{lstlisting}
Since $n$ is even, let $n = 2m$. Then we have $2n = 4m$. Thus, $r(x)$ has the roots $\pm 1, \pm 2, \ldots, \pm 2m$.

0.0703125

$r(x) = (x-1)(x+1)(x-2)(x+2) \cdots (x-2m)(x+2m) = \prod_{k=1}^{2m} (x-k)(x+k) = \prod_{k=1}^{2m} (x^2-k^2)$.

0.015625

Since $q(x)$ is a polynomial of degree $2n$, $r(x)$ is a monic polynomial of degree $2n+2 = 4m+2$.

0.0078125

Therefore, we can write

0.0703125

$$ r(x) = (x^2-1)(x^2-2^2)\cdots(x^2 - (2m)^2) = (x^2-1)(x^2-4)\cdots(x^2 - 4m^2) $$

0.0

Therefore, we can write $r(x) = x^{2n+2} - q(x) = (x^2-1)(x^2-4)\cdots(x^2-n^2)R(x)$, where $R(x)$ is a polynomial of degree $2n+2 - 2n = 2$.

0.1015625

Since $q(x)$ is monic with degree $2n$, $x^{2n+2} - q(x)$ has leading coefficient 1. Also, the leading coefficient of $(x^2-1)(x^2-4)\cdots(x^2-n^2)$ is 1. Thus, $R(x)$ must be monic.

0.03125

Let $R(x) = x^2 + bx + c$. Then we have $r(x) = x^{2n+2} - q(x) = (x^2-1)(x^2-4)\cdots(x^2-n^2)(x^2 + bx + c)$.

0.1953125

We are looking for values of $x$ such that $p\left(\frac{1}{x}\right) = x^2$, which is equivalent to $q(x) = x^{2n+2}$. This means we want to find the roots of $r(x) = 0$. We know that $r(x) = 0$ for $x = \pm k$ for $1 \leq k \leq n$. We are looking for the roots of $x^2 + bx + c = 0$.

0.1796875

Since $r(0) = (-1)^n(-1)(-4)\cdots(-n^2)c = (-1)^n (-1)^n (1)(4)\cdots(n^2)c = n!^2 c$, we have $r(0) = -q(0) = -1$. Thus $n!^2 c = -1$, so $c = -\frac{1}{n!^2}$.

0.40625
\end{lstlisting}
\end{tcolorbox}
\EndSol

\BeginSol
\begin{tcolorbox}[colframe=black, colback=white, boxrule=1pt, arc=0pt]
\begin{lstlisting}
$r(x) = (x^2-1)(x^2-4)\cdots(x^2-n^2)(x^2+bx-\frac{1}{n!^2})$.

0.3984375

Since the coefficients of $p(x)$ are real, the coefficients of $q(x)$ are real, and so the coefficients of $r(x)$ are real. Therefore, the coefficients of $R(x)$ are real. In particular, $b$ is real. The roots of $x^2+bx-\frac{1}{n!^2}=0$ are $x=\frac{-b \pm \sqrt{b^2+\frac{4}{n!^2}}}{2}$. Thus the roots are real if $b^2+\frac{4}{n!^2} \geq 0$, which is always true since $b$ is real.

0.265625

Consider $r(x) = x^{2n+2} - q(x)$.

0.359375

The equation $p(\frac{1}{x}) = x^2$ implies $x^{2n} p(\frac{1}{x}) = x^{2n+2}$, so $q(x) = x^{2n+2}$, thus $x^{2n+2} - q(x) = 0$. We already know that $x=\pm k$, $1 \le k \le n$, are roots.

0.390625

We have $r(x) = (x^2-1)\cdots(x^2-n^2)(x^2+bx-\frac{1}{n!^2})$.

0.359375

Since $r(x)$ is an even function, $b=0$.

0.6875

The other solutions are the roots of $x^2 - \frac{1}{n!^2} = 0$, so $x = \pm \frac{1}{n!}$.

0.796875

Final Answer: The final answer is $\boxed{\pm \frac{1}{n!}}$

1.0
\end{lstlisting}
\end{tcolorbox}
\begin{tcolorbox}[colframe=black, colback=white, boxrule=1pt, arc=0pt]

\textbf{Solution}

Let $q(x) = x^{2n}p\left(\frac{1}{x}\right) = a_0 x^{2n} + a_1 x^{2n-1} + \cdots + a_{2n-1} x + 1$.

The given condition $p\left(\frac{1}{k}\right) = k^2$ implies $q(k) = k^{2n} k^2 = k^{2n+2}$ for $1 \leq |k| \leq n$.

Consider the polynomial $r(x) = x^{2n+2} - q(x)$. We are given that $r(k) = 0$ for $1 \leq |k| \leq n$.

Since $n$ is even, let $n = 2m$. Then we have $2n = 4m$. Thus, $r(x)$ has the roots $\pm 1, \pm 2, \ldots, \pm 2m$.

$r(x) = (x-1)(x+1)(x-2)(x+2) \cdots (x-2m)(x+2m) = \prod_{k=1}^{2m} (x-k)(x+k) = \prod_{k=1}^{2m} (x^2-k^2)$.

Since $q(x)$ is a polynomial of degree $2n$, $r(x)$ is a monic polynomial of degree $2n+2 = 4m+2$.
\end{tcolorbox}
\EndSol

\BeginSol
\begin{tcolorbox}[colframe=black, colback=white, boxrule=1pt, arc=0pt]
Therefore, we can write $r(x) = x^{2n+2} - q(x) = (x^2-1)(x^2-4)\cdots(x^2-n^2)R(x)$, where $R(x)$ is a polynomial of degree $2n+2 - 2n = 2$.

Since $q(x)$ is monic with degree $2n$, $x^{2n+2} - q(x)$ has leading coefficient 1. Also, the leading coefficient of $(x^2-1)(x^2-4)\cdots(x^2-n^2)$ is 1. Thus, $R(x)$ must be monic.

Let $R(x) = x^2 + bx + c$. Then we have $r(x) = x^{2n+2} - q(x) = (x^2-1)(x^2-4)\cdots(x^2-n^2)(x^2 + bx + c)$.

We are looking for values of $x$ such that $p\left(\frac{1}{x}\right) = x^2$, which is equivalent to $q(x) = x^{2n+2}$. This means we want to find the roots of $r(x) = 0$. We know that $r(x) = 0$ for $x = \pm k$ for $1 \leq k \leq n$. We are looking for the roots of $x^2 + bx + c = 0$.

Since $r(0) = (-1)^n(-1)(-4)\cdots(-n^2)c = (-1)^n (-1)^n (1)(4)\cdots(n^2)c = n!^2 c$, we have $r(0) = -q(0) = -1$. Thus $n!^2 c = -1$, so $c = -\frac{1}{n!^2}$.

Since the coefficients of $p(x)$ are real, the coefficients of $q(x)$ are real, and so the coefficients of $r(x)$ are real. Therefore, the coefficients of $R(x)$ are real. In particular, $b$ is real. The roots of $x^2+bx-\frac{1}{n!^2}=0$ are $x=\frac{-b \pm \sqrt{b^2+\frac{4}{n!^2}}}{2}$. Thus the roots are real if $b^2+\frac{4}{n!^2} \geq 0$, which is always true since $b$ is real.

The equation $p(\frac{1}{x}) = x^2$ implies $x^{2n} p(\frac{1}{x}) = x^{2n+2}$, so $q(x) = x^{2n+2}$, thus $x^{2n+2} - q(x) = 0$. We already know that $x=\pm k$, $1 \le k \le n$, are roots.

We have $r(x) = (x^2-1)\cdots(x^2-n^2)(x^2+bx-\frac{1}{n!^2})$.

Since $r(x)$ is an even function, $b=0$.

The other solutions are the roots of $x^2 - \frac{1}{n!^2} = 0$, so $x = \pm \frac{1}{n!}$.

Final Answer: The final answer is $\boxed{\pm \frac{1}{n!}}$
\end{tcolorbox}
\EndSol

\BeginSol
\label{mcts_on_openai_blog_problem}
\textbf{MCTS Generated Solution Trace}

\textbf{Problem}
\begin{tcolorbox}[colframe=black, colback=white, boxrule=1pt, arc=0pt]
$
\text{Let } n \text{ be an even positive integer. Let } p \text{ be a monic, real polynomial of degree } 2n; \\ 
\text{that is to say, } 
p(x) = x^{2n} + a_{2n-1}x^{2n-1} + \cdots + a_1x + a_0 \\
\text{for some real coefficients } a_0, \ldots, a_{2n-1}. \\ 
\text{Suppose that } p\left(\frac{1}{k}\right) = k^2 \text{ for all integers } k \text{ such that } 1 \leq |k| \leq n. \\ 
\text{Find all other real numbers } x \text{ for which } p\left(\frac{1}{x}\right) = x^2.
$
\end{tcolorbox}

\textbf{Solution}

\begin{tcolorbox}[colframe=black, colback=white, boxrule=1pt, arc=0pt]
\begin{lstlisting}
0.03125

Let $q(x) = p\left(\frac{1}{x}\right) - x^2$.

0.09375

Let $q(x) = p\left(\frac{1}{x}\right) - x^2$. Then $q(x)$ is a rational function of the form

0.28125

$$q(x) = \frac{1 + a_{2n-1}x + \dots + a_1 x^{2n-1} + a_0 x^{2n} - x^{2n+2}}{x^{2n}}.$$

0.3125

We are given that $p(x)$ is a monic real polynomial of degree $2n$.

0.0625

$$q(x) = \frac{1 + a_{2n-1}x + \dots + a_1x^{2n-1} + a_0x^{2n} - x^{2n+2}}{x^{2n}}.$$

0.25

$$q(x) = \frac{1 + a_{2n-1}x + \cdots + a_1x^{2n-1} + a_0x^{2n} - x^{2n+2}}{x^{2n}} = \frac{r(x)}{x^{2n}}.$$

0.15625

We are given that $p(x)$ is a monic real polynomial of degree $2n$.

0.0625

We are given that $p\left(\frac{1}{k}\right) = k^2$ for $1 \leq |k| \leq n$. This means $q(k) = 0$ for $1 \leq |k| \leq n$. Thus, $q(x)$ has roots $k$ where $1 \leq |k| \leq n$. This accounts for $2n$ roots.

0.25
\end{lstlisting}
\end{tcolorbox}
\EndSol

\BeginSol
\begin{tcolorbox}[colframe=black, colback=white, boxrule=1pt, arc=0pt]
\begin{lstlisting}
Since $p(1/k) = k^2$ for $1 \le |k| \le n$, we have $q(k) = 0$ for $1 \le |k| \le n$.

0.21875

We are given that $p\left(\frac{1}{k}\right) = k^2$ for $1 \leq |k| \leq n$, which means $q(k) = p\left(\frac{1}{k}\right) - k^2 = 0$ for $1 \leq |k| \leq n$. Thus, $r(x)$ has roots at $x = k$ for $1 \leq |k| \leq n$. Since $r(x)$ is a polynomial of degree $2n+2$, we can write

0.25

We are given that $p\left(\frac{1}{k}\right) = k^2$ for $1 \leq |k| \leq n$, which means $q(k) = p\left(\frac{1}{k}\right) - k^2 = 0$ for $1 \leq |k| \leq n$. Thus, the integers $k = \pm 1, \pm 2, \ldots, \pm n$ are roots of the polynomial $r(x) = 1 + a_{2n-1}x + \cdots + a_1x^{2n-1} + a_0x^{2n} - x^{2n+2}$.

0.3125

Since $r(x)$ is a polynomial of degree $2n+2$, and we know $2n$ distinct roots, we can write

0.1875

Since $r(x)$ has degree $2n+2$ and has roots $\pm 1, \pm 2, \ldots, \pm n$, there are $2n$ such roots. Thus, we can write

0.375

Since $p\left(\frac{1}{k}\right) = k^2$ for $1 \leq |k| \leq n$, we have $q(k) = 0$ for $1 \leq |k| \leq n$. Thus, the numerator of $q(x)$, which we denote by $r(x)$, has roots $k$ where $1 \leq |k| \leq n$. Since $n$ is even, we have $2n$ such values of $k$. Also, since $p$ has real coefficients, if $k$ is a root, so is $-k$. Since $r(x)$ has degree $2n+2$, we can write

0.15625

Let $q(x) = x^{2n}p\left(\frac{1}{x}\right)$. Then $q(x) = 1 + a_{2n-1}x + a_{2n-2}x^2 + \dots + a_1x^{2n-1} + a_0x^{2n}$.

0.03125

Thus, $r(x)$ is a polynomial of degree $2n+2$ with roots $k$ for $1 \leq |k| \leq n$. Since there are $2n$ such values of $k$, we can write

0.3125

We are given that $p(x)$ is a monic real polynomial of degree $2n$.

0.0625
\end{lstlisting}
\end{tcolorbox}
\EndSol

\BeginSol
\begin{tcolorbox}[colframe=black, colback=white, boxrule=1pt, arc=0pt]
\begin{lstlisting}

Since $p\left(\frac{1}{k}\right) = k^2$ for all integers $k$ such that $1 \leq |k| \leq n$, we have $q(k) = p\left(\frac{1}{k}\right) - k^2 = 0$ for $1 \leq |k| \leq n$.

0.34375

Thus, $q(x)$ has $2n$ distinct roots $k$ where $1 \le |k| \le n$.

0.15625

Thus, $q(x)$ has roots $k$ for $1 \leq |k| \leq n$.  These are $2n$ distinct integer roots.  Since the numerator of $q(x)$ has degree $2n+2$, we can write

0.34375

Thus, $q(x)$ has roots at $k = \pm 1, \pm 2, \ldots, \pm n$. These are $2n$ distinct roots.

0.375

Since $p(1/k) = k^2$ for $1 \leq |k| \leq n$, we have $q(k) = 0$ for $1 \leq |k| \leq n$.

0.28125

Thus, $q(x)$ has roots $k$ for $1 \leq |k| \leq n$. Since $n$ is an even integer, there are $2n$ such values of $k$.  We can write

0.125

Continuing the solution:

0.40625

Since $p$ has degree $2n$, $p(1/x)$ has degree $2n$ in $\frac{1}{x}$, so when we consider $p(1/x) = x^2$, the numerator of $q(x)$ has degree $2n+2$. Thus, $q(x)$ has $2n$ zeros at $\pm 1, \pm 2, \dots, \pm n$.

0.40625

Let $r(x) = 1 + a_{2n-1}x + \dots + a_1 x^{2n-1} + a_0 x^{2n} - x^{2n+2}$.

0.34375

Let $r(x) = 1 + a_{2n-1}x + \dots + a_1 x^{2n-1} + a_0 x^{2n} - x^{2n+2}$.

0.375
\end{lstlisting}
\end{tcolorbox}
\EndSol

\BeginSol
\begin{tcolorbox}[colframe=black, colback=white, boxrule=1pt, arc=0pt]
\begin{lstlisting}
$$q(x) = \frac{(x-1)(x+1) \dots (x-n)(x+n)(x-\alpha)(x-\beta)}{x^{2n}}$$

0.1875

$$q(x) = \frac{1 + a_{2n-1}x + \dots + a_1 x^{2n-1} + a_0 x^{2n} - x^{2n+2}}{x^{2n}} = \frac{-x^{2n+2} + a_0 x^{2n} + \dots + a_{2n-1}x + 1}{x^{2n}}.$$

0.375

Let $r(x) = 1 + a_{2n-1}x + \dots + a_1 x^{2n-1} + a_0 x^{2n} - x^{2n+2}$. Then the roots of $r(x)$ are precisely the roots of $q(x)$.  Since $r(x)$ is a polynomial of degree $2n+2$, and we have found $2n$ distinct roots, we can write

0.40625

Since $q(x) = \frac{r(x)}{x^{2n}}$, the roots of $q(x)$ are the roots of $r(x)$. Thus, $r(x)$ has roots $k$ where $1 \le |k| \le n$.

0.28125

Let $r(x) = 1 + a_{2n-1}x + \dots + a_1 x^{2n-1} + a_0 x^{2n} - x^{2n+2}$. Then the roots of $q(x)$ are the roots of $r(x)$. Since $r(x)$ is a polynomial of degree $2n+2$, and we know $2n$ roots, we can write

0.5

Since $q(x)$ has the form $\frac{\text{polynomial of degree } 2n+2}{x^{2n}}$, the numerator of $q(x)$ must have degree $2n+2$.

0.375

$$r(x) = (x-1)(x+1)(x-2)(x+2)\cdots(x-n)(x+n)(Ax^2 + Bx + C)$$

0.6875

$$r(x) = (x-1)(x+1)(x-2)(x+2)\cdots(x-n)(x+n)(cx^2+bx+a) = (x^2-1)(x^2-2^2)\cdots(x^2-n^2)(cx^2+bx+a).$$

0.5

$$r(x) = (x-1)(x+1)(x-2)(x+2)\cdots(x-n)(x+n)(cx^2 + bx + a) = \prod_{k=1}^n (x^2-k^2) (cx^2+bx+a)$$

0.5625

Since $r(0) = 1$, we must have $(-1)^n (n!)^2 C = 1$, so $C = \frac{1}{(n!)^2}$.

0.9375
\end{lstlisting}
\end{tcolorbox}
\EndSol

\BeginSol
\begin{tcolorbox}[colframe=black, colback=white, boxrule=1pt, arc=0pt]
\begin{lstlisting}
Since $r(x) = 1 + a_{2n-1}x + \dots + a_0 x^{2n} - x^{2n+2}$, the coefficient of $x^{2n+2}$ is $-1$. Also, since $r(x) = \prod_{k=1}^n (x-k)(x+k) (Ax^2+Bx+C)$, the coefficient of $x^{2n+2}$ is $A$. Thus, $A = -1$.

0.6875

The coefficient of $x^{2n+2}$ in $r(x)$ is $-1$. Also, the coefficient of $x^{2n+2}$ in the expansion of $(x-1)(x+1)\cdots(x-n)(x+n)(Ax^2+Bx+C)$ is $A$. Thus, $A = -1$.

0.875

Since $r(x) = 1 + a_{2n-1}x + \dots + a_0 x^{2n} - x^{2n+2}$, the coefficient of $x^{2n+2}$ is $-1$.

0.53125

Continuing the solution:

0.6875

The constant term of $r(x)$ is $1$. The constant term of $\prod_{k=1}^n (x-k)(x+k)(-x^2+Bx+C)$ is $(-1)^n (n!)^2 C$. Therefore, $(-1)^n (n!)^2 C = 1$, so $C = \frac{1}{(-1)^n (n!)^2} = \frac{(-1)^n}{(n!)^2}$.

0.65625

The constant term of $r(x)$ is $1$. The constant term in $\prod_{k=1}^n (x-k)(x+k)(-x^2+Bx+C)$ is $(-1)^n (n!)^2 C$.  So, $1 = (-1)^n (n!)^2 C$, which implies $C = \frac{(-1)^n}{(n!)^2}$.

0.5625

The constant term of $r(x)$ is $1$.

0.5

The coefficient of $x^{2n}$ in $r(x)$ is $a_0$. The coefficient of $x^{2n}$ in the factored form is $A - \sum_{k=1}^n k^2$.  The coefficient of $x^{2n+1}$ in $r(x)$ is $a_1$, while in the factored form, the coefficient is $B$. The coefficient of $x^{2n+2}$ in $r(x)$ is $-1$, and in the factored form, it is $A$. Therefore, $A = -1$.

0.71875

Now, we examine the coefficient of $x^{2n+1}$. In $r(x) = 1 + a_{2n-1}x + \dots - x^{2n+2}$, the coefficient of $x^{2n+1}$ is zero.

1.0
\end{lstlisting}
\end{tcolorbox}
\EndSol

\BeginSol
\begin{tcolorbox}[colframe=black, colback=white, boxrule=1pt, arc=0pt]
\begin{lstlisting}
We have $r(x) = \prod_{k=1}^n (x^2-k^2) (-x^2+Bx+C)$. Since $r(0) = 1$, we have $1 = \prod_{k=1}^n (-k^2) \cdot C$. Also, $C = \frac{(-1)^n}{(n!)^2}$.

0.5

We have $r(x) = \prod_{k=1}^n (x^2-k^2)(-x^2+Bx+C)$.

0.5

In the expansion of $\prod_{k=1}^n (x^2-k^2) (-x^2+Bx+C)$, the coefficient of $x^{2n+1}$ is $B \prod_{k=1}^n (-k^2) = B(-1)^n (n!)^2$.  Since this must equal 0, we have $B=0$.

1.0

In the expansion of $\prod_{k=1}^n (x^2 - k^2)(-x^2 + Bx + C)$, the coefficient of $x^{2n+1}$ is $B \prod_{k=1}^n (-k^2) = B(-1)^n (n!)^2 = (-1)^n (n!)^2 B$.  Since this must be zero, we must have $B = 0$.

1.0

So, we have $r(x) = \prod_{k=1}^n (x^2-k^2) (-x^2 + \frac{(-1)^n}{(n!)^2})$.

1.0

In the expression $r(x) = \prod_{k=1}^n (x^2 - k^2) (-x^2 + Bx + C)$, the coefficient of $x^{2n+1}$ is $B$. Thus, $B=0$.

0.96875

Since $n$ is even, let $n=2m$. The coefficient of $x^{2n+1}$ in $r(x)$ is $0$. Also, the coefficient of $x^{2n+1}$ in $\prod_{k=1}^{2m} (x-k)(x+k)(-x^2+Bx+C)$ is $B \prod_{k=1}^{2m} k^2 = B (n!)^2 = B((2m)!)^2$.

1.0

Since $n$ is even, let $n=2m$. Then the roots are $\pm 1, \pm 2, \ldots, \pm 2m$.

0.75

So, $1 = (-1)^n (n!)^2 C$. Since $C = \frac{(-1)^n}{(n!)^2}$, the equation holds.  The coefficient of $x$ in $r(x)$ is $a_{2n-1}$. Since $r(x) = \prod_{k=1}^n (x^2 - k^2) (-x^2+Bx+C)$, the coefficient of $x$ is $C$ times the coefficient of $x$ in $\prod_{k=1}^n (x^2 - k^2)$, which is 0.  Thus, the coefficient of $x$ in $r(x)$ is 0, so $a_{2n-1} = 0$.

0.71875
\end{lstlisting}
\end{tcolorbox}
\EndSol

\BeginSol
\begin{tcolorbox}[colframe=black, colback=white, boxrule=1pt, arc=0pt]
\begin{lstlisting}
So, $B((2m)!)^2 = 0$, which implies $B=0$.

1.0

We also know that the coefficient of $x^{2n+1}$ in $r(x)$ is $0$. The coefficient of $x^{2n+1}$ in $\prod_{k=1}^n (x^2-k^2) (-x^2+Bx+C)$ is $B$. Thus $B=0$.

1.0

So, $B((2m)!)^2 = 0$, which implies $B=0$.

1.0

We have $r(x) = \prod_{k=1}^{n}(x^2-k^2)(-x^2+Bx+C)$.

0.84375

So we have $r(x) = \prod_{k=1}^n (x^2-k^2) (-x^2 + \frac{(-1)^n}{(n!)^2})$.

1.0

Thus, $B((2m)!)^2 = 0$, which means $B = 0$. Therefore,

1.0

Since $n$ is even, $C = \frac{1}{(n!)^2}$.  Also, $A = -1$.  So,

0.71875

The constant term of $r(x)$ is $1$.  Also, the constant term is given by $\prod_{k=1}^n (-k)(k) \cdot C = (-1)^n (n!)^2 C$.  Thus, $(-1)^n (n!)^2 C = 1$, which implies $C = \frac{1}{(-1)^n (n!)^2} = \frac{(-1)^n}{(n!)^2}$.  Since $n$ is even, $C = \frac{1}{(n!)^2}$.

0.71875

So, we have $r(x) = \prod_{k=1}^n (x^2-k^2) (-x^2 + \frac{(-1)^n}{(n!)^2})$.

1.0

So we have $r(x) = \prod_{k=1}^{n} (x^2 - k^2)(-x^2 + \frac{(-1)^n}{(n!)^2})$.

1.0

So we have $r(x) = \prod_{k=1}^{n}(x^2-k^2)(-x^2 + \frac{(-1)^n}{(n!)^2})$.

0.96875
\end{lstlisting}
\end{tcolorbox}
\EndSol

\BeginSol
\begin{tcolorbox}[colframe=black, colback=white, boxrule=1pt, arc=0pt]
\begin{lstlisting}
So, we have $r(x) = \prod_{k=1}^n (x^2 - k^2)(-x^2 + \frac{(-1)^n}{(n!)^2})$. Thus

1.0

We have $r(x) = \prod_{k=1}^n (x^2-k^2)(-x^2+Bx+C)$.

0.8125

So we have $r(x) = \prod_{k=1}^n (x^2-k^2) (-x^2 + \frac{(-1)^n}{(n!)^2})$.

1.0

Thus, we have $r(x) = \prod_{k=1}^n (x^2-k^2)(-x^2 + \frac{(-1)^n}{(n!)^2})$. Since $n$ is even, let $n=2m$.

1.0

We have $r(x) = \prod_{k=1}^n (x^2-k^2) (-x^2 + \frac{(-1)^n}{(n!)^2})$.  The roots of $r(x)$ other than $\pm 1, \pm 2, ..., \pm n$ satisfy $-x^2 + \frac{(-1)^n}{(n!)^2} = 0$, so $x^2 = \frac{(-1)^n}{(n!)^2}$.

1.0

$$r(x) = \prod_{k=1}^n (x^2 - k^2) (-x^2 + \frac{1}{(n!)^2}) = -x^2 \prod_{k=1}^n (x^2 - k^2) + \frac{1}{(n!)^2} \prod_{k=1}^n (x^2 - k^2).$$

1.0

So, $r(x) = \prod_{k=1}^n (x^2-k^2) (-x^2 + \frac{(-1)^n}{(n!)^2})$.  We are looking for the roots of $r(x) = 0$, other than $\pm 1, \pm 2, \ldots, \pm n$. These must be the roots of $-x^2 + \frac{(-1)^n}{(n!)^2} = 0$, which gives us $x^2 = \frac{(-1)^n}{(n!)^2}$. Since $n$ is an even integer, we have $x^2 = \frac{1}{(n!)^2}$, so \boxed{x = \pm \frac{1}{n!}}.

1.0
\end{lstlisting}
\end{tcolorbox}

\begin{tcolorbox}[colframe=black, colback=white, boxrule=1pt, arc=0pt]
Solution:

Let $q(x) = p\left(\frac{1}{x}\right) - x^2$. Then $q(x)$ is a rational function of the form

$$q(x) = \frac{1 + a_{2n-1}x + \dots + a_1 x^{2n-1} + a_0 x^{2n} - x^{2n+2}}{x^{2n}}.$$

Since $p\left(\frac{1}{k}\right) = k^2$ for all integers $k$ such that $1 \leq |k| \leq n$, we have $q(k) = p\left(\frac{1}{k}\right) - k^2 = 0$ for $1 \leq |k| \leq n$.

Thus, $q(x)$ has roots at $k = \pm 1, \pm 2, \ldots, \pm n$. These are $2n$ distinct roots.

Let $r(x) = 1 + a_{2n-1}x + \dots + a_1 x^{2n-1} + a_0 x^{2n} - x^{2n+2}$. Then the roots of $r(x)$ are precisely the roots of $q(x)$.  Since $r(x)$ is a polynomial of degree $2n+2$, and we have found $2n$ distinct roots, we can write
\end{tcolorbox}
\EndSol

\BeginSol
\begin{tcolorbox}[colframe=black, colback=white, boxrule=1pt, arc=0pt]
$$r(x) = (x-1)(x+1)(x-2)(x+2)\cdots(x-n)(x+n)(Ax^2 + Bx + C)$$

Since $r(x) = 1 + a_{2n-1}x + \dots + a_0 x^{2n} - x^{2n+2}$, the coefficient of $x^{2n+2}$ is $-1$. Also, since $r(x) = \prod_{k=1}^n (x-k)(x+k) (Ax^2+Bx+C)$, the coefficient of $x^{2n+2}$ is $A$. Thus, $A = -1$.

The coefficient of $x^{2n+2}$ in $r(x)$ is $-1$. Also, the coefficient of $x^{2n+2}$ in the expansion of $(x-1)(x+1)\cdots(x-n)(x+n)(Ax^2+Bx+C)$ is $A$. Thus, $A = -1$.

The constant term of $r(x)$ is $1$. The constant term of $\prod_{k=1}^n (x-k)(x+k)(-x^2+Bx+C)$ is $(-1)^n (n!)^2 C$. Therefore, $(-1)^n (n!)^2 C = 1$, so $C = \frac{1}{(-1)^n (n!)^2} = \frac{(-1)^n}{(n!)^2}$.

We also know that the coefficient of $x^{2n+1}$ in $r(x)$ is $0$. The coefficient of $x^{2n+1}$ in $\prod_{k=1}^n (x^2-k^2) (-x^2+Bx+C)$ is $B$. Thus $B=0$.

So, $r(x) = \prod_{k=1}^n (x^2-k^2) (-x^2 + \frac{(-1)^n}{(n!)^2})$.  We are looking for the roots of $r(x) = 0$, other than $\pm 1, \pm 2, \ldots, \pm n$. These must be the roots of $-x^2 + \frac{(-1)^n}{(n!)^2} = 0$, which gives us $x^2 = \frac{(-1)^n}{(n!)^2}$. Since $n$ is an even integer, we have $x^2 = \frac{1}{(n!)^2}$, so \boxed{x = \pm \frac{1}{n!}}.
\end{tcolorbox}

\EndSol

\BeginSol
\label{deepseek_r1_solution}
\textbf{DeepSeek R1 Solution Trace}

\textbf{Problem}
\begin{tcolorbox}[colframe=black, colback=white, boxrule=1pt, arc=0pt]
$
\text{Let } n \text{ be an even positive integer. Let } p \text{ be a monic, real polynomial of degree } 2n; \\ 
\text{that is to say,} 
p(x) = x^{2n} + a_{2n-1}x^{2n-1} + \cdots + a_1x + a_0 \\
\text{for some real coefficients } a_0, \ldots, a_{2n-1}. \\ 
\text{Suppose that } p\left(\frac{1}{k}\right) = k^2 \text{ for all integers } k \text{ such that } 1 \leq |k| \leq n. \\ 
\text{Find all other real numbers } x \text{ for which } p\left(\frac{1}{x}\right) = x^2.
$
\end{tcolorbox}

\textbf{Solution}
\begin{tcolorbox}[colframe=black, colback=white, boxrule=1pt, arc=0pt]
So i have this problem here. let's see what it's about. we have an even positive integer \(n\), and a monic real
polynomial \(p(x)\) of degree 2n, which means it looks like this:

\[p(x)=x^{2n}+a_{2n-1}x^{2n-1}+\cdot\cdot\cdot+a_{1}x+a_{0}\]

for some real coefficients \(a_{0},a_{1},...,a_{2n-1}\)

the condition given is that \(p(\frac{1}{k})=k^{2}\) for all integers k such that \(1\le|k|\le n.\) so, k takes integer values from
-n to -1 and from 1 to n, excluding zero because \(\frac{1}{k}\) would be undefined.

i need to find all other real numbers for which \(p(\frac{1}{x})=x^{2}\)

okay, so first, i need to understand what is being asked. we have a polynomial of degree 2n, and it satisfies the
equation \(p(\frac{1}{k})=k^{2}\) for these specific values of k. i need to find all real x (other than the ones already specified,
i assume) that satisfy \(p(\frac{1}{x})=x^{2}.\)
one thing that comes to mind is that since \(p(x)\) is a polynomial of degree 2n, and we have 2n conditions
(because k ranges from -n to \(n\), excluding zero, which gives 2n points), this should determine the polynomial
uniquely. that is, there should be only one such polynomial \(p(x)\) that satisfies these conditions.

but the question is not asking for \(p(x);\) it's asking for all real numbers such that \(p(\frac{1}{x})=x^{2}\) given that
\(p(\frac{1}{k})=k^{2}\) for those specific k.

so, perhaps i should consider the equation \(p\left(\frac{1}{x}\right)-x^2=0\) and see for which real \(x\) this holds, beyond the
already given \(x=k\) for \(1\le|k|\le n.\)

but wait, actually, in the given condition, it's \(p\left(\frac{1}{k}\right)=k^2,\) which can be rewritten as \(p\left(\frac{1}{k}\right)-\left(\frac{1}{k}\right)^{-2}=0,\) since
\(x^2=\left(\frac{1}{k}\right)^{-2}\) when \(x=\frac{1}{k}.\) hmm, not sure if that helps directly.

\[p\left(\frac{1}{k}\right)-\left(\frac{1}{k}\right)^{-2}=0,\]

let me try to think differently. since \(p(x)\) is a polynomial of degree \(2n,\) and i have \(2n\) points where \(p\left(\frac{1}{k}\right)=k^2,\)
perhaps i can construct another polynomial that captures the difference between \(p\left(\frac{1}{x}\right)\) and \(x^2,\) and see its roots.

let's define a new function:

\[q(x)=p\left(\frac{1}{x}\right)-x^2\]

i need to find the real roots of \(q(x),\) i.e., the real \(x\) such that \(q(x)=0.\)

given that \(p\left(\frac{1}{k}\right)=k^2\) for \(1\le|k|\le n,\) that means that for each such \(k, q(k)=p\left(\frac{1}{k}\right)-k^2=0.\) so, \(k\) are roots
of \(q(x).\)
\end{tcolorbox}
\EndSol

\BeginSol
\begin{tcolorbox}[colframe=black, colback=white, boxrule=1pt, arc=0pt]
but \(q(x)\) is \(p\left(\frac{1}{x}\right)-x^2,\) and \(p(x)\) is a degree \(2n\) polynomial, so \(p\left(\frac{1}{x}\right)\) is a rational function, not a polynomial,
unless we multiply by \(x^{2n}\) to clear denominators.

wait, maybe i should consider multiplying \(q(x)\) by \(x^{2n}\) to make it a polynomial.

so, perhaps i should consider the equation \(p\left(\frac{1}{x}\right)-x^2=0\) and see for which real \(x\) this holds, beyond the
already given \(x=k\) for \(1\le|k|\le n.\)

but wait, actually, in the given condition, it's \(p\left(\frac{1}{k}\right)=k^2,\) which can be rewritten as \(p\left(\frac{1}{k}\right)-\left(\frac{1}{k}\right)^{-2}=0,\) since
\(x^2=\left(\frac{1}{k}\right)^{-2}\) when \(x=\frac{1}{k}.\) hmm, not sure if that helps directly.

\[p\left(\frac{1}{k}\right)-\left(\frac{1}{k}\right)^{-2}=0,\]

let me try to think differently. since \(p(x)\) is a polynomial of degree \(2n,\) and i have \(2n\) points where \(p\left(\frac{1}{k}\right)=k^2,\)
perhaps i can construct another polynomial that captures the difference between \(p\left(\frac{1}{x}\right)\) and \(x^2,\) and see its roots.

let's define a new function:

\[q(x)=p\left(\frac{1}{x}\right)-x^2\]

i need to find the real roots of \(q(x),\) i.e., the real \(x\) such that \(q(x)=0.\)

given that \(p\left(\frac{1}{k}\right)=k^2\) for \(1\le|k|\le n,\) that means that for each such \(k, q(k)=p\left(\frac{1}{k}\right)-k^2=0.\) so, \(k\) are roots
of \(q(x).\)

but \(q(x)\) is \(p\left(\frac{1}{x}\right)-x^2,\) and \(p(x)\) is a degree \(2n\) polynomial, so \(p\left(\frac{1}{x}\right)\) is a rational function, not a polynomial,
unless we multiply by \(x^{2n}\) to clear denominators.

so, \(x^{2n}q(x)=x^{2n}p\left(\frac{1}{x}\right)-x^{2n+2}=\left(1+a_{2n-1}x+\cdots+a_0x^{2n}\right)-x^{2n+2}\), which is a degree \(2n+2\)
polynomial if \(x^{2n+2}\) is the highest degree term.

wait, but \(x^{2n+2}\) is degree \(2n+2\) and the other terms are up to degree \(2n\), so yes, the highest degree is \(2n+2\).

now, the roots of \(q(x)\) are the roots of \(p\left(\frac{1}{x}\right)-x^2=0\), which are the \(x\) for which \(p\left(\frac{1}{x}\right)=x^2\). we know that for
\(x=k\) where \(1\le|k|\le n\), this holds, so these \(x\) are roots of \(q(x)\).

but \(q(x)\) is a rational function, while \(x^{2n}q(x)\) is a polynomial of degree \(2n+2\), which has roots at \(x=k\) for \(1\le\)
\(|k|\le n\), assuming that \(x^{2n}q(x)\) has these roots.

wait, actually, since \(q(k)=0\), then \(x^{2n}q(x)\) would have roots at \(x=k\), but \(x^{2n}q(x)\) is a polynomial of degree
\(2n+2\), so it should have \(2n+2\) roots (counting multiplicities) in the complex plane, but we are interested in real
roots.

but perhaps there is a better approach.

let me consider defining a new polynomial that incorporates the given conditions.

given that \(p\left(\frac{1}{k}\right)=k^2\) for \(1\le|k|\le n\), and \(p(x)\) is a monic polynomial of degree \(2n\), perhaps i can express
\(p(x)-\frac{1}{x^2}\) in terms of its roots.

wait, but \(p(x)\) is defined for all real \(x\), and \(\frac{1}{x^2}\) is also defined for \(x\ne 0\).

alternatively, maybe i can consider the function \(p(x)-\frac{1}{x^2}\) , but i'm not sure if that helps directly.

let me try to rearrange the given condition.

given \(p\left(\frac{1}{k}\right)=k^2,\) i can write:

\[p\left(\frac{1}{k}\right)-k^2=0\]

for \(k=\pm 1,\pm 2,\ldots,\pm n.\) so, there are \(2n\) such equations.

since \(p(x)\) is a monic polynomial of degree \(2n\), these \(2n\) equations should determine \(p(x)\) uniquely.
\end{tcolorbox}
\EndSol

\BeginSol
\begin{tcolorbox}[colframe=black, colback=white, boxrule=1pt, arc=0pt]
but i need to find \(x\) such that \(p\left(\frac{1}{x}\right)=x^2,\) which is similar to the condition given, but for arbitrary real \(x\), not just
integer \(k\).

so, perhaps i can consider the equation \(p\left(\frac{1}{x}\right)-x^2=0,\) and find its real roots, given that it already has roots at
\(x=k\) for \(1\le|k|\le n\).

but to find the other roots, i need to know more about this equation.

alternatively, maybe i can consider the polynomial \(q(x)=p(x)-\frac{1}{x^2}\) , but again, that seems messy because of
the \(\frac{1}{x^2}\) term.

wait, perhaps i should consider multiplying both sides of the equation by \(x^2\) to eliminate the denominator.

so, starting from \(p\left(\frac{1}{x}\right)=x^2,\) multiply both sides by \(x^2:\)

\[x^2p\left(\frac{1}{x}\right)=x^4\]

now, \(x^2p\left(\frac{1}{x}\right)\) can be expressed in terms of \(p(x),\) but i'm not sure.

alternatively, since \(p(x)\) is a polynomial, \(p\left(\frac{1}{x}\right)\) is a rational function, and multiplying by \(x^{2n}\) gives a polynomial.

let me consider that again.

define \(q(x)=x^{2n}p\left(\frac{1}{x}\right)-x^{2n+2},\) which is a polynomial of degree \(2n+2,\) as previously established.

we know that \(q(k)=0\) for \(k=\pm 1,\pm 2,\ldots,\pm n,\) because \(p\left(\frac{1}{k}\right)=k^2,\) so \(q(k)=k^{2n}p\left(\frac{1}{k}\right)-k^{2n+2}=k^{2n}\cdot k^2-k^{2n+2}=k^{2n+2}-k^{2n+2}=0.\)

therefore, \(q(x)\) has \(2n\) roots at \(x=\pm 1,\pm 2,\ldots,\pm n.\)

since \(q(x)\) is a degree \(2n+2\) polynomial, there are two more roots beyond these \(2n\) roots.

so, \(q(x)\) can be factored as:

\[q(x)=(x^2-1)(x^2-4)\cdots(x^2-n^2)\cdot(x^2-a)\]

for some constant \(a,\) because the roots come in pairs \(x=k\) and \(x=-k,\) and possibly another pair \(x=\sqrt{a}\) and
\(x=-\sqrt{a}.\)

wait, but \(q(x)\) is degree \(2n+2,\) and \((x^2-k^2)\) for \(k=1\) to \(n\) gives \(n\) quadratic factors, which is degree \(2n\). so,
there are two more roots to account for.

alternatively, perhaps \(q(x)=(x^2-1)(x^2-4)\cdots(x^2-n^2)\cdot(x^2-a)(x^2-b),\) where \(a\) and \(b\) are the
remaining roots.

but, to determine \(a\) and \(b\), i need more information about \(q(x)\).

alternatively, perhaps i can look at the leading coefficient.

since \(q(x)=x^{2n}p\left(\frac{1}{x}\right)-x^{2n+2},\) and \(p\left(\frac{1}{x}\right)=x^{-2n}+a_{2n-1}x^{-2n+1}+\cdots+a_0,\) then \(x^{2n}p\left(\frac{1}{x}\right)=1+\)
\(a_{2n-1}x+\cdots+a_0x^{2n},\) so \(q(x)=1+a_{2n-1}x+\cdots+a_0x^{2n}-x^{2n+2}\).

the leading term is \(-x^{2n+2}+a_0x^{2n}\) but since \(p(x)\) is monic, \(a_0\) is just a constant.

wait, perhaps there's a better approach. maybe i can consider the reciprocal polynomial or something like that.

alternatively, since \(p(x)\) is a polynomial of degree \(2n,\) and i have \(p\left(\frac{1}{k}\right)=k^2\) for \(k=\pm 1,\pm 2,\ldots,\pm n,\) perhaps i
can consider defining a new polynomial that has roots at \(x=\frac{1}{k},\) but that seems off-track.

let me try to think about the problem differently.

suppose i define a new variable \(y=\frac{1}{x},\) then, the equation \(p\left(\frac{1}{x}\right)=x^2\) becomes \(p(y)=\left(\frac{1}{y}\right)^2=\frac{1}{y^2},\) assuming
\(y\ne 0.\)

so, \(p(y)=\frac{1}{y^2},\) or \(p(y)y^2=1.\)
\end{tcolorbox}
\EndSol

\BeginSol
\begin{tcolorbox}[colframe=black, colback=white, boxrule=1pt, arc=0pt]
so, \(p(y)y^2-1=0.\)

now, \(p(y)y^2-1\) is a polynomial of degree \(2n+2,\) since \(p(y)\) is degree \(2n\) and multiplied by \(y^2.\)

given that \(p\left(\frac{1}{k}\right)=k^2,\) for \(k=\pm 1,\pm 2,\ldots,\pm n,\) that means \(p(y)=\frac{1}{y^2}\) at \(y=\frac{1}{k},\) so \(p(y)y^2-1=0\) at \(y=\frac{1}{k}.\)

therefore, \(y=\frac{1}{k}\) for \(k=\pm 1,\pm 2,\ldots,\pm n\) are roots of \(p(y)y^2-1=0.\)

but \(p(y)y^2-1\) is a degree \(2n+2\) polynomial, so there are \(2n+2\) roots in total (counting multiplicities).

we already have \(2n\) roots at \(y=\frac{1}{k}\) for \(k=\pm 1,\pm 2,\ldots,\pm n,\) so there are two more roots that i need to find.

let me denote these additional roots as \(y=\alpha\) and \(y=\beta,\) so:

\[p(y)y^2-1=\left(y-\frac{1}{1}\right)\left(y+\frac{1}{1}\right)\cdots\left(y-\frac{1}{n}\right)\left(y+\frac{1}{n}\right)(y-\alpha)(y-\beta)\]

but since \(p(y)y^2-1\) is a polynomial of degree \(2n+2,\) and we have \(2n\) roots from \(y=\pm\frac{1}{k},\) there must be two
more roots, say \(y=\alpha\) and \(y=\beta.\)

now, to find \(\alpha\) and \(\beta,\) perhaps i can look at the behavior of \(p(y)y^2-1\) as \(y\) approaches infinity.

the leading term of \(p(y)y^2\) is \(y^{2n}\cdot y^2=y^{2n+2},\) and the constant term is \(-1\), so as \(y\to\infty,\) \(p(y)y^2-1\approx y^{2n+2}\)
, which goes to infinity.

similarly, as \(y\to 0,\) \(p(y)y^2-1\approx a_0y^2-1,\) which approaches \(-1.\)

but i'm not sure if that helps directly.

alternatively, perhaps i can consider that \(p(y)y^2-1\) has a double root at \(y=0,\) but that seems unlikely.

wait, let's consider the multiplicity of roots.

given that \(p(y)y^2-1=0\) has roots at \(y=\pm\frac{1}{k}\) for \(k=1\) to \(n\), and two additional roots \(y=\alpha\) and \(y=\beta,\) i
need to determine what \(\alpha\) and \(\beta\) are.

alternatively, perhaps i can consider expressing \(p(y)y^2-1\) in terms of its roots.

but that seems complicated.

maybe there's a better way to approach this problem.

let me try to consider the function \(p\left(\frac{1}{x}\right)-x^2\) and see if i can find a relationship between \(x\) and \(\frac{1}{x}.\)

alternatively, perhaps i can consider substituting \(x=\frac{1}{k}\) into the equation \(p\left(\frac{1}{x}\right)=x^2,\) but that seems to just give
back the original condition.

wait, perhaps i can consider the fact that \(p(x)\) is a polynomial, and consider its reciprocal polynomial.

the reciprocal polynomial of \(p(x)\) is \(x^{2n}p\left(\frac{1}{x}\right),\) which is indeed a polynomial, as we've seen earlier.

let me denote this reciprocal polynomial as \(r(x)=x^{2n}p\left(\frac{1}{x}\right),\) which is a degree \(2n\) polynomial.

but earlier, i had \(q(x)=p\left(\frac{1}{x}\right)-x^2,\) and \(x^{2n}q(x)=r(x)-x^{2n+2},\) which is a degree \(2n+2\) polynomial.

but perhaps focusing on \(r(x)\) directly might help.

given that \(p\left(\frac{1}{k}\right)=k^2,\) then \(r(k)=k^{2n}p\left(\frac{1}{k}\right)=k^{2n}\cdot k^2=k^{2n+2}.\)

wait, but earlier i had \(q(k)=0,\) which led to \(r(k)-k^{2n+2}=0,\) so \(r(k)=k^{2n+2}.\)

therefore, \(r(k)-k^{2n+2}=0,\) meaning that \(r(x)-x^{2n+2}\) has roots at \(x=k\) for \(k=\pm 1,\pm 2,\ldots,\pm n.\)
\end{tcolorbox}
\EndSol

\BeginSol
\begin{tcolorbox}[colframe=black, colback=white, boxrule=1pt, arc=0pt]
but \(r(x)\) is a degree \(2n\) polynomial, and \(x^{2n+2}\) is a monomial of degree \(2n+2,\) so \(r(x)-x^{2n+2}\) is a degree
\(2n+2\) polynomial with \(2n\) roots at \(x=\pm 1,\pm 2,\ldots,\pm n,\) and two additional roots.

wait, but \(r(x)\) is degree \(2n,\) and \(x^{2n+2}\) is degree \(2n+2,\) so their difference is degree \(2n+2,\) as previously
established.

but if \(r(x)-x^{2n+2}\) has \(2n\) roots at \(x=\pm 1,\pm 2,\ldots,\pm n,\) and it's a degree \(2n+2\) polynomial, then there are
two more roots.

now, perhaps i can factor \(r(x)-x^{2n+2}\) as:

\[r(x)-x^{2n+2}=(x^2-1)(x^2-4)\cdots(x^2-n^2)\cdot(x^2-a)\]

for some constant \(a,\) since the roots come in pairs \(x=\pm k\) for \(k=1\) to \(n,\) and possibly another pair \(x=\pm\sqrt{a}.\)

but i need to find \(a\).

alternatively, perhaps i can look at the behavior of \(r(x)-x^{2n+2}\) at specific points to determine \(a\).

but this seems complicated. maybe there's a different approach.

let me consider that \(p(x)\) is a monic polynomial of degree \(2n,\) and \(p\left(\frac{1}{k}\right)=k^2\) for \(k=\pm 1,\pm 2,\ldots,\pm n.\)

perhaps i can consider the polynomial \(p(x)-\frac{1}{x^2},\) but as before, that introduces a rational function.

alternatively, maybe i can consider building \(p(x)\) explicitly.

since \(p(x)\) is monic of degree \(2n,\) and it satisfies \(p\left(\frac{1}{k}\right)=k^2\) for \(k=\pm 1,\pm 2,\ldots,\pm n,\) perhaps i can write \(p(x)\)
in terms of these roots.

wait, but \(p\left(\frac{1}{k}\right)=k^2,\) which is like saying \(p\left(\frac{1}{k}\right)-k^2=0,\) meaning that \(\frac{1}{k}\) are roots of the equation \(p(x)-\)
\(\frac{1}{x^2}=0,\) but again, that's a rational function.

alternatively, perhaps i can consider that \(p(x)-\frac{1}{x^2}\) has roots at \(x=\frac{1}{k}\) for \(k=\pm 1,\pm 2,\ldots,\pm n,\) and try to
factor it accordingly.

but this seems tricky because of the \(\frac{1}{x^2}\) term.

maybe a better approach is to consider the polynomial \(x^2p(x)-1,\) which is a degree \(2n+2\) polynomial.

given that \(p\left(\frac{1}{k}\right)=k^2,\) then \(x^2p(x)-1\) evaluated at \(x=\frac{1}{k}\) gives \(\left(\frac{1}{k}\right)^2p\left(\frac{1}{k}\right)-1=\frac{1}{k^2}\cdot k^2-1=1-1=0,\)
so \(x=\frac{1}{k}\) are roots of \(x^2p(x)-1=0.\)

therefore, \(x^2p(x)-1\) is a degree \(2n+2\) polynomial with roots at \(x=\pm\frac{1}{1},\pm\frac{1}{2},\ldots,\pm\frac{1}{n},\) and possibly two more
roots.

wait, but \(x^2p(x)-1\) is a polynomial, and its roots are the solutions to \(x^2p(x)=1,\) which is similar to the original
equation \(p\left(\frac{1}{x}\right)=x^2.\)

so, perhaps i can consider the roots of \(x^2p(x)-1=0,\) which are the same as the solutions to \(p\left(\frac{1}{x}\right)=x^2\) due
to the substitution \(y=\frac{1}{x}.\)

therefore, finding the roots of \(x^2p(x)-1=0\) is equivalent to finding \(x\) such that \(p\left(\frac{1}{x}\right)=x^2.\)

given that, and knowing that \(p(x)\) is a monic polynomial of degree \(2n,\) and that \(p\left(\frac{1}{k}\right)=k^2\) for \(k=\)
\(\pm 1,\pm 2,\ldots,\pm n,\) perhaps i can find a relationship between \(p(x)\) and these roots.

alternatively, perhaps i can consider the lagrange interpolation formula for \(p(x),\) given its values at specific points.

since \(p(x)\) is a monic polynomial of degree \(2n,\) and i have its values at \(2n\) points, i can indeed find \(p(x)\) explicitly
using interpolation, but that seems complicated for this problem.
\end{tcolorbox}
\EndSol

\BeginSol
\begin{tcolorbox}[colframe=black, colback=white, boxrule=1pt, arc=0pt]
moreover, the problem is not asking for \(p(x),\) but for the real numbers \(x\) that satisfy \(p\left(\frac{1}{x}\right)=x^2,\) beyond the
given \(x=k\) for \(k=\pm 1,\pm 2,\ldots,\pm n.\)

so, perhaps there is a symmetry or a particular property of \(p(x)\) that i can exploit.

wait, perhaps i can consider that \(p(x)\) and \(p\left(\frac{1}{x}\right)\) are related in a specific way.

given that \(p(x)\) is a polynomial, \(p\left(\frac{1}{x}\right)\) is a rational function, and multiplying by \(x^{2n}\) gives a polynomial.

alternatively, perhaps there is a functional equation that \(p(x)\) satisfies.

let me consider writing \(p(x)\) in terms of its roots.

but \(p(x)\) is a monic polynomial of degree \(2n\), so it can be written as:

\[p(x)=(x-r_1)(x-r_2)\cdots(x-r_{2n})\]

for some roots \(r_1,r_2,\ldots,r_{2n}.\)

but i don't know these roots, so that might not help directly.

alternatively, perhaps i can consider the fact that \(p(x)\) has certain symmetries given the condition \(p\left(\frac{1}{k}\right)=k^2.\)

wait, maybe i can consider the function \(p(x)-\frac{1}{x^2},\) but again, that introduces a rational function.

alternatively, perhaps i can consider the derivative of \(p(x),\) but i don't see an immediate connection.

let me try to think differently.

suppose i define a new function \(f(x)=p\left(\frac{1}{x}\right)-x^2,\) as i did earlier.

then, \(f(x)=0\) for \(x=\pm 1,\pm 2,\ldots,\pm n,\) as given.

i need to find all real \(x\) such that \(f(x)=0,\) beyond these known roots.

now, \(f(x)\) is a rational function, but \(x^{2n}f(x)\) is a polynomial of degree \(2n+2,\) as established earlier.

therefore, \(x^{2n}f(x)=0\) has roots at \(x=\pm 1,\pm 2,\ldots,\pm n,\) and possibly two more roots.

so, if i can find this polynomial \(x^{2n}f(x),\) i can factor it and find the additional roots.

but to find \(x^{2n}f(x),\) i need to know more about \(p(x),\) which i don't.

alternatively, perhaps i can consider the behavior of \(f(x)\) as \(x\) approaches infinity or zero.

as \(x\to\infty,\) \(p\left(\frac{1}{x}\right)\) behaves like \(\left(\frac{1}{x}\right)^{2n},\) which approaches 0, while \(x^2\) approaches infinity, so \(f(x)\to-\infty.\)

as \(x\to 0,\) \(p\left(\frac{1}{x}\right)\) behaves like \(\left(\frac{1}{x}\right)^{2n},\) which goes to infinity, while \(x^2\) approaches 0, so \(f(x)\to\infty.\)

this suggests that there are additional roots beyond the given \(x=\pm 1,\pm 2,\ldots,\pm n,\) possibly at infinity or near
zero, but i need to find them explicitly.

wait, perhaps there's a smarter way to approach this.

given that \(p(x)\) is a monic polynomial of degree \(2n,\) and it satisfies \(p\left(\frac{1}{k}\right)=k^2\) for \(k=\pm 1,\pm 2,\ldots,\pm n,\)
perhaps i can consider the minimal polynomial that satisfies these conditions.

alternatively, maybe i can consider that \(p(x)\) can be expressed in terms of a product involving these roots.

wait, perhaps i can consider that \(p(x)-\frac{1}{x^2}\) has roots at \(x=\pm\frac{1}{1},\pm\frac{1}{2},\ldots,\pm\frac{1}{n},\) but again, that introduces a
rational function.

alternatively, perhaps i can consider that \(x^2p(x)-1\) has roots at \(x=\pm\frac{1}{1},\pm\frac{1}{2},\ldots,\pm\frac{1}{n},\) as previously
established, and try to find a general form for this polynomial.

let me consider that \(x^2p(x)-1\) is a degree \(2n+2\) polynomial with roots at \(x=\pm\frac{1}{1},\pm\frac{1}{2},\ldots,\pm\frac{1}{n},\) and two
additional roots.
\end{tcolorbox}
\EndSol

\BeginSol
\begin{tcolorbox}[colframe=black, colback=white, boxrule=1pt, arc=0pt]
given that, perhaps i can express \(x^2p(x)-1\) as:

\[x^2p(x)-1=\prod_{k=1}^n\left(x^2-\frac{1}{k^2}\right)\cdot(x^2-a)\]

for some constant \(a,\) since the roots come in pairs \(x=\pm\frac{1}{k}\) and possibly \(x=\pm\sqrt{a}.\)

but i need to determine \(a.\)

alternatively, perhaps i can consider that the product \(\prod_{k=1}^n\left(x^2-\frac{1}{k^2}\right)\) is a degree \(2n\) polynomial, and \(x^2p(x)-\)
\(1\) is degree \(2n+2,\) so there are two more roots.

but without knowing more about \(p(x),\) it's hard to pin down \(a.\)

wait, perhaps i can consider the behavior at \(x=0.\)

at \(x=0,\) \(x^2p(x)-1=-1,\) so \(x=0\) is not a root.

similarly, as \(x\to\infty,\) \(x^2p(x)-1\) behaves like \(x^{2n+2},\) which goes to infinity.

but i still don't see how to find \(a.\)

alternatively, perhaps i can consider the value of \(x^2p(x)-1\) at a specific point, say \(x=1,\) but that's already a
root.

wait, perhaps i can consider \(x=\infty,\) but that's not helpful.

alternatively, perhaps i can consider that \(p(x)\) is determined by its values at \(2n\) points, so i can use interpolation
to find \(p(x),\) but that seems too involved for this problem.

alternatively, perhaps there's a symmetry in the equation that i can exploit.

given that \(p(x)\) is a polynomial of even degree \(2n,\) and the conditions are given for \(k=\pm 1,\pm 2,\ldots,\pm n,\)
perhaps \(p(x)\) has some symmetry properties.

for example, maybe \(p(x)\) is an even function, but i don't know that for sure.

wait, let's check: if \(p(x)\) were even, then \(p(-x)=p(x),\) but the conditions are given for both positive and
negative \(k,\) so perhaps that's a clue.

given that \(p\left(\frac{1}{k}\right)=k^2\) for both \(k\) and \(-k,\) and \(k^2\) is the same for both, it suggests that \(p(x)\) might be even, but i
need to confirm.

however, \(p(x)\) is a general monic polynomial of degree \(2n,\) and its coefficients could make it even or not.

but in this case, since \(p\left(\frac{1}{k}\right)=p\left(\frac{1}{-k}\right)=k^2,\) which is the same, it suggests that \(p\left(\frac{1}{x}\right)\) is even in \(x,\) but \(p(x)\)
itself doesn't have to be even.

alternatively, perhaps i can consider that \(p(x)\) has a reciprocal property, given the conditions involving \(\frac{1}{k}.\)

wait, perhaps i should consider the reciprocal polynomial of \(p(x),\) which is \(x^{2n}p\left(\frac{1}{x}\right)\) as before.

let me denote this as \(r(x)=x^{2n}p\left(\frac{1}{x}\right),\) which is a degree \(2n\) polynomial.

given that, and knowing that \(p\left(\frac{1}{k}\right)=k^2,\) then \(r(k)=k^{2n}p\left(\frac{1}{k}\right)=k^{2n}\cdot k^2=k^{2n+2}.\)

therefore, \(r(k)=k^{2n+2}\) for \(k=\pm 1,\pm 2,\ldots,\pm n.\)

now, since \(r(x)\) is a degree \(2n\) polynomial, and it equals \(x^{2n+2}\) at \(2n\) points, which is more points than the
degree, unless there's some dependency.

wait, but \(r(x)\) is degree \(2n,\) and \(x^{2n+2}\) is degree \(2n+2,\) so they can't be equal at more points than their degree
difference unless they are the same polynomial, which they aren't.

this suggests that my earlier assumption is incorrect, or perhaps i need to consider that \(r(x)-x^{2n+2}\) is a
polynomial of degree \(2n+2,\) but that seems inconsistent because \(r(x)\) is degree 2n and \(x^{2n+2}\) is degree \(2n+\)
\(2,\) so their difference should be degree \(2n+2.\)

wait, perhaps i should correct that.

given \(r(x)=x^{2n}p\left(\frac{1}{x}\right),\) which is degree \(2n,\) and \(x^{2n+2}\) is degree \(2n+2,\) their difference \(r(x)-x^{2n+2}\) is
degree \(2n+2,\) as the highest degree term comes from \(-x^{2n+2}.\)
\end{tcolorbox}
\EndSol

\BeginSol
\begin{tcolorbox}[colframe=black, colback=white, boxrule=1pt, arc=0pt]
given that \(r(k)=k^{2n+2}\) for \(k=\pm 1,\pm 2,\ldots,\pm n,\) then \(r(x)-x^{2n+2}=0\) at these points.

therefore, \(r(x)-x^{2n+2}\) has \(2n\) roots at \(x=\pm 1,\pm 2,\ldots,\pm n,\) and since it's degree \(2n+2,\) there are two more
roots.

now, perhaps i can factor \(r(x)-x^{2n+2}\) as:

\[r(x)-x^{2n+2}=(x^2-1)(x^2-4)\cdots(x^2-n^2)\cdot(x^2-a)\]

for some constant \(a,\) since the roots come in pairs \(x=\pm k,\) and possibly another pair \(x=\pm\sqrt{a}.\)

now, to find \(a,\) perhaps i can consider the behavior at infinity or use the fact that the leading coefficients must
match.

given that \(r(x)=x^{2n}p\left(\frac{1}{x}\right)\) and \(p(x)\) is monic of degree \(2n,\) the leading term of \(p(x)\) is \(x^{2n},\) so \(p\left(\frac{1}{x}\right)=\)
\(x^{-2n}+\cdots,\) and thus \(r(x)=x^{2n}\cdot x^{-2n}+\cdots=1+\cdots,\) so the leading term of \(r(x)\) is 1.

on the other hand, \(x^{2n+2}\) has leading term \(x^{2n+2},\) so \(r(x)-x^{2n+2}\) has leading term \(-x^{2n+2}.\)

now, the right-hand side, \((x^2-1)(x^2-4)\cdots(x^2-n^2)(x^2-a),\) expands to \(x^{2n+2}-(1+4+\cdots+n^2+\)
\(a)x^{2n}+\cdots.\)

wait, actually, the product \((x^2-1)(x^2-4)\cdots(x^2-n^2)(x^2-a)\) would have leading term \(x^{2n+2},\) and the
coefficient of \(x^{2n}\) would be \(-(1+4+\cdots+n^2+a).\)

but \(r(x)-x^{2n+2}\) has leading term \(-x^{2n+2},\) and the coefficient of \(x^{2n}\) would be something else, depending on

on the other hand, \(x^{2n+2}\) has leading term \(x^{2n+2},\) so \(r(x)-x^{2n+2}\) has leading term \(-x^{2n+2}.\)

now, the right-hand side, \((x^2-1)(x^2-4)\cdots(x^2-n^2)(x^2-a),\) expands to \(x^{2n+2}-(1+4+\cdots+n^2+\)
\(a)x^{2n}+\cdots.\)

wait, actually, the product \((x^2-1)(x^2-4)\cdots(x^2-n^2)(x^2-a)\) would have leading term \(x^{2n+2},\) and the
coefficient of \(x^{2n}\) would be \(-(1+4+\cdots+n^2+a).\)

but \(r(x)-x^{2n+2}\) has leading term \(-x^{2n+2},\) and the coefficient of \(x^{2n}\) would be something else, depending on
the expansion of \(r(x).\)

therefore, to match the leading coefficients, i must have:

\[-x^{2n+2}=x^{2n+2}-(1+4+\cdots+n^2+a)x^{2n}+\cdots\]

this implies that:

\[-x^{2n+2}=x^{2n+2}-(1+4+\cdots+n^2+a)x^{2n}+\cdots\]

comparing coefficients, the coefficient of \(x^{2n+2}\) on the left is \(-1,\) and on the right is 1. this suggests that there's
an inconsistency unless i made a mistake in my earlier steps.

wait, perhaps i need to reconsider the factorization.

given that \(r(x)-x^{2n+2}=(x^2-1)(x^2-4)\cdots(x^2-n^2)(x^2-a),\) and the leading term on the left is \(-x^{2n+2},\)
while on the right, it's \(x^{2n+2},\) this suggests that there is a sign discrepancy.

therefore, perhaps the correct factorization should include a negative sign:

\[r(x)-x^{2n+2}=-(x^2-1)(x^2-4)\cdots(x^2-n^2)(x^2-a)\]

this way, the leading coefficient matches: \(-x^{2n+2}=-x^{2n+2}.\)

now, to find \(a,\) perhaps i can consider the constant term.

the constant term of \(r(x)-x^{2n+2}\) is \(r(0)-0=r(0).\)
\end{tcolorbox}
\EndSol

\BeginSol
\begin{tcolorbox}[colframe=black, colback=white, boxrule=1pt, arc=0pt]
but \(r(x)=x^{2n}p\left(\frac{1}{x}\right),\) so \(r(0)=\lim_{x\to 0}x^{2n}p\left(\frac{1}{x}\right).\)

but \(p\left(\frac{1}{x}\right)\) behaves like \(\left(\frac{1}{x}\right)^{2n}\) as \(x\to 0,\) so \(x^{2n}\cdot\left(\frac{1}{x}\right)^{2n}=1,\) so \(r(0)=1.\)

on the other hand, the constant term of \(-(x^2-1)(x^2-4)\cdots(x^2-n^2)(x^2-a)\) is \(-(-1)^n\cdot 1\cdot 4\cdot 9\cdots n^2\cdot a.\)

wait, actually, expanding the product:

\((x^2-1)(x^2-4)\cdots(x^2-n^2)(x^2-a)=x^{2n+2}-(1+4+\cdots+n^2+a)x^{2n}+\cdots+(-1)^{n+1}(1\cdot 4\cdot 9\cdots n^2\cdot a)\)

therefore, the constant term is \((-1)^{n+1}(1\cdot 4\cdot 9\cdots n^2\cdot a).\)

but since \(r(x)-x^{2n+2}\) has constant term \(r(0)=1,\) and the right-hand side has constant term \(-(-1)^{n+1}(1\cdot\)
\(4\cdot 9\cdots n^2\cdot a)=(-1)^n(1\cdot 4\cdot 9\cdots n^2\cdot a),\) we have:

\[(-1)^n(1\cdot 4\cdot 9\cdots n^2\cdot a)=1\]

therefore:

\[a=\frac{1}{(-1)^n(1\cdot 4\cdot 9\cdots n^2)}\]

but \(1\cdot 4\cdot 9\cdots n^2=(n!)^2,\) since \(1=1^2,4=2^2,9=3^2,\) etc., up to \(n^2.\)

therefore:

\[a=\frac{1}{(-1)^n(n!)^2}\]

but since \(a\) is a real number, and \(n\) is even (as given in the problem), because \(n\) is an even positive integer,
\((-1)^n=1,\) so:

\[a=\frac{1}{(n!)^2}\]

therefore, the factorization is:

\[r(x)-x^{2n+2}=-\left(x^2-1\right)\left(x^2-4\right)\cdots\left(x^2-n^2\right)\left(x^2-\frac{1}{(n!)^2}\right)\]

now, since \(r(x)=x^{2n}p\left(\frac{1}{x}\right),\) and \(r(x)-x^{2n+2}=0\) when \(p\left(\frac{1}{x}\right)=x^2,\) the roots of this equation are \(x=\)
\(\pm 1,\pm 2,\ldots,\pm n,\pm\frac{1}{n!}.\)

wait, no. actually, from the factorization, the roots are \(x=\pm 1,\pm 2,\ldots,\pm n,\pm\frac{1}{n!}\) because
\(x^2-\frac{1}{(n!)^2}=0\) gives

\[x=\pm\frac{1}{n!}.\]

therefore, the solutions to \(p\left(\frac{1}{x}\right)=x^2\) are \(x=\pm 1,\pm 2,\ldots,\pm n,\pm\frac{1}{n!}.\)

but the problem states that \(p\left(\frac{1}{k}\right)=k^2\) for \(1\le|k|\le n,\) so these \(x=\pm 1,\pm 2,\ldots,\pm n\) are already known
solutions.

therefore, the "other real numbers" \(x\) that satisfy \(p\left(\frac{1}{x}\right)=x^2\) are \(x=\pm\frac{1}{n!}.\)

wait, but in the factorization, i have

\[x^2-\frac{1}{(n!)^2},\]

which gives \(x=\pm\frac{1}{n!},\) assuming that's correct.

but i need to verify this conclusion.
\end{tcolorbox}
\EndSol

\BeginSol
\begin{tcolorbox}[colframe=black, colback=white, boxrule=1pt, arc=0pt]
given that \(a=\frac{1}{(n!)^2}\) and the factorization is

\[r(x)-x^{2n+2}=-\left(x^2-1\right)\left(x^2-4\right)\cdots\left(x^2-n^2\right)\left(x^2-\frac{1}{(n!)^2}\right),\]

then indeed, the roots are \(x=\pm 1,\pm 2,\ldots,\pm n,\pm\frac{1}{n!}.\)

therefore, the additional solutions beyond \(x=\pm 1,\pm 2,\ldots,\pm n\) are

\[x=\pm\frac{1}{n!}.\]

but wait, is \(n!\) correct here? let me double-check the calculation of \(a.\)

earlier, i had:

\[a=\frac{1}{(-1)^n(1\cdot 4\cdot 9\cdots n^2)}=\frac{1}{(n!)^2}\]

since \(n\) is even, \((-1)^n=1,\) so indeed, \(a=\frac{1}{(n!)^2}.\)

therefore, the additional roots are \(x=\pm\frac{1}{n!}.\)

hence, the other real numbers \(x\) that satisfy \(p\left(\frac{1}{x}\right)=x^2\) are \(x=\pm\frac{1}{n!}.\)

**final answer**

\[x=\pm\frac{1}{n!}\]

\end{tcolorbox}
\EndSol

\end{document}